\documentclass[journal]{IEEEtran}

\usepackage{adjustbox}
\usepackage{bbding}
\usepackage{url}
\usepackage{color}
\usepackage{hyperref}
\usepackage{multirow}
\usepackage{graphicx}
\usepackage{amsmath}
\usepackage{amssymb}
\newcommand{\etal}{\emph{et al.}~}
\newcommand{\ie}{\emph{i.e.}}
\newcommand{\eg}{\emph{e.g.}}

\begin{document}

\title{Image Fine-grained Inpainting}

\author{Zheng~Hui,
        Jie~Li,
        Xinbo~Gao,~\IEEEmembership{Senior~Member,~IEEE}
        and~Xiumei~Wang% <-this % stops a space
\thanks{The authors are with the Video \& Image Processing System (VIPS) Lab, School of Electronic Engineering, Xidian University, No.2, South Taibai Road, Xi'an 710071, China. (e-mail: zheng\_hui@aliyun.com, leejie@mail.xidian.edu.cn, xbgao@mail.xidian.edu.cn, wangxm@xidian.edu.cn)}}% <-this % stops a space

\markboth{A correspondence submitted to Journal}%
{Hui \MakeLowercase{\textit{et al.}}: Image fine-grained inpainting}

\maketitle

\begin{abstract}
Image inpainting techniques have shown promising improvement with the assistance of generative adversarial networks (GANs) recently. However, most of them often suffered from completed results with unreasonable structure or blurriness. To mitigate this problem, in this paper, we present a one-stage model that utilizes dense combinations of dilated convolutions to obtain larger and more effective receptive fields. Benefited from the property of this network, we can more easily recover large regions in an incomplete image. To better train this efficient generator, except for frequently-used VGG feature matching loss, we design a novel self-guided regression loss for concentrating on uncertain areas and enhancing the semantic details. Besides, we devise a geometrical alignment constraint item (feature center coordinates alignment) to compensate for the pixel-based distance between prediction features and ground-truth ones. We also employ a discriminator with local and global branches to ensure local-global contents consistency. To further improve the quality of generated images, discriminator feature matching on the local branch is introduced, which dynamically minimizes the similarity of intermediate features between synthetic and ground-truth patches. Extensive experiments on several public datasets demonstrate that our approach outperforms current state-of-the-art methods. Code is available at~\url{https://github.com/Zheng222/DMFN}.
\end{abstract}


\IEEEpeerreviewmaketitle

\section{Introduction}
Image inpainting (a.k.a. image completion) aims to synthesize proper contents in missing regions of an image, which can be used in many applications. For instance, it allows removing unwanted objects in image editing tasks, while filling the contents that are visually realistic and semantically correct. Early approaches to image inpainting are mostly based on patches of low-level features. PatchMatch~\cite{patch-match}, a typical method, iteratively searches optimal patches to fill in the holes. It can produce plausible results when painting image background or repetitive textures. However, it cannot generate pleasing results for cases where completing regions include complex scenes, faces, and objects, which is due to PatchMatch cannot synthesize new image contents, and missing patches cannot be found in remaining regions for challenging cases.

With the rapid development of deep convolutional neural networks (CNN) and generative adversarial networks (GAN)~\cite{GAN}, image inpainting approaches have achieved remarkable success. Pathak~\etal proposed context-encoder~\cite{CE}, which employs a deep generative model to predict missing parts of the scene from their surroundings using reconstruction and adversarial losses. Yang~\etal~\cite{high-resolution-multi-scale} introduced style transfer into image inpainting to improve textural quality that propagates the high-frequency textures from the boundary to the hole. Li~\etal~\cite{GFC} presented semantic parsing in the generation to restrict synthesized semantically valid contents for the missing facial key parts from random noise. To be able to complete large regions, Iizuka~\etal~\cite{globally-and-locally} adopted stacked dilated convolutions in their image completion network to obtain lager spatial support and reached realistic results with the assistance of a globally and locally consistent adversarial training approach. Shortly afterward, Yu~\etal~\cite{contextual-attention} extended this insight and developed a novel contextual attention layer, which uses the features of known patches as convolutional kernels to compute the correlation between the foreground and background patches. More specifically, they calculated attention score for each pixel and then performed transposed convolution on attention score to reconstruct missing patches with known patches. It might be failing when the relationship between unknown and known patches is not close (\eg masking all of the critical components of a facial image). Wang~\etal~\cite{GMCNN} proposed a generative multi-column convolutional neural network (GMCNN) that uses varied receptive fields in branches by adopting different sizes of convolution kernels (\ie~$3\times 3$, $ 5\times 5$, and $7 \times 7$) in a parallel manner. This method produces advanced performance but suffers from substantial model parameters (12.562M) caused by large convolution kernels. In terms of image quality (more photo-realistic, fewer artifacts), it is still room for improvement.

The goals pursued by image inpainting are ensuring produced images with global semantic structure and finely detailed textures. Additionally, completed image should be approaching the ground truth as much as possible, especially for building and face images. Previous techniques more focus on solving how to yield holistically reasonable and photo-realistic images. This problem has been mitigated by GAN~\cite{GAN} or its improved version WGAN-GP~\cite{WGAN-GP} that is frequently utilized in image inpainting methods~\cite{CE,globally-and-locally,high-resolution-multi-scale,GFC,contextual-attention,contextual-based,Shift-Net,GMCNN,PICNet,PEN-Net}. However, concerning fine-grained details, there is still much room to enhance. Besides, these existing methods haven't taken into account the consistency between outputs and targets,~\ie, semantic structures should be as much similar as possible for facial images and building images.

\begin{figure*}[htpb]
	\centering
	\includegraphics[width=\textwidth]{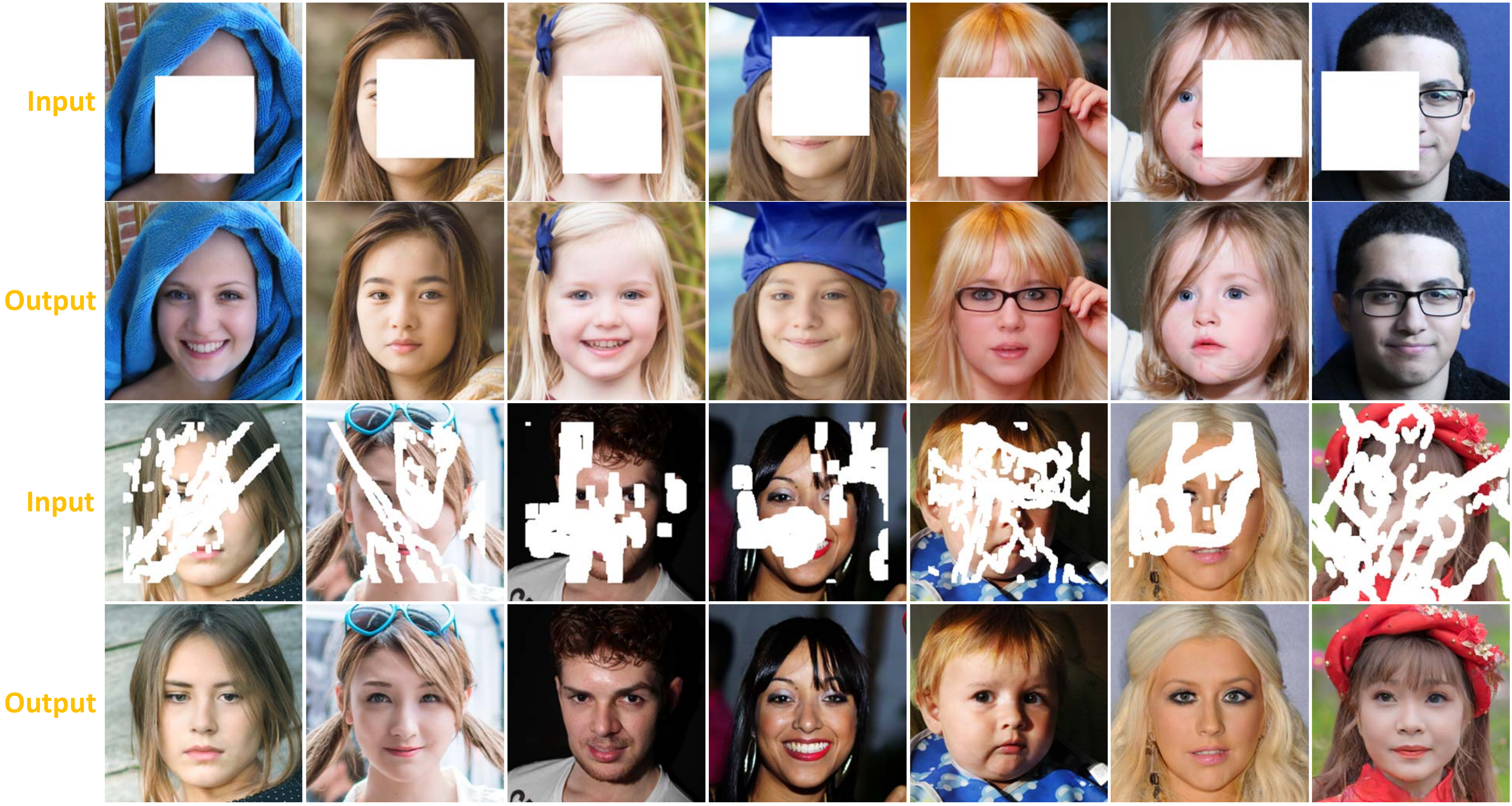}
	\caption{The inpainted results on FFHQ dataset~\cite{ffhq} by using our method. The missing areas are shown in white. \textit{It is worth noting that they also recover well in terms of lighting and texture.}}
	\label{fig:teaser}
\end{figure*}

To overcome the limitations of the methods as mentioned above, we present a unified generative network for image inpainting, which is denoted as \textit{dense multi-scale fusion network} (DMFN). The \textit{dense multi-scale fusion block} (DMFB), serving as the basic block of DMFN, is composed of four-way dilated convolutions as illustrated in Figure~\ref{fig:DMFB}. This basic block adopts the combination and fusion of hierarchical features extracted from various convolutions with different dilation rates to obtain better multi-scale features, compared with general dilated convolution (dense v.s. sparse). For generating images with the realistic and semantic structure, we design a \textit{self-guided regression loss} that constrains low-level features of the generated content according to the normalized discrepancy map (the difference between the output and target). \textit {Geometrical alignment constraint} is developed for penalizing the coordinate center of estimated image high-level features away from the ground-truth. This loss can further help the processing of image fine-grained inpainting. We improve the discriminator using relativistic average GAN (RaGAN)~\cite{RaGAN}. It is noteworthy that we use global and local branches in the discriminator as in~\cite{globally-and-locally}, where one branch focuses on the global image while the other concentrates on the local patch of the missing region. To explicitly constraint the output and ground-truth images, we utilize the hidden layers of the local branch that belongs to the whole discriminator to evaluate their discrepancy through an adversarial training process. With all these improvements, the proposed method can produce high-quality results on multiple datasets, including faces, building, and natural scene images.

\begin{table*}[htpb]
	\centering
	\caption{Image inpainting methodology employed by some representative models.}
	\label{tab:methods_summary}
	\scalebox{0.9}{
		\begin{tabular}{|r|c|c|c|}
			\hline
			Method & GMCNN (NeurIPS'2018)~\cite{GMCNN} & GC (ICCV'2019)~\cite{GC} & DMFN (Ours)\\
			\hline
			\hline
			Stage & one-stage & two-stage & one-stage \\
			\hline
			Generator & multi-column CNNs & gated CNN & dense multi-scale fusion network \\
			\hline
			Discriminator & WGAN-GP & SN-PatchGAN & RelativisticGAN \\
			\hline
			Losses & reconstruction + adversarial + ID-MRF & $l1$ reconstruction + SN-PatchGAN  & $l1$ + self-guided regression + fm\_vgg + fm\_dis + adversarial + alignment \\
			\hline
			
	\end{tabular} }
\end{table*}

Our contributions are summarized as follows:
\begin{itemize}
	
	\item Self-guided regression loss corrects semantic structure errors to some extent through re-weighing VGG features guided by discrepancy map, which is novel for image/video completion task.
	
	\item We present the geometrical alignment constraint to supplement the shortage of pixel-based VGG features matching loss, which restrains the results with a more reasonable semantic spatial location.
	
	\item We propose the dense multiple fusion block (DMFB, enhancing the dilated convolution) to improve the network representation, which increases the receptive field while maintaining an acceptable parameter size. Our generative image inpainting framework achieves compelling visual results (as illustrated in Figure~\ref{fig:teaser}) on challenging datasets.
\end{itemize}
As shown in Table~\ref{tab:methods_summary}, we summarized the difference between the typical method and the proposed approach. We are committed to improving the dilated convolution that frequently used in image completion, and developing more losses to measure the matching degree of features from different viewpoint. More details can be found in Section~\ref{sec:proposed-method}.

The rest of this paper is organized as follows. Section~\ref{sec:related-work} provides a brief review of related inpainting methods. Section~\ref{sec:proposed-method} describes the proposed approach and loss functions in detail. In Section~\ref{sec:experiments}, we explain the experiments conducted for this work, experimental comparisons with other state-of-the-art methods, and model analysis. In Section~\ref{sec:conclusion}, we conclude the study.

\section{Related Work}\label{sec:related-work}
A variety of algorithms for image inpainting have been proposed. Traditional diffusion-based methods~\cite{image-inpainting,joint-interpolation} propagate information from neighboring regions to the holes. They can work well for small and narrow holes, where the texture and color variance are the same. However, these methods fail to recover meaning contents in the large missing regions. Patch-based approaches, such as~\cite{image-quilting,texture-optimization}, search for relevant patches from the known regions in an iterative fashion. Simakov~\etal~\cite{bidirectional-similarity} proposed bidirectional similarity scheme to capture better and summarize non-stationary visual data. However, these methods are computationally expensive due to calculating the similarity scores of each output-target pair. To relieve this problem, PatchMatch~\cite{patch-match} is proposed, which speeds it up by designing a faster similar patch searching algorithm. Ding~\etal~\cite{nonlocal-texture-matching} proposed a Gaussian-weighted nonlocal texture similarity measure to obtain multiple candidate patches for each target patch.

Recently, deep learning and GAN-based algorithms have been a remarkable paradigm for image inpainting. Context Encoders (CE)~\cite{CE} embed the $128 \times 128$ image with a $64 \times 64$ center hole as a low dimensional feature vector and then decode it to a $64 \times 64$ image. Iizuka~\etal~\cite{globally-and-locally} proposed a high-performance completion network with both global and local discriminators that is critical in obtaining semantically and locally consistent image inpainting results. Also, the authors employ the dilated convolution layers to increase receptive fields of the output neurons. Yang~\etal~\cite{high-resolution-multi-scale} use intermediate features extracted by pre-trained VGG network~\cite{VGG19} to find hole's most similar patch outside the hole. This approach performs multi-scale neural patch synthesis in a coarse-to-fine manner, which noticeably takes a long time to fill a large image during the inference stage. For face completion, Li~\etal~\cite{GFC} trained a deep generative model with a combination of reconstruction loss, global and local adversarial losses, and a semantic parsing loss specialized for face images. Contextual Attention (CA)~\cite{contextual-attention} adopted two-stage network architecture where the first step produces a crude result, and the second refinement network using attention mechanism takes the coarse prediction as inputs and improves fine details. Liu~\etal~\cite{partial-convolutions} introduced partial convolution that employs computational operations only on valid pixels and presented an auto-update binary mask to determinate whether the current pixels are valid. Substituting convolutional layers with partial convolutions can help a UNet-like architecture~\cite{img-to-img} achieve the state-of-the-art inpainting results. Yan~\etal~\cite{Shift-Net} introduced a special shift-connection to the U-Net architecture for enhancing the sharp structures and fine-detailed textures in the filled holes. This method was mainly developed on building and natural landscape images. Similar to~\cite{high-resolution-multi-scale,contextual-attention}, Song~\etal~\cite{contextual-based} decoupled the completion process into two stages: coarse inference and fine textures translation. Nazeri~\etal~\cite{EdgeConnect} also proposed a two-stage network that comprises of an edge generator and an image completion network. Similar to this method, Li~\etal~\cite{PRVS} progressively incorporated edge information into the feature to output more structured image. Xiong~\etal~\cite{Foreground} inferred the contours of the objects in the image, then used the completed contours as a guidance to complete the image. Different from frequently-used two-stage processing~\cite{StructureFlow}, Sagong~\etal~\cite{PEPSI} proposed parallel path for semantic inpainting to reduce the computational costs.

\section{Proposed Method}\label{sec:proposed-method}
\begin{figure}[htpb]
	\begin{center}
		\includegraphics[width=0.42\textwidth, height=0.25\textheight]{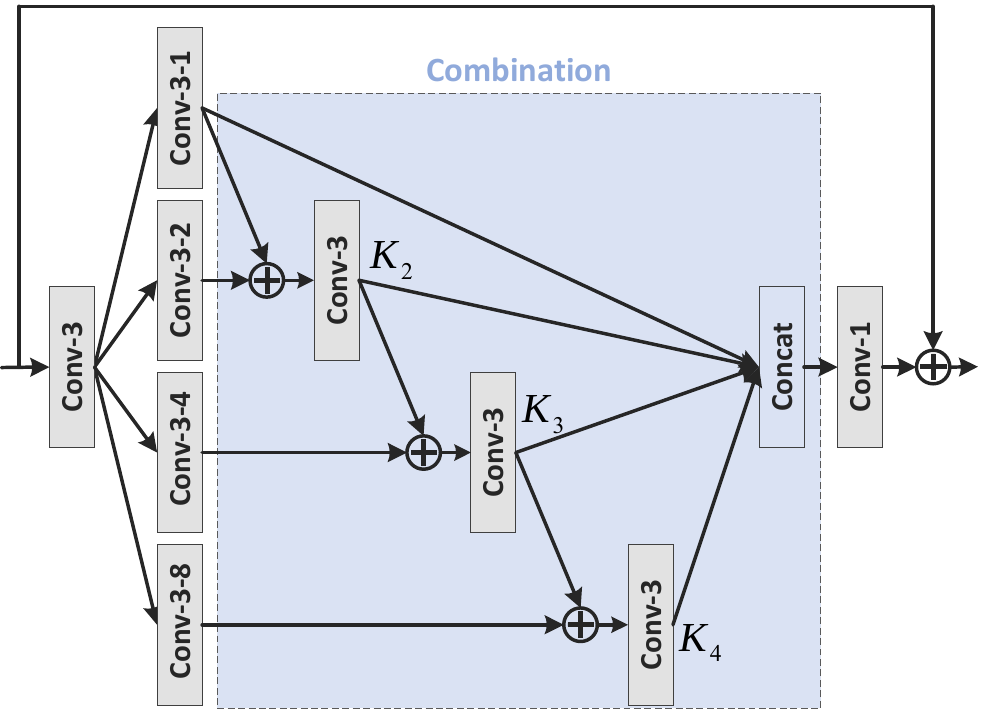}
	\end{center}
	\caption{The architecture of the proposed dense multi-scale fusion block (DMFB). Here, ``Conv-3-8'' indicates $3 \times 3$ convolution layer with the dilation rate of $8$ and $ \oplus $ is element-wise summation. Instance normalization (IN) and ReLU activation layers followed by the first convolution, second column convolutions and concatenation layer are omitted for brevity. The last convolutional layer only connects an IN layer. The number of output channels for each convolution is set to $64$ except for the last $1 \times 1$ convolution (256 channels) in DMFB.} 
	\label{fig:DMFB}
\end{figure}

\begin{figure*}[htpb]
	\begin{center}
		\includegraphics[width=\textwidth]{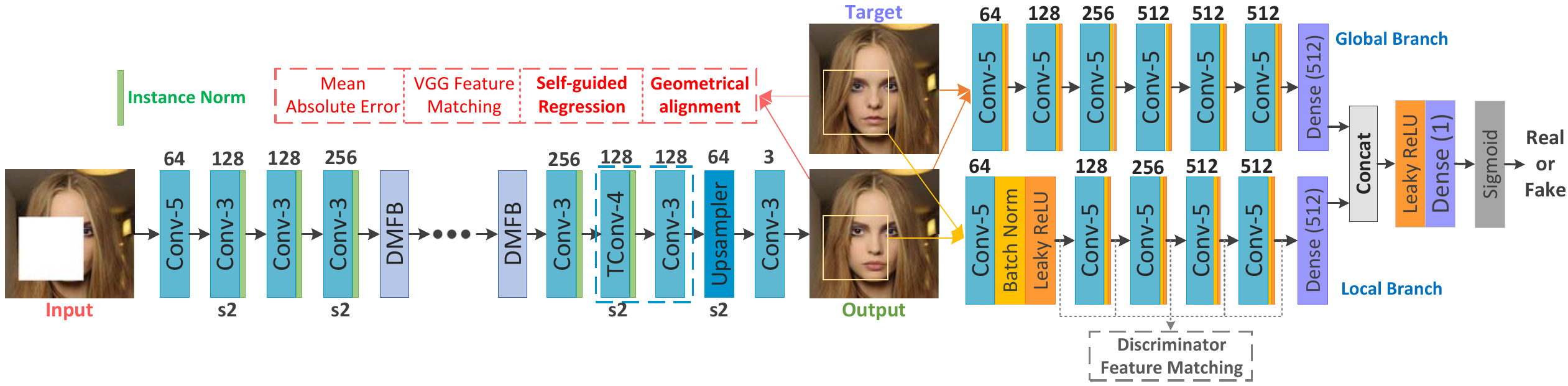}
	\end{center}
	\caption{The framework of our method. The activation layer followed by each ``convolution + norm'' or convolution layer in the generator is omitted for conciseness. The activation function adopts ReLU except for the last convolution (Tanh) in the generator. \textcolor{blue}{Blue} dotted box indicates our upsampler module (TConv-4 is $4 \times 4$ transposed convolution) and  ``$s2$'' denotes the stride of 2.}
	\label{fig:framework}
\end{figure*}

Our proposed inpainting system is trained in an end-to-end way. Given an input image with hole ${ \mathbf{I}_{in}}$, its corresponding binary mask $\mathbf{M}$ (value $0$ for known pixels and 
$1$ denotes unknown ones), the output ${\mathbf{I}_{out}}$ predicted by the network, and the ground-truth image ${\mathbf{I}_{gt}}$. We take the input image and mask as inputs,~\ie, $[{\mathbf{I}_{in}},\mathbf{M}]$. We now elaborate on our network as follows.

\subsection{Network structure}
As depicted in Figure~\ref{fig:framework}, our framework consists of a generator, and a discriminator with two branches. The generator produces plausible painted results, and the discriminator conducts adversarial training. 

For image inpainting task, the size of the receptive fields should be sufficiently large. The dilated convolution is popularly adopted in the previous works~\cite{globally-and-locally,contextual-attention} to accomplish this purpose. This way increases the area that can use as input without increasing the number of learnable weights. However, the kernel of dilated convolution is sparse, which skips many pixels during applying to compute. Large convolution kernel (\eg $7 \times 7$) is applied in~\cite{GMCNN} to implement this intention. However, this solution introduces heavy model parameters. To enlarge the receptive fields and ensure dense convolution kernels simultaneously, we propose our dense multi-scale fusion block (DMFB, see in Figure~\ref{fig:DMFB}) inspired by~\cite{Hui-PPON-2019}. Specifically, the first convolution on the left in DMFB reduces the channels of input features to $64$ for decreasing the parameters, and then these processed features are sent to four branches to extract multi-scale features, denoted as ${\mathbf{x}_i}$ ($i = 1,2,3,4$), by using dilated convolutions with different dilation factors. Except for ${\mathbf{x}_1}$, each ${\mathbf{x}_i}$ has a corresponding $3 \times 3$ convolution, denoted by ${K_i}\left(  \cdot  \right)$. Through a cumulative addition fashion, we can get dense multi-scale features from the combination of various sparse multi-scale features. We denote by ${\mathbf{y}_i}$ the output of ${K_i}\left(  \cdot  \right)$. The combination part can be formulated as
\begin{equation}
{\mathbf{y}_i} = \begin{cases}
{\mathbf{x}_i}, & i=1; \\
{K_i}\left( {{\mathbf{x}_{i - 1}} + {\mathbf{x}_i}} \right), & i=2; \\
{K_i}\left( {{\mathbf{y}_{i - 1}} + {\mathbf{x}_i}} \right), & 2 < i \le 4.
\end{cases}
\end{equation}
The following step is the fusion of concatenated features simply using a $1 \times 1$ convolution. In a word, this basic block especially enhances the general dilated convolution and has fewer parameters than large kernels.

Different from previous generative inpainting networks~\cite{contextual-attention,GMCNN} that apply WGAN-GP~\cite{WGAN-GP} for adversarial training, we propose to use RaGAN~\cite{RaGAN} to pursue more photo-realistic generated images~\cite{ESRGAN}. This discriminator also considers the consistency of global and local images.

\subsection{Loss functions}

\begin{figure}[htpb]
	\centering
	\includegraphics[width=0.48\textwidth]{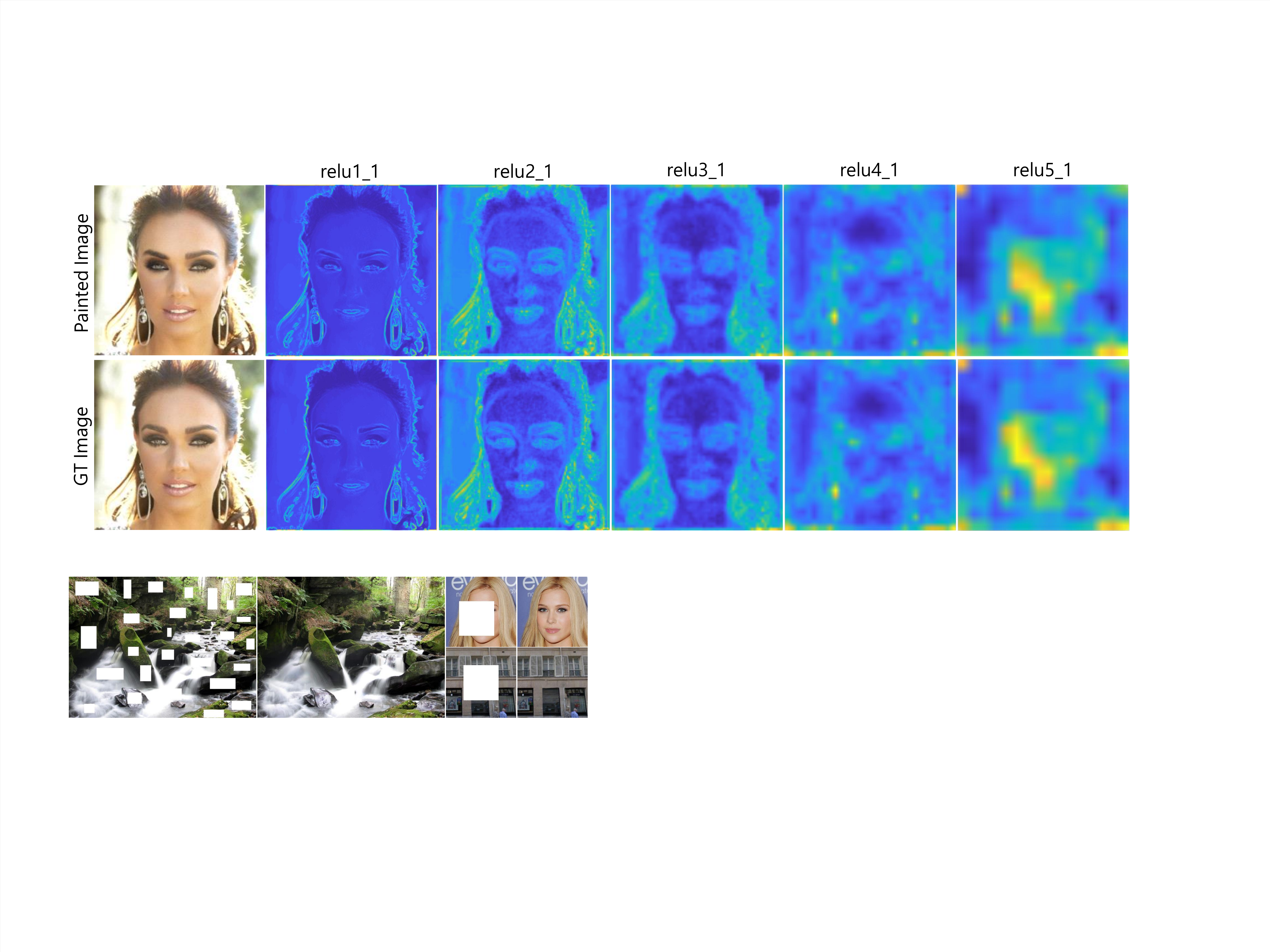}
	\caption{Visualization of average VGG feature maps.}
	\label{fig:avg-features}
\end{figure}

\subsubsection{Self-guided regression loss}
Here, we address the semantic structure preservation issue. We scheme to take self-guided regression constraint to correct the image semantic level estimation. Briefly, we compute the discrepancy map between generated contents and corresponding ground truth to navigate the similarity measure of the feature map hierarchy from the pre-trained VGG19~\cite{VGG19} network. At first, we investigate the characteristic of VGG feature maps. Given an input image ${\mathbf{I}_A}$, it is first fed forward through the VGG19 to yield a five-level feature map pyramid, where their spatial resolution reduces low progressively. Specifically, the $l$-th ($l = 1,2,3,4,5$) level is set to the feature tensor produced by relu$l$\_1 layer of VGG19. These feature tensors are denoted by $F_A^l$. We give an illustration of average feature maps $F_{A\_avg}^l = \frac{1}{M}\sum\limits_{m = 1}^M {F_{A\_m}^l}$ in Figure~\ref{fig:avg-features}, which suggests that the deeper layers of a pre-trained network represent higher-level semantic information, while lower-level features more focus on textural or structural details, such as edges, corners, and other simple conjunctions. In this paper, we would intend to improve the detail fidelity of the completed image, especially for building and face images.

\begin{figure}[htpb]
	\centering
	\includegraphics[width=0.48\textwidth]{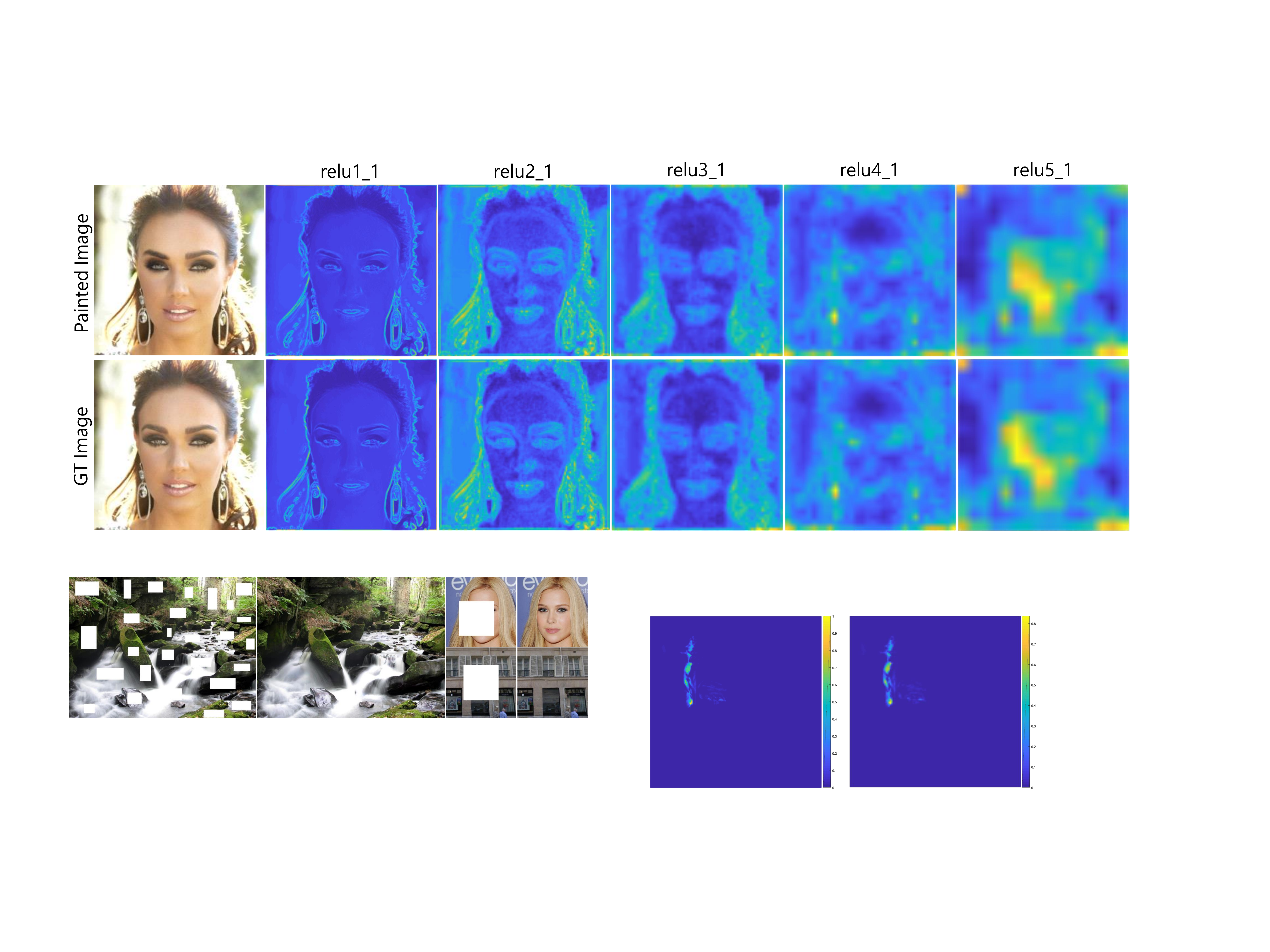}
	\caption{Visualization of guidance maps. (Left) Guidance map $\mathbf{M}_{guidance}^1$ for ``relu1\_1'' layer. (Right) Guidance map $\mathbf{M}_{guidance}^2$ for ``relu2\_1'' layer. These are corresponding to Figure~\ref{fig:avg-features}.}
	\label{fig:guidance-map}
\end{figure}

To this end, through the error map between the output image produced by the generator and ground truth, we get the guidance map to distinguish between areas of challenging and manageable. Therefore, we propose to use the following equation to gain the average error map:
\begin{equation}\label{eq:error-map}
{\mathbf{M}_{error}} = \frac{1}{3}\sum\limits_{c \in {\mathcal C}} {{{\left( {{\mathbf{I}_{out,c}} - {\mathbf{I}_{gt,c}}} \right)}^2 }},
\end{equation}

where ${\mathcal C}$ are the three color channels, $\mathbf{I}_{out,c}$ denotes $c$-th channel of the output image. Then, the normalized guidance mask can be calculated by
\begin{equation}\label{eq:normalized-guidance-mask}
{\mathbf{M}_{guidance,p}} = \frac{{{\mathbf{M}_{error,p}} - \min \left( {{\mathbf{M}_{error}}} \right)}}{{\max \left( {{\mathbf{M}_{error}}} \right) - \min \left( {{\mathbf{M}_{error}}} \right)}},
\end{equation}
where ${{\mathbf{M}_{error,p}}}$ is the error map value at position $p$. Note that our guidance mask with continuous values between $0$ and $1$, which is soft instead of binary. $\mathbf{M}_{guidence}^l$ corresponds $l$-th level feature maps and it can be expressed by
\begin{equation}\label{eq:l-th-guidance-map}
\mathbf{M}_{guidance}^{l + 1} = AP\left( {\mathbf{M}_{guidance}^l} \right),
\end{equation}
where $AP$ denotes \emph{average pooling} with kernel size of $2$ and stride of $2$. Here, $\mathbf{M}_{guidance}^1 = {\mathbf{M}_{guidance}}$ (Equation~\ref{eq:normalized-guidance-mask}). In this way, the value range of $\mathbf{M}_{guidance}^l$ is still between $0$ and $1$. In view of the fact that lower-level feature map contains more detailed information, we choose feature tensors from ``relu1\_1'' and ``relu2\_1'' layers to describe image semantic structures. Thus, our self-guided regression loss is defined as
\begin{small}
	\begin{equation}\label{eq:self-guided-loss}
	{{\mathcal L}_{self - guided}} = \sum\limits_{l = 1}^2 {{w^l}\frac{{{{\left\| {\mathbf{M}_{guidance}^l \odot \left( {\Psi _{{\mathbf{I}_{gt}}}^l - \Psi _{{\mathbf{I}_{output}}}^l} \right)} \right\|}_1}}}{{{N_{\Psi _{{\mathbf{I}_{gt}}}^l}}}}} ,
	\end{equation}
\end{small}
where $\Psi _{{\mathbf{I}_ * }}^{l}$ is the activation map of the relu$l$\_1 layer given original input ${\mathbf{I}_ * }$, ${{N_{\Psi _{{\mathbf{I}_{gt}}}^l}}}$ is the number of elements in ${\Psi _{{\mathbf{I}_{gt}}}^{l}}$, $ \odot $ is the element-wise product operator, and ${w^l} = \frac{{1e3}}{{{{\left( {{C_{\Psi _{{\mathbf{I}_{gt}}}^{l}}}} \right)}^2}}}$ followed by~\cite{non-stationary}. Here, $C$ is the channel size of feature map $\Psi _{{\mathbf{I}_{gt}}}^{l}$. 

An obvious benefit for this regularization is to suppress regions with higher uncertainty (as shown in Figure~\ref{fig:guidance-map}). ${\mathbf{M}_{guidance}}$ can be viewed as a spatial attention map, which preferably optimizes areas that are difficult to handle. Our self-guided regression loss is performed lower-level semantic space instead of pixel space. The merit of this way would appear in the perceptual image synthesis with pleasant structural information.

\subsubsection{Geometrical alignment constraint}
In the typical solutions, the metric evaluation in higher-level feature space is only achieved using pixel-based loss,~\eg, L1 or L2. It doesn't take the alignment of each high-level feature map semantic hub into account. To better measure the distance between high-level features belong to prediction and ground-truth, we impose the geometrical alignment constraint on the response maps of ``relu4\_1'' layer. This term can help the generator create a plausible image that aligned with the target image in position. Specifically, this term encourages the output feature center to be spatially close to the target feature center. The geometrical center for the \textit{k}-th feature map along axis $u$ is calculated as
\begin{equation}\label{geometrical-center}
c_u^k = \sum\limits_{u,v} {u \cdot \left( {\mathbf{R}\left( {k,u,v} \right)/\sum\limits_{u,v} {\mathbf{R}\left( {k,u,v} \right)} } \right)} ,
\end{equation}
where response maps $\mathbf{R} = \text{VGG}\left( {\mathbf{I};{\theta _\text{vgg}}} \right) \in {\mathbb{R}^{K \times H \times W}}$. ${{\mathbf{R}\left( {k,u,v} \right)} \mathord{\left/
		{\vphantom {{R\left( {k,u,v} \right)} {\sum\limits_{u,v} R }}} \right.
		\kern-\nulldelimiterspace} {\sum\limits_{u,v} \mathbf{R} }}\left( {k,u,v} \right)$ represents a spatial probability distribution function. $c_u^k$ denotes coordinate expectation along axis $u$. Then, we pass both the completed image ${\textbf{I}_{output}}$ and ground-truth image ${\mathbf{I}_{gt}}$ through the VGG network and obtain the corresponding response maps $\mathbf{R'}$ and $\mathbf{R}$. Given these response maps, we compute the centers $\left\langle {c{{_u^k}^\prime },c{{_v^k}^\prime }} \right\rangle $ and $\left\langle {c_u^k,c_v^k} \right\rangle $ using Equation~\ref{geometrical-center}. Then, we formulate the geometrical alignment constraint as
\begin{equation}\label{geometrical-alignment}
%{{\mathcal L}_{align}} = \sum\limits_k {\sum\limits_{u,v} {\left\| {\left\langle {c_u^{k'},c_v^{k'}} \right\rangle  - \left\langle {c_u^k,c_v^k} \right\rangle } \right\|_2^2} }.
{\mathcal{L}_{align}} = \sum\limits_k {\left\| {\left\langle {c{{_u^k}^\prime },c{{_v^k}^\prime }} \right\rangle  - \left\langle {c_u^k,c_v^k} \right\rangle } \right\|_2^2} .
\end{equation}

\subsubsection{Feature matching losses} 
The VGG feature matching loss ${{\mathcal L}_{fm\_vgg}}$ compares the activation maps in the intermediate layers of well-trained VGG19~\cite{VGG19} model, which can be written as
\begin{equation}\label{eq:vgg-fm}
{{\mathcal L}_{fm\_vgg}} = \sum\limits_{l = 1}^5 {{w^l}\frac{{{{\left\| {\Psi _{{\mathbf{I}_{gt}}}^l - \Psi _{{\mathbf{I}_{output}}}^l} \right\|}_1}}}{{{N_{\Psi _{{\mathbf{I}_{gt}}}^l}}}}} ,
\end{equation}
where ${N_{\Psi _{{\mathbf{I}_{gt}}}^l}}$ is the number of elements in $\Psi _{{\mathbf{I}_{gt}}}^l$. Inspired by~\cite{perceptual-adversarial}, we also introduce local branch in discriminator feature matching loss ${{\mathcal L}_{fm\_dis}}$, which is reasonable to assume that the output image are consistent with the ground-truth images under any measurements (\ie, any high-dimensional spaces). This feature matching loss is defined as
\begin{equation}\label{eq:dis-fm}
{{\mathcal L}_{fm\_dis}} = \sum\limits_{l = 1}^5 {{w^l}\frac{{{{\left\| {D_{local}^l\left( {{\mathbf{I}_{gt}}} \right) - D_{local}^l\left( {{\mathbf{I}_{output}}} \right)} \right\|}_1}}}{{{N_{D_{local}^l\left( {{\mathbf{I}_{gt}}} \right)}}}}} ,
\end{equation}
where $D_{local}^l\left( {{\mathbf{I}_ * }} \right)$ is the activation in the $l$-th selected layer of the discriminator given input ${\mathbf{I}_ * }$ (see in Figure~\ref{fig:framework}). Note that the hidden layers of the discriminator are trainable, which is slightly different from the well-trained VGG19 network trained on the ImageNet dataset. It can adaptively update based on specific training data. This complementary feature matching can dynamically extract features that may be not mined in VGG model. 

\subsubsection{Adversarial loss}
For improving the visual quality of inpainted results, we use relativistic average discriminator~\cite{RaGAN} as in ESRGAN~\cite{ESRGAN}, which is the recent state-of-the-art perceptual image super-resolution algorithm. For the generator, the adversarial loss is defined as
\begin{equation}\label{eq:advsarial-loss}
\begin{aligned}
{{\mathcal L}_{adv}} &=  - {\mathbb{E}_{{x_r}}}\left[ {\log \left( {1 - D_{Ra}\left( {{x_r},{x_f}} \right)} \right)} \right] \\&- {\mathbb{E}_{{x_f}}}\left[ {\log \left( {D_{Ra}\left( {{x_f},{x_r}} \right)} \right)} \right],
\end{aligned}
\end{equation}
where ${D_{Ra}}\left( {{x_r},{x_f}} \right) = sigmoid \left( {C\left( {{x_r}} \right) - {\mathbb{E}_{{x_f}}}\left[ {C\left( {{x_f}} \right)} \right]} \right)$ and $C\left(\cdot\right)$ indicates the discriminator network without the last \emph{sigmoid} function. Here, real/fake data pairs $\left( {{x_r},{x_f}} \right)$ are sampled from ground-truth and output images.

\subsubsection{Final objective}
With self-guided regression loss, geometrical alignment constraint, VGG feature matching loss, discriminator feature matching loss, adversarial loss, and mean absolute error (MAE) loss, our overall loss function is defined as
\begin{equation}\label{eq:total-loss}
\begin{aligned}
{{\mathcal L}_{total}} &= {{\mathcal L}_{mae}} + \lambda \left( {{{\mathcal L}_{self - guided}}  + {{\mathcal L}_{fm\_vgg}}} \right) \\&+ \eta {{\mathcal L}_{fm\_dis}} + \mu {{\mathcal L}_{adv}} + \gamma {{\mathcal L}_{align}},
\end{aligned}
\end{equation}
where $\lambda $, $\eta $, $\mu $, and $\gamma $ are used to balance the effects between the losses mentioned above.

\begin{figure}[htpb]
	\centering
	\begin{adjustbox}{valign=t}
		\begin{tabular}{cccc}
			\includegraphics[width=0.11\textwidth]{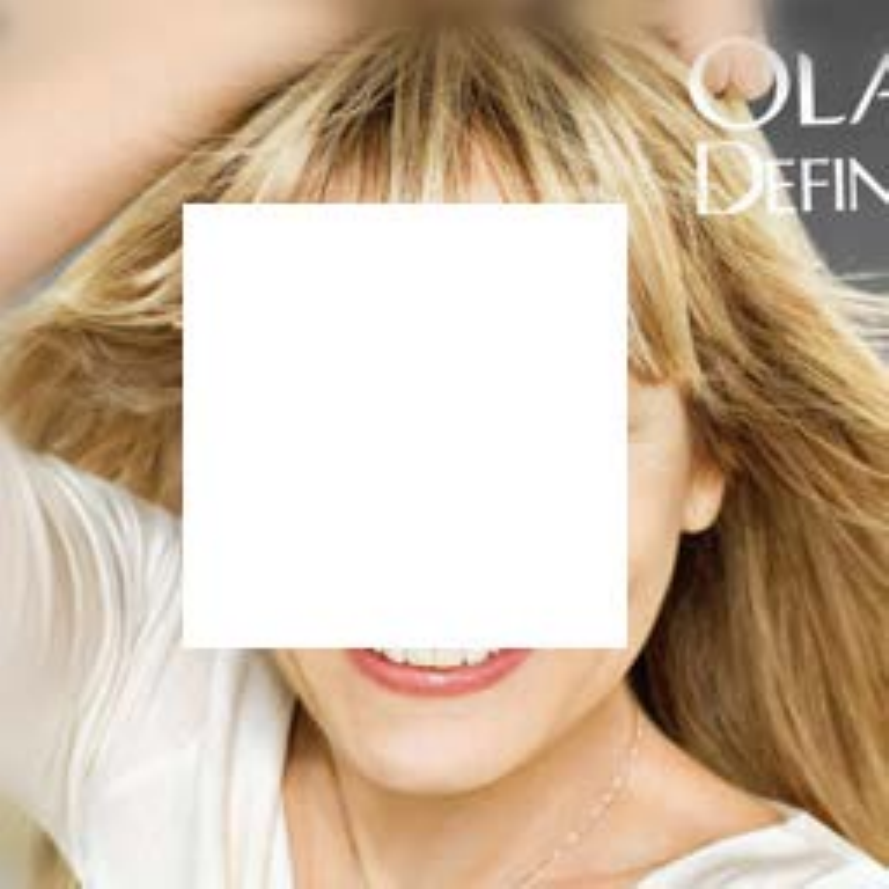} &
			\hspace{-4mm}
			\includegraphics[width=0.11\textwidth]{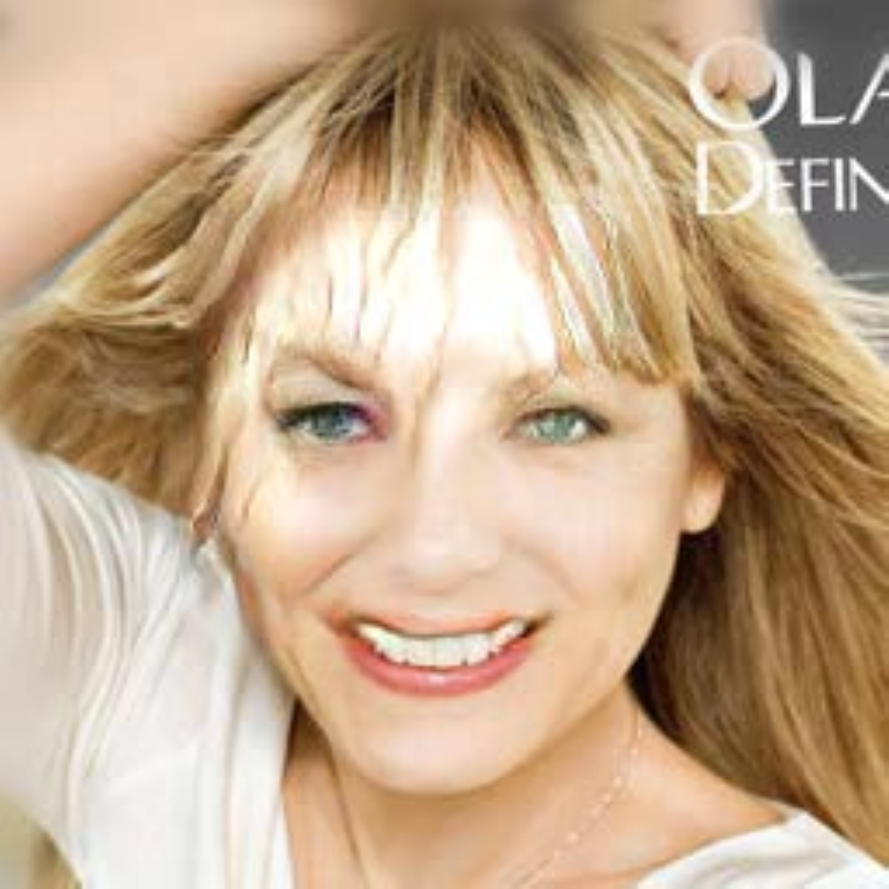} &
			\hspace{-4mm}
			\includegraphics[width=0.11\textwidth]{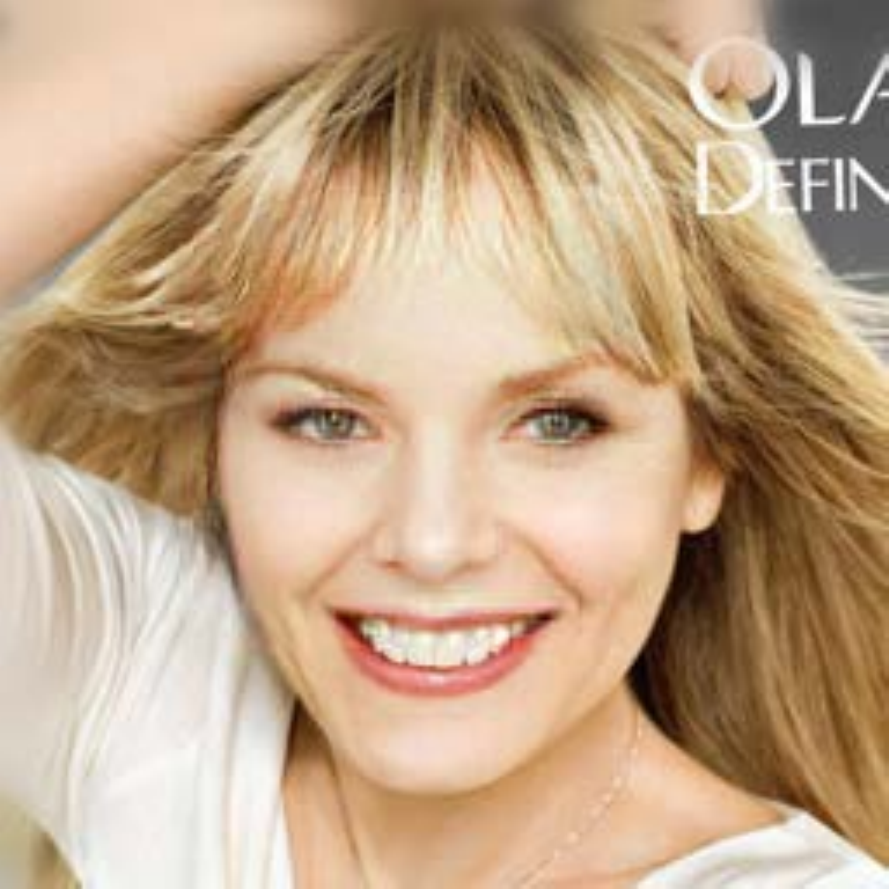} &
			\hspace{-4mm}
			\includegraphics[width=0.11\textwidth]{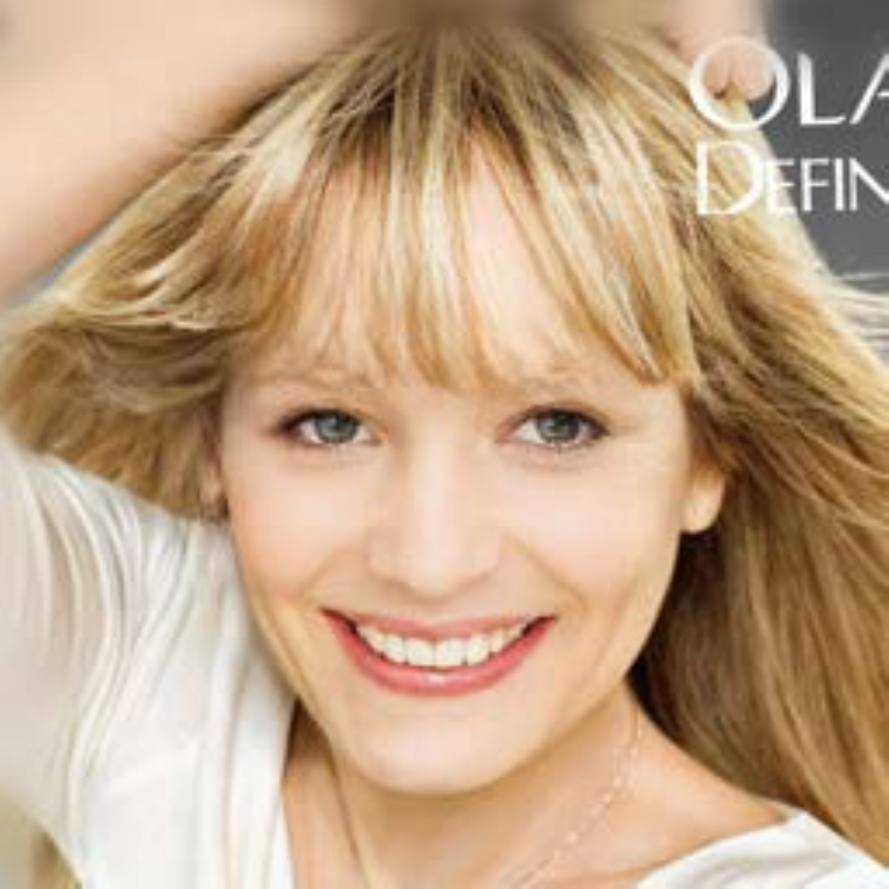} \\

			\includegraphics[width=0.11\textwidth]{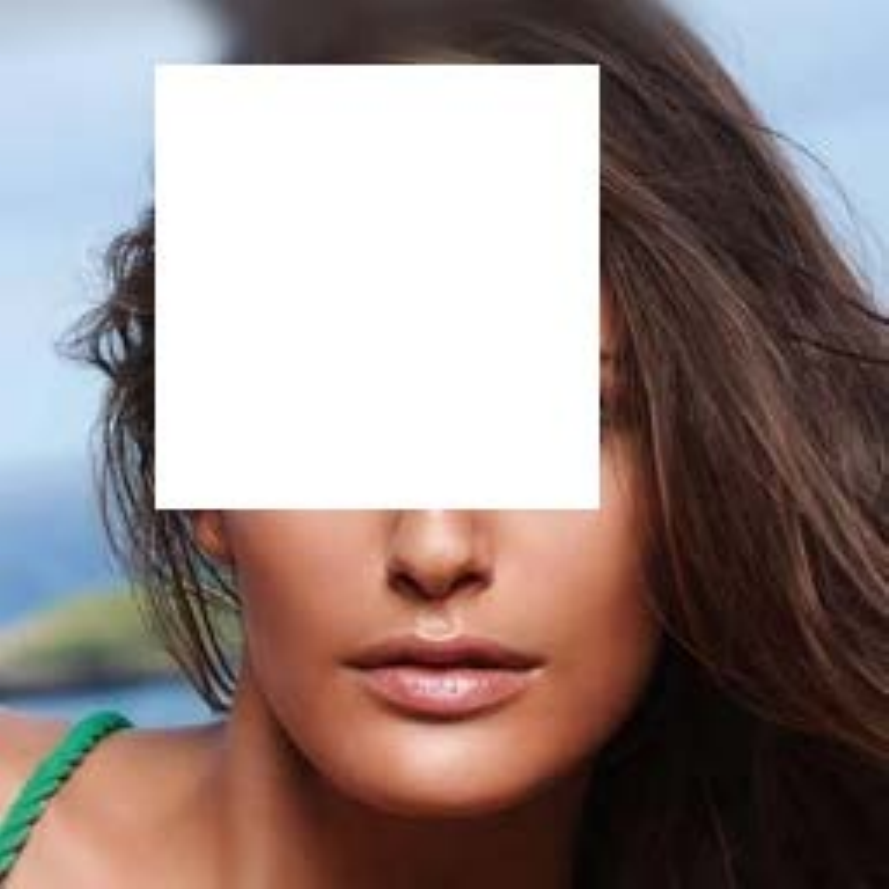} &
			\hspace{-4mm}
			\includegraphics[width=0.11\textwidth]{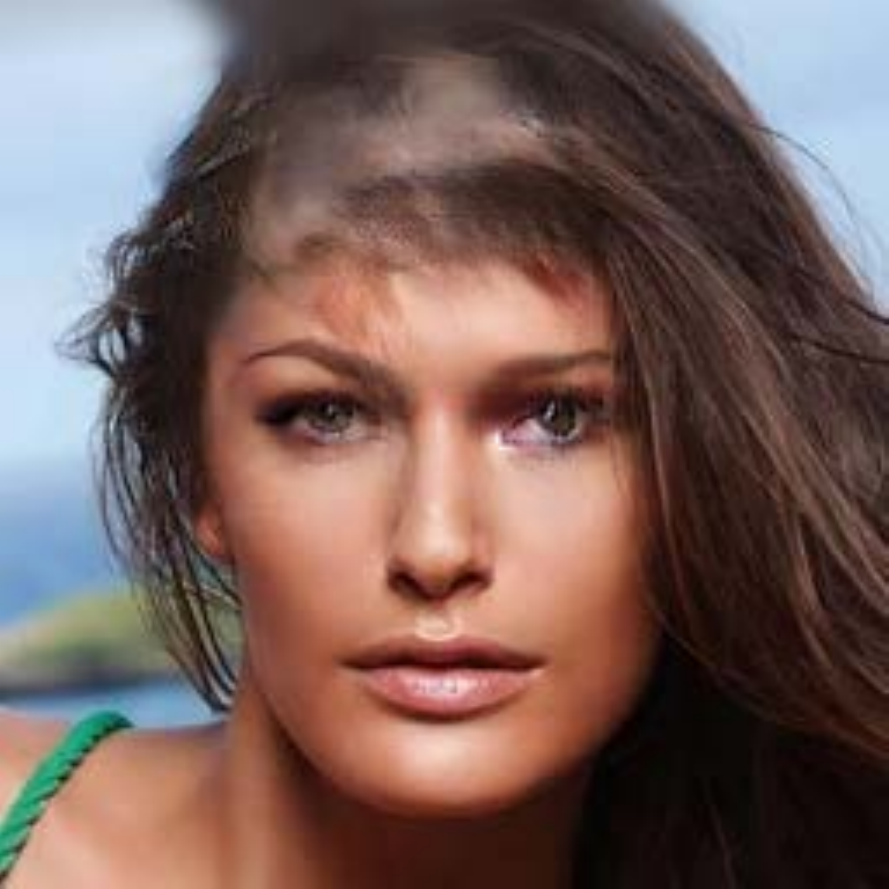} &
			\hspace{-4mm}
			\includegraphics[width=0.11\textwidth]{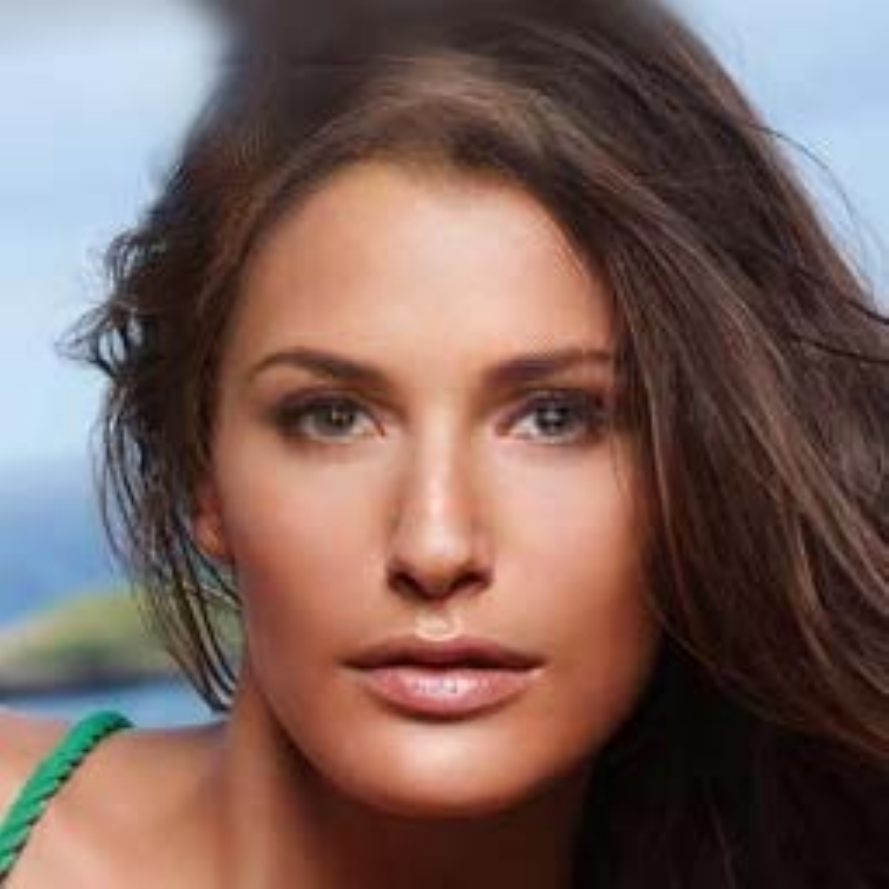} &
			\hspace{-4mm}
			\includegraphics[width=0.11\textwidth]{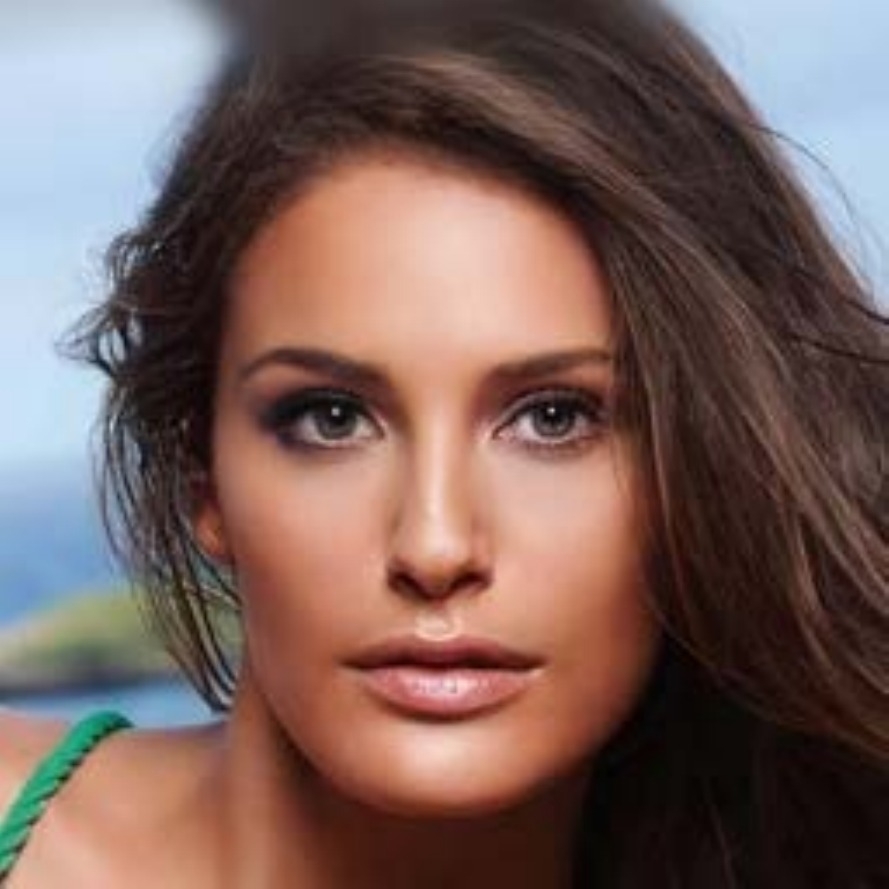} \\
			
			Input &\hspace{-4mm} CA~\cite{contextual-attention} &\hspace{-4mm} GMCNN~\cite{GMCNN} &\hspace{-4mm} DMFN (Ours) \\
		\end{tabular}
	\end{adjustbox}
	\caption{Visual comparisons on CelebA-HQ. \textit{Best viewed with zoom-in.}}
	\label{fig:celeba-hq}
\end{figure}

\section{Experiments}\label{sec:experiments}
We evaluate the proposed inpainting model on Paris Street View~\cite{CE}, Places2~\cite{places2}, CelebA-HQ~\cite{celeba-hq}, and a new challenging facial dataset FFHQ~\cite{ffhq}.

\subsection{Experimental settings}
For our experiments, we empirically set $\lambda = 25$, $\eta = 5$, $\mu = 0.003$ and $\gamma = 1$ in Equation~\ref{eq:total-loss}. The training procedure is optimized using Adam optimizer~\cite{Adam} with ${\beta _1} = 0.5$ and ${\beta _2} = 0.9$. We set the learning rate to $2e - 4$. The batch size is $16$. We apply PyTorch framework to implement our model and train them using NVIDIA TITAN Xp GPU (12GB memory). 

For training, given a raw image $\mathbf{I}_{gt}$, a binary image mask $\mathbf{M}$ (value $0$ for known pixels and $1$ denotes unknown ones) at a random position. In this way, the input image ${\mathbf{I}_{in}}$ is obtained from the raw image as ${\mathbf{I}_{in}} = {\mathbf{I}_{gt}} \odot \left( {\mathbf{1} - \mathbf{M}} \right)$. Our inpainting generator takes $\left[ {{\mathbf{I}_{in}},\mathbf{M}} \right]$ as input, and produces prediction ${\mathbf{I}_{pred}}$. The final output image is ${\mathbf{I}_{out}} = {\mathbf{I}_{in}} + {\mathbf{I}_{pred}} \odot \mathbf{M}$. All input and output are linearly scaled to $\left[ { - 1,1} \right]$. We train our network on the training set and evaluate it on the validation set (Places2, CelebA-HQ, and FFHQ) or testing set (Paris street view and CelebA). For training, we use images of resolution $256 \times 256$ with the largest hole size $128 \times 128$ as in~\cite{contextual-attention,GMCNN}. For Paris street view ($936 \times 537$), we randomly crop patches with resolution $537 \times 537$ and then scale down them to $256 \times 256$ for training. Similarly, for Places2 ($512 \times *$), $512 \times 512$ sub-images are cropped at a random location. These images are scaled down to $256 \times 256$ for our model. For CelebA-HQ and FFHQ face datasets ($1024 \times 1024$), images are directly scaled to $256 \time 256$. We use the irregular mask dataset provided by~\cite{partial-convolutions}. For irregular masks, the random regular regions are cropped and sent to the local discriminator. All results generated by our model are not post-processed.

\subsection{Qualitative comparisons}
\begin{figure*}[ht]
	\centering
	\begin{adjustbox}{valign=t}
		\begin{tabular}{cccccc}
			% 12
			\includegraphics[width=0.15\textwidth]{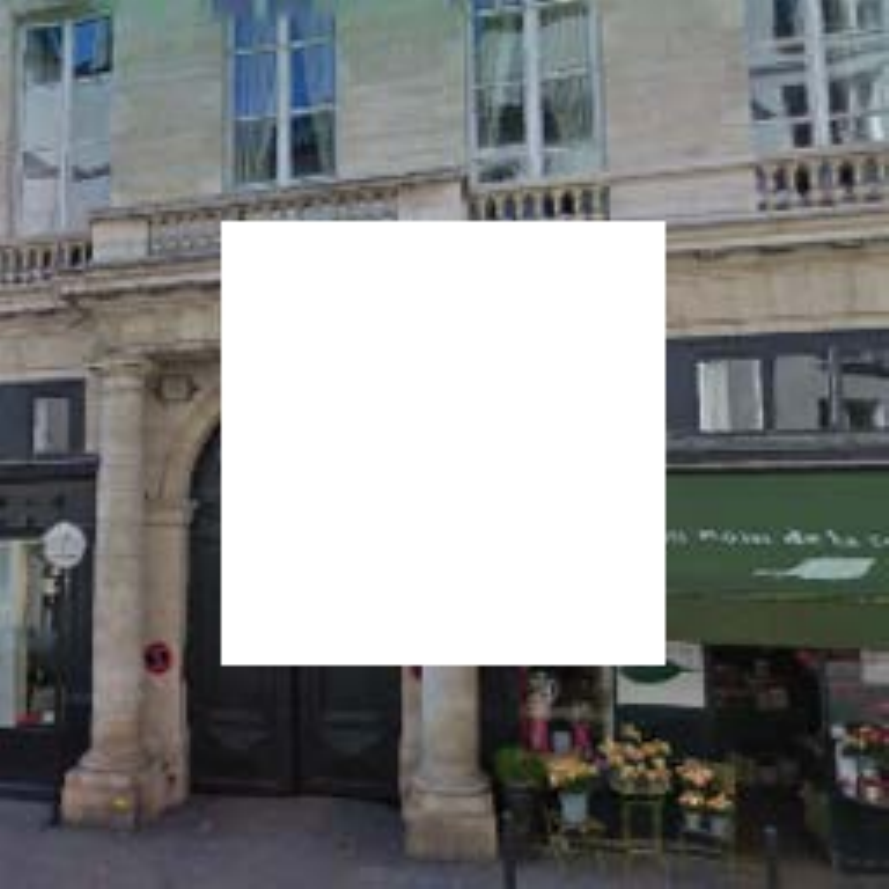} &
			\hspace{-3mm}
			\includegraphics[width=0.15\textwidth]{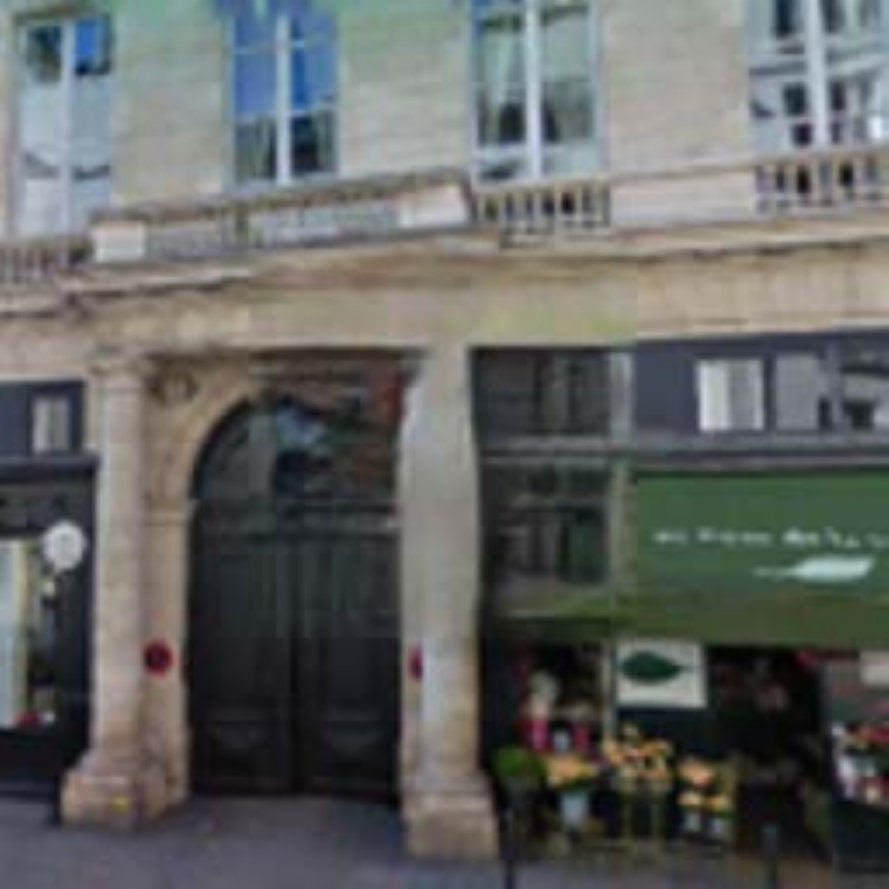} &
			\hspace{-3mm}
			\includegraphics[width=0.15\textwidth]{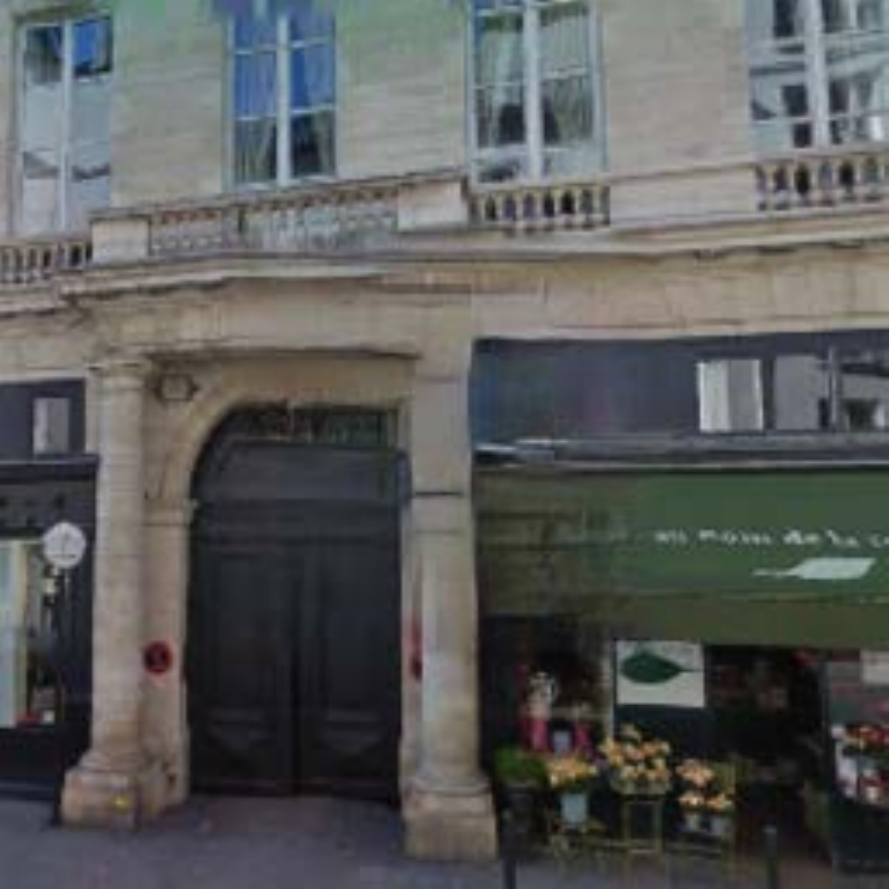} &
			\hspace{-3mm}
			\includegraphics[width=0.15\textwidth]{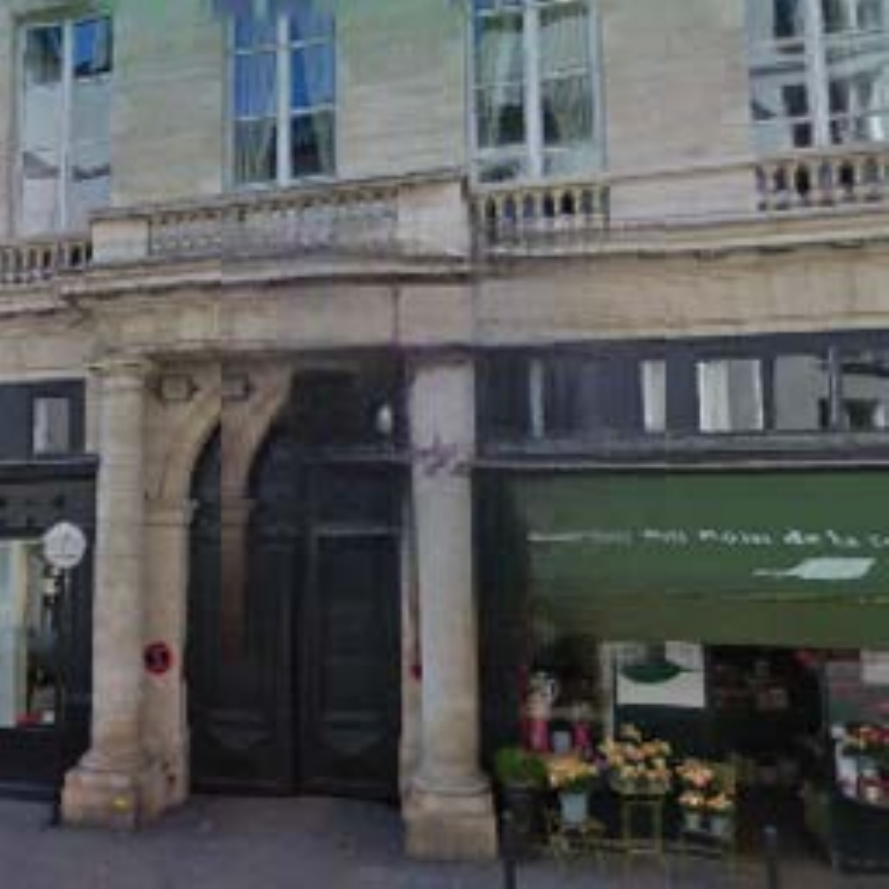} &
			\hspace{-3mm}
			\includegraphics[width=0.15\textwidth]{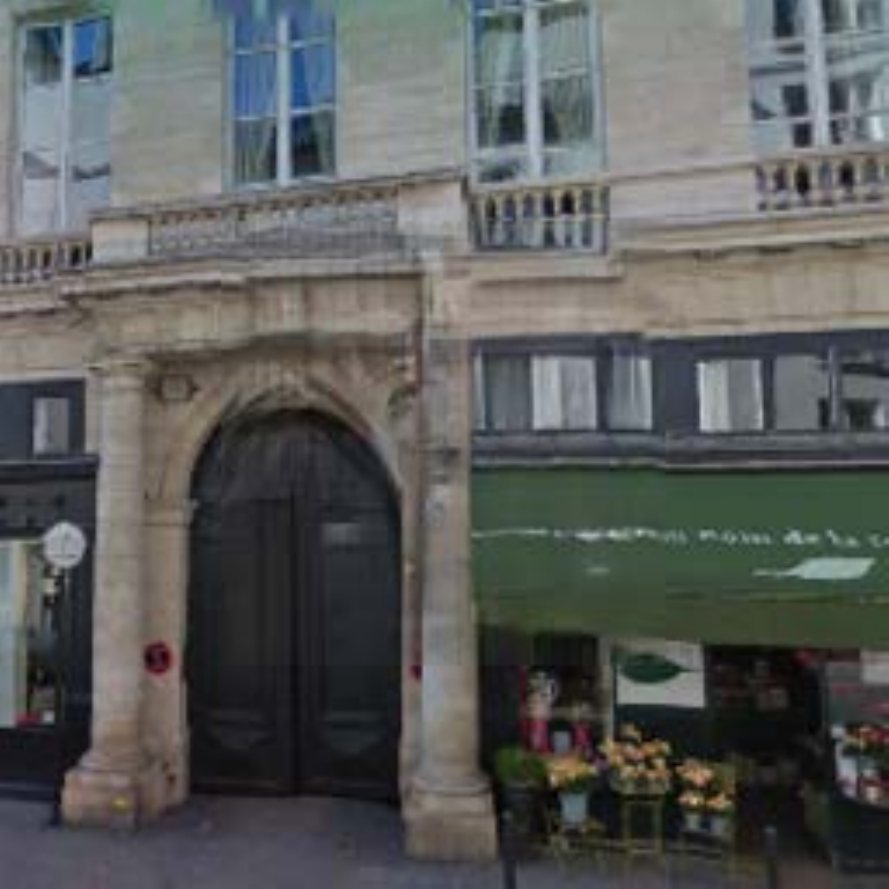} &
			\hspace{-3mm}
			\includegraphics[width=0.15\textwidth]{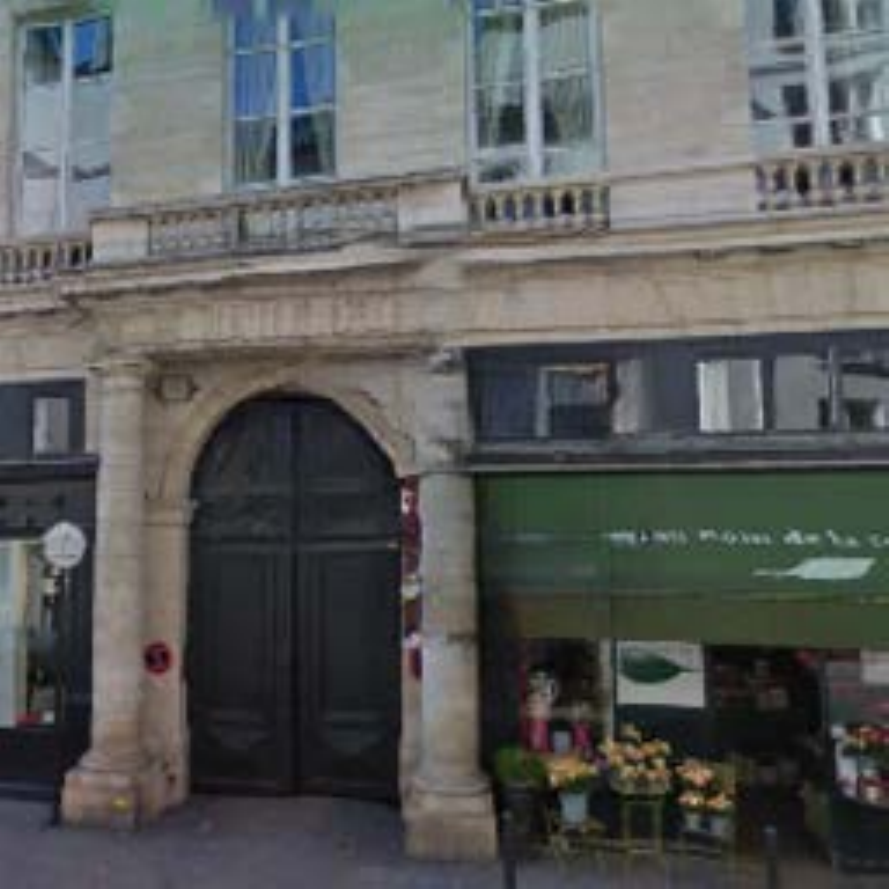} \\
			
			% 026
			\includegraphics[width=0.15\textwidth]{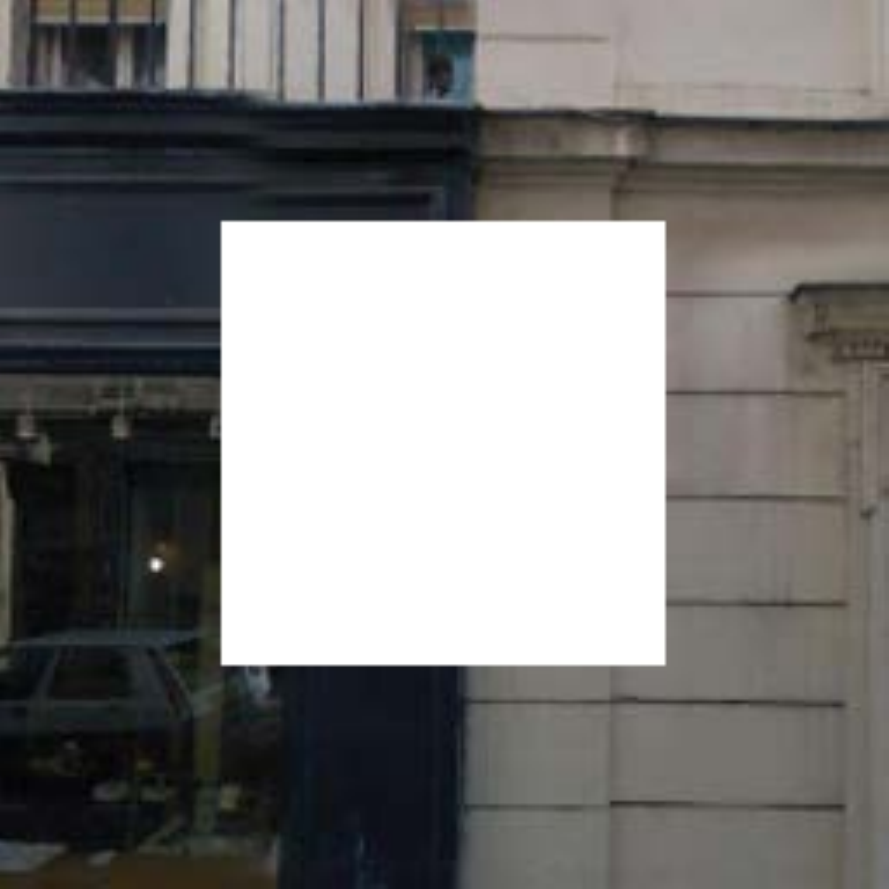} &
			\hspace{-3mm}
			\includegraphics[width=0.15\textwidth]{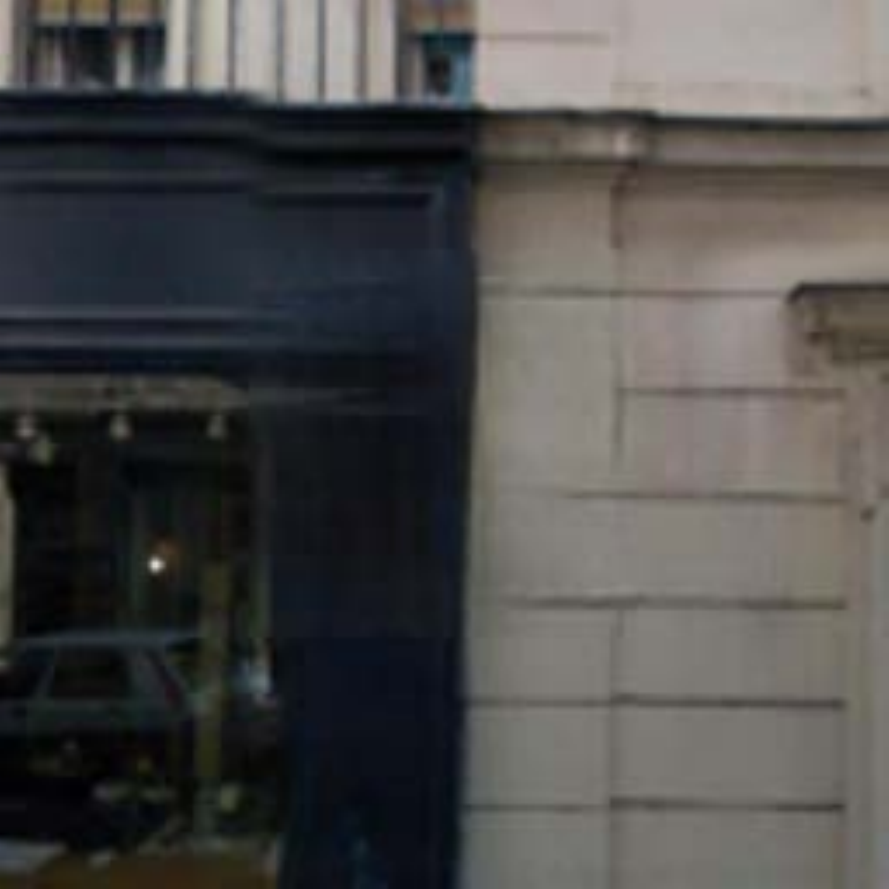} &
			\hspace{-3mm}
			\includegraphics[width=0.15\textwidth]{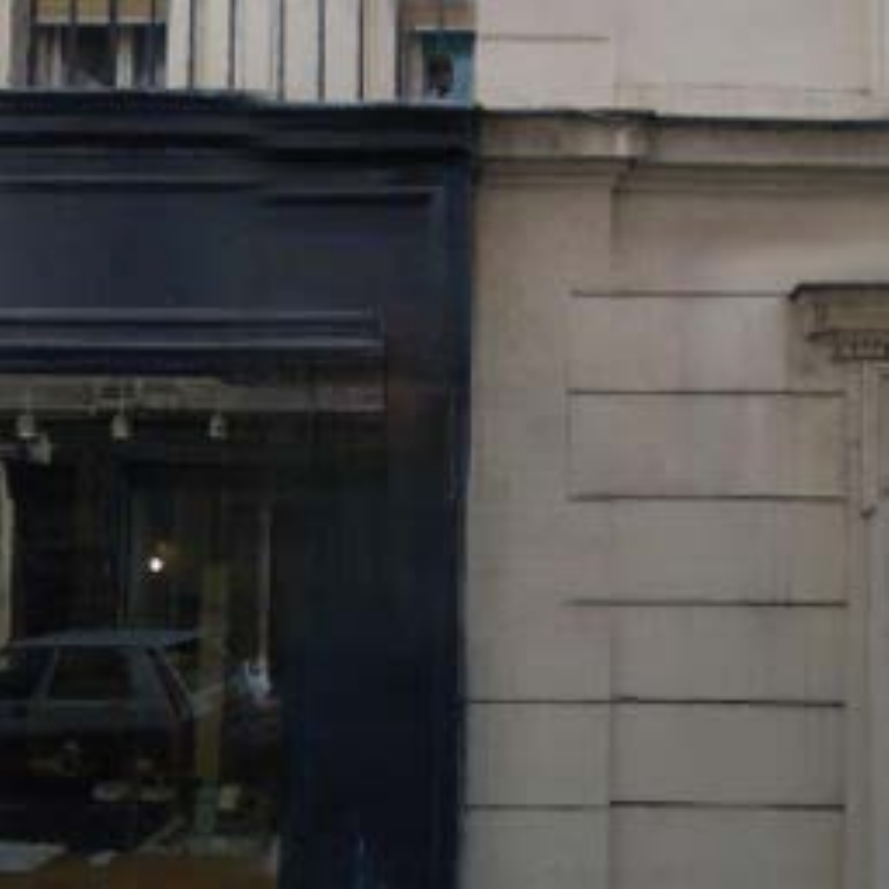} &
			\hspace{-3mm}
			\includegraphics[width=0.15\textwidth]{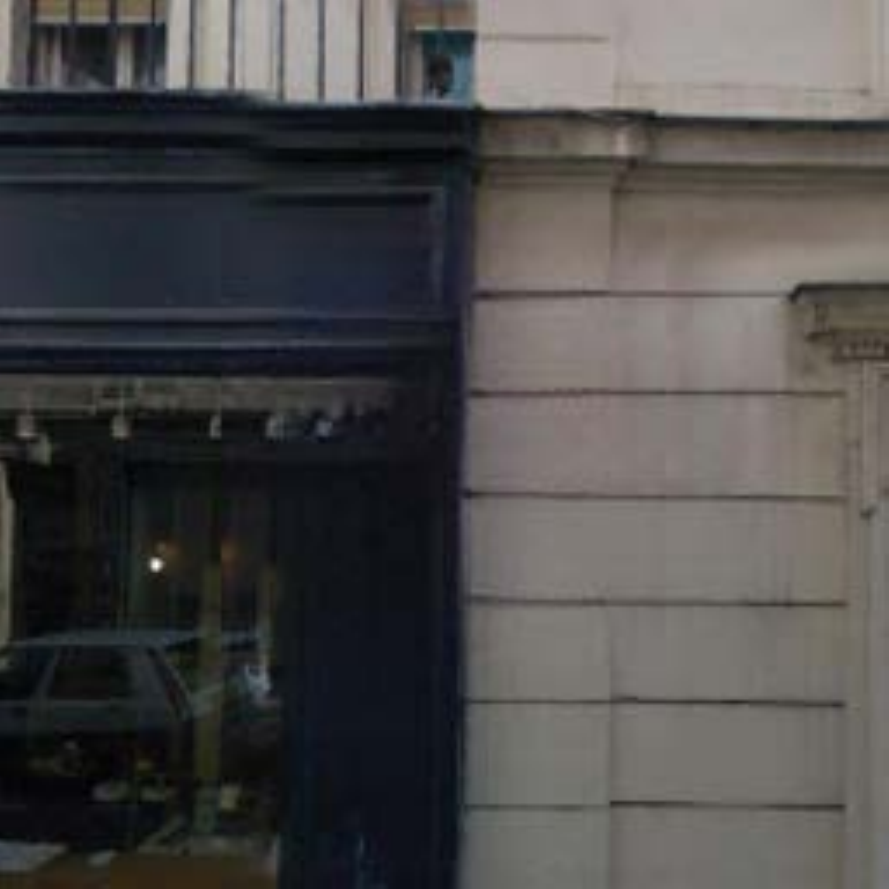} &
			\hspace{-3mm}
			\includegraphics[width=0.15\textwidth]{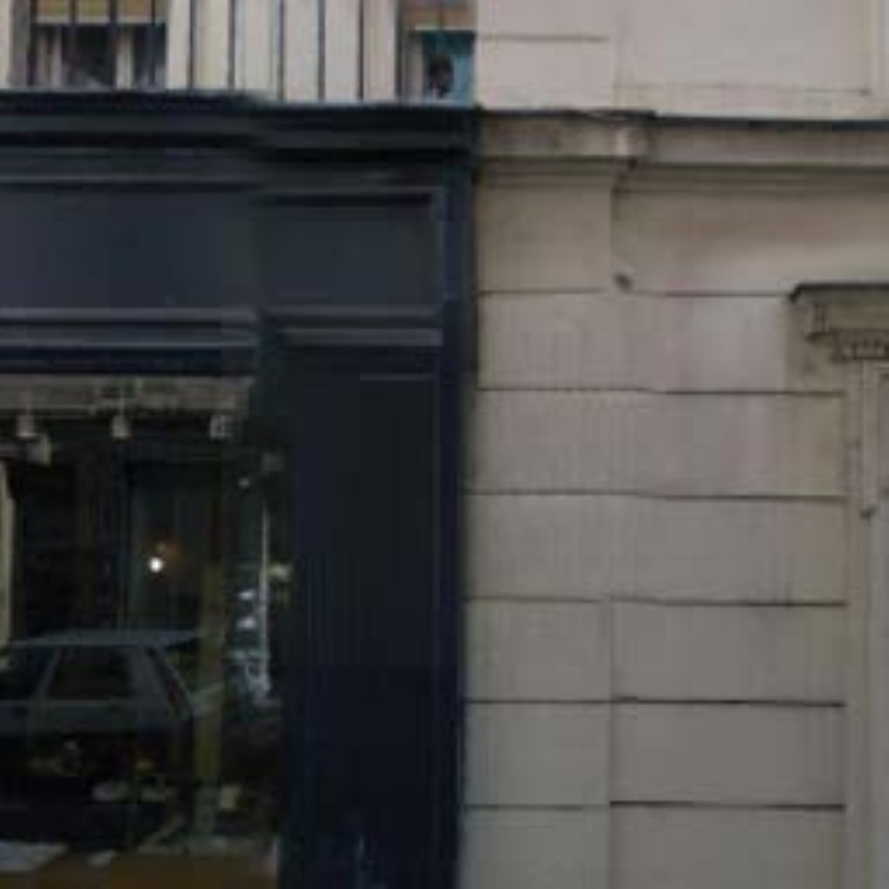} &
			\hspace{-3mm}
			\includegraphics[width=0.15\textwidth]{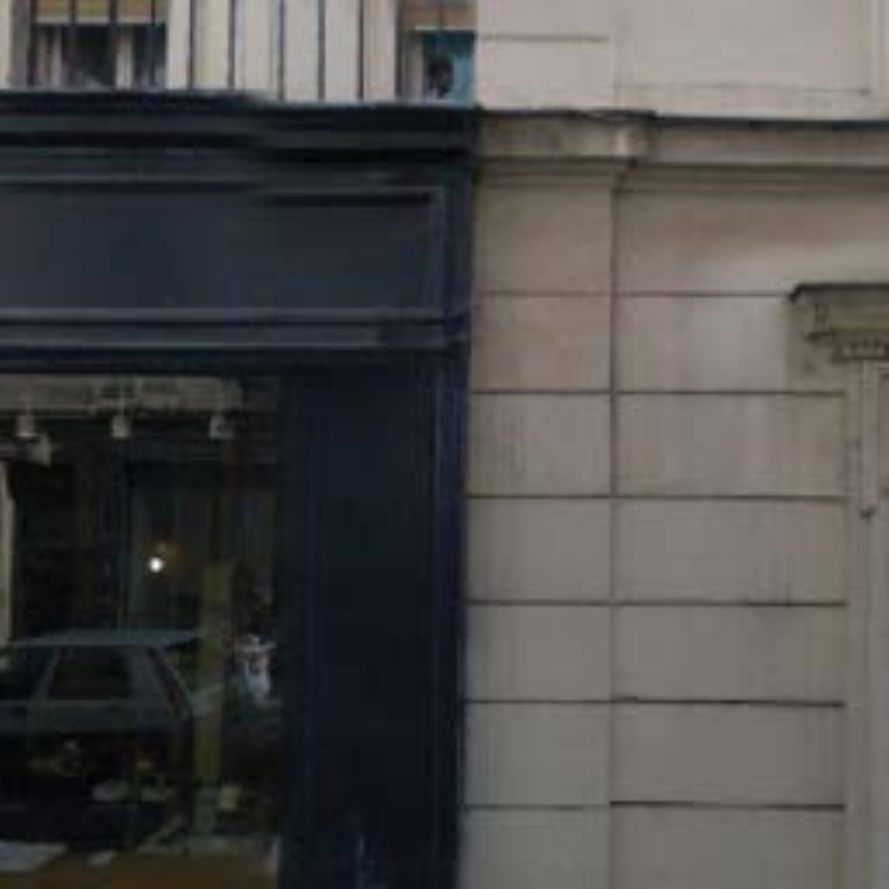} \\
			
			Input Image & \hspace{-3mm} CE~\cite{CE} & \hspace{-3mm} Shift-Net~\cite{Shift-Net} & \hspace{-3mm} GMCNN~\cite{GMCNN} & PICNet~\cite{PICNet} & \hspace{-3mm} DMFN (Ours) \\
		\end{tabular}
	\end{adjustbox}
	\caption{Visual comparisons on Paris street view.}
	\label{fig:paris-streetview}
\end{figure*}

\begin{figure}[ht]
	\centering
	\begin{adjustbox}{valign=t}
		\begin{tabular}{cccc}
			
			\includegraphics[width=0.12\textwidth]{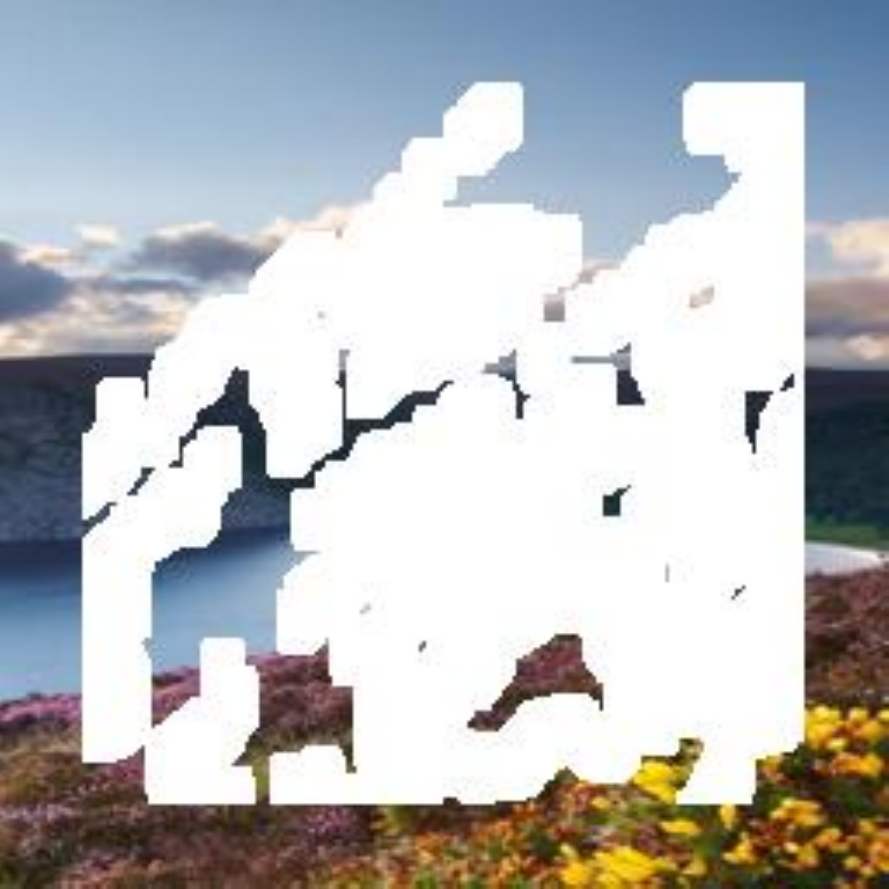} &
			\hspace{-4mm}
			\includegraphics[width=0.12\textwidth]{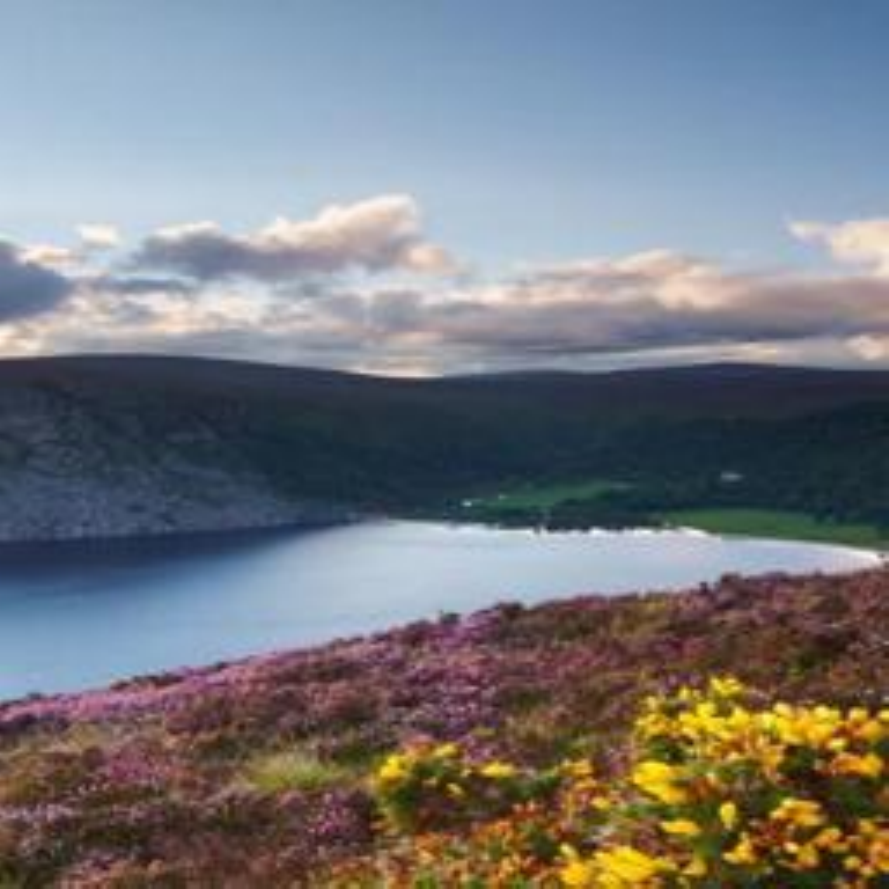} &
			\hspace{-4mm}
			\includegraphics[width=0.12\textwidth]{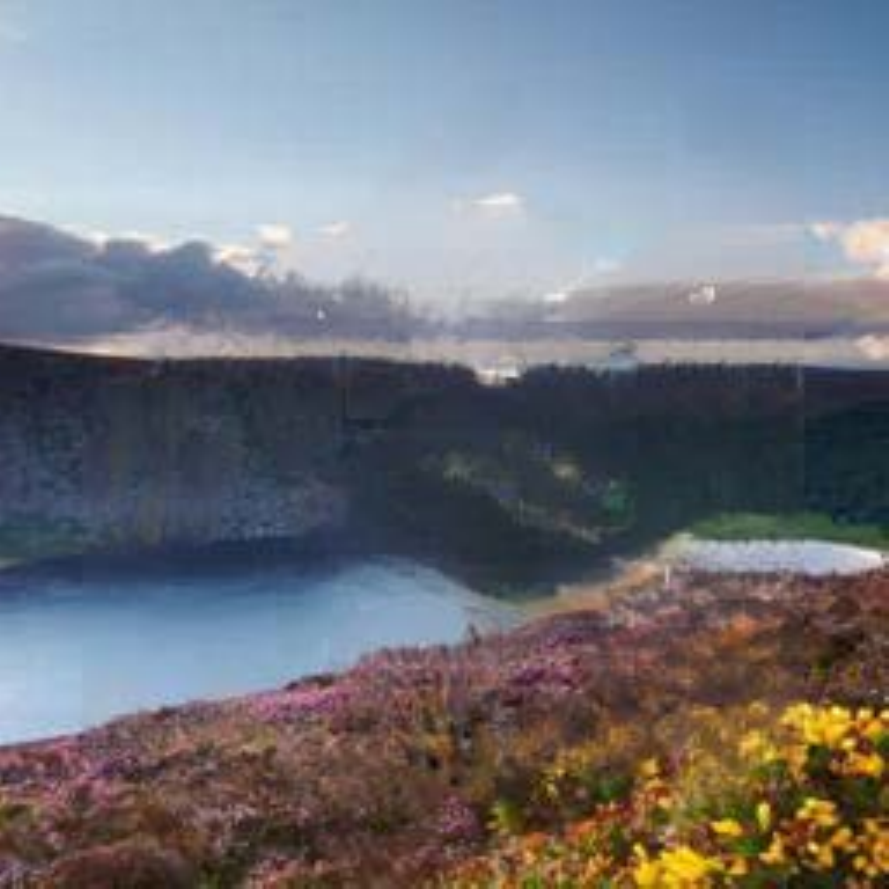} &
			\hspace{-4mm}
			\includegraphics[width=0.12\textwidth]{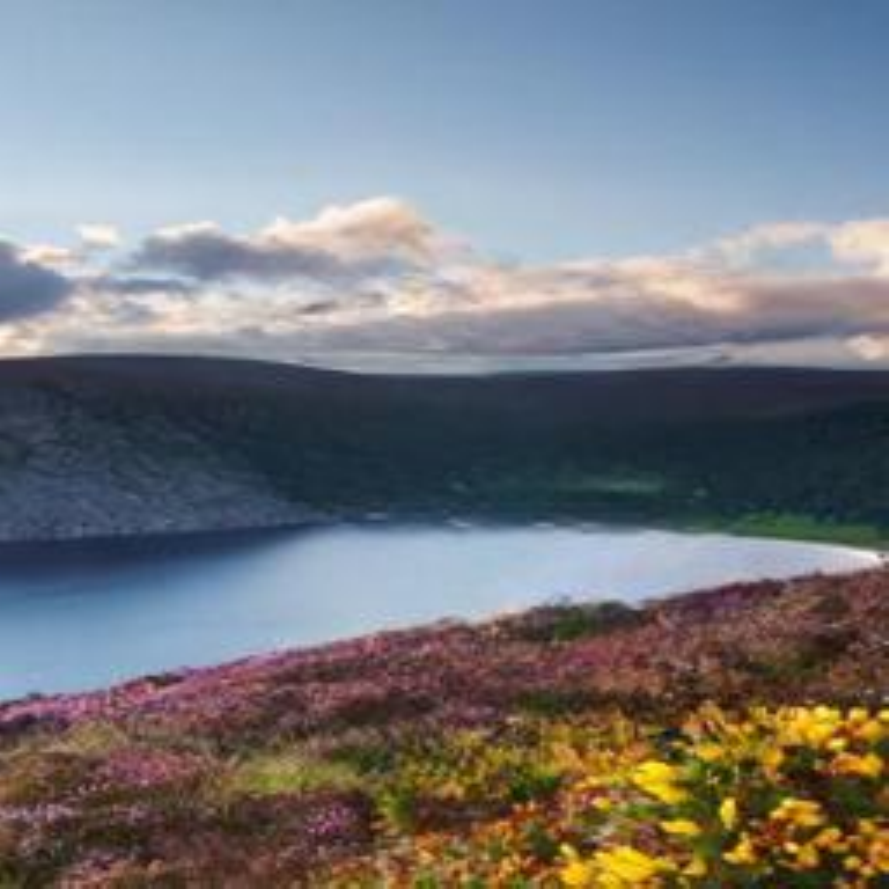}\\
			
			\includegraphics[width=0.12\textwidth]{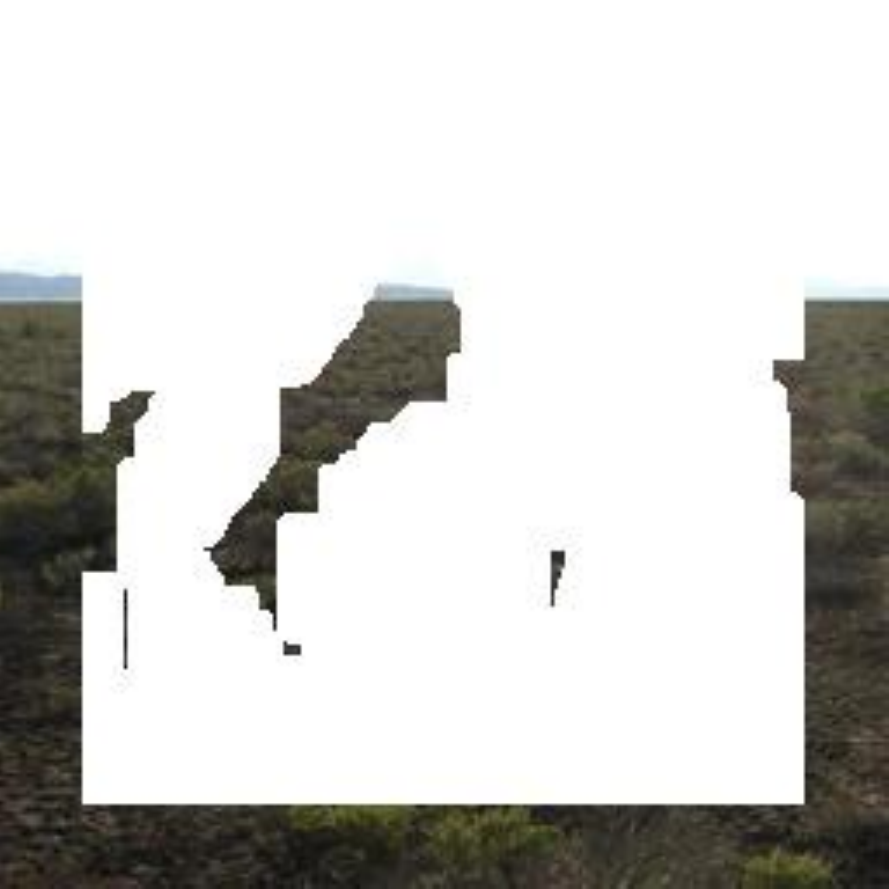} &
			\hspace{-4mm}
			\includegraphics[width=0.12\textwidth]{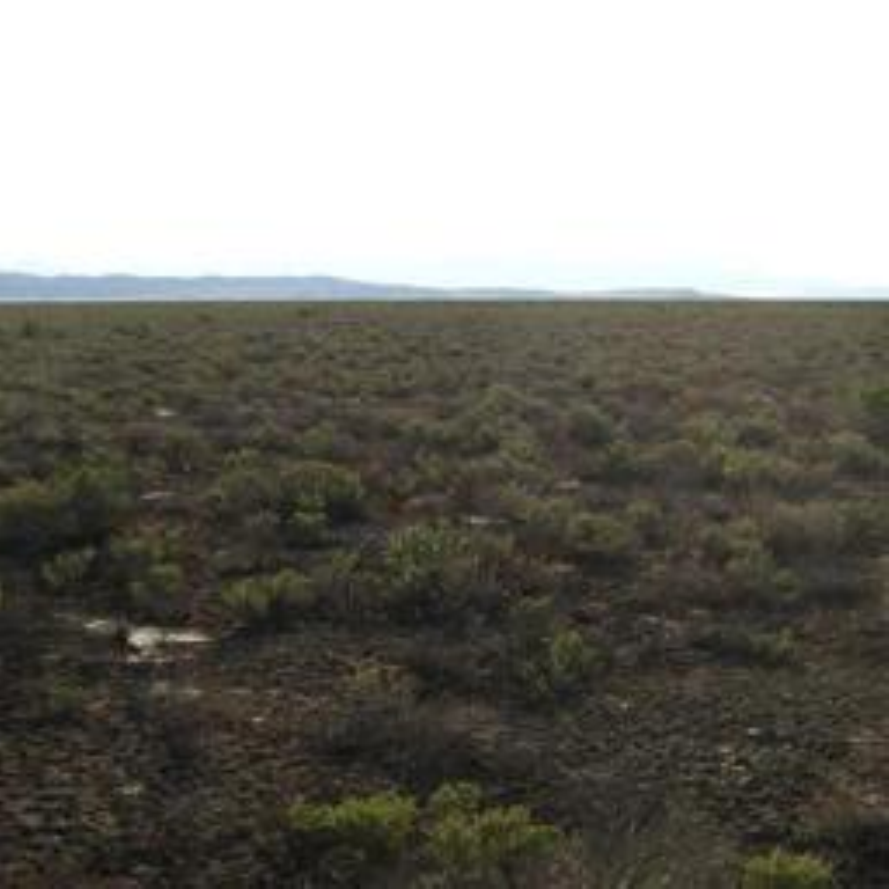} &
			\hspace{-4mm}
			\includegraphics[width=0.12\textwidth]{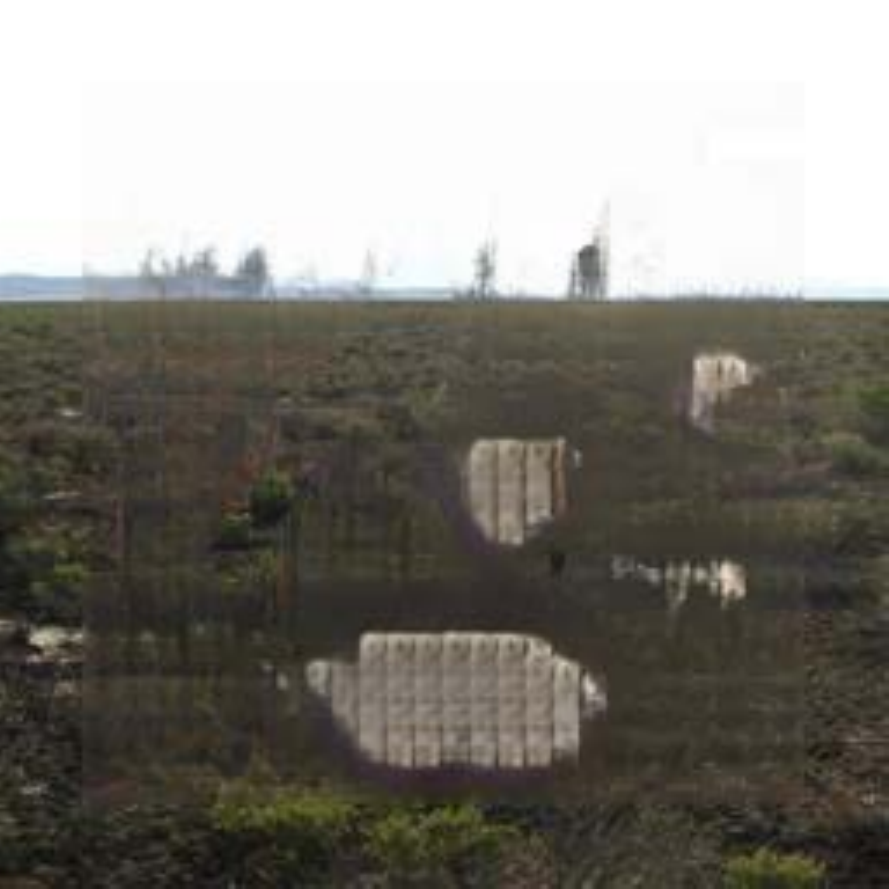} &
			\hspace{-4mm}
			\includegraphics[width=0.12\textwidth]{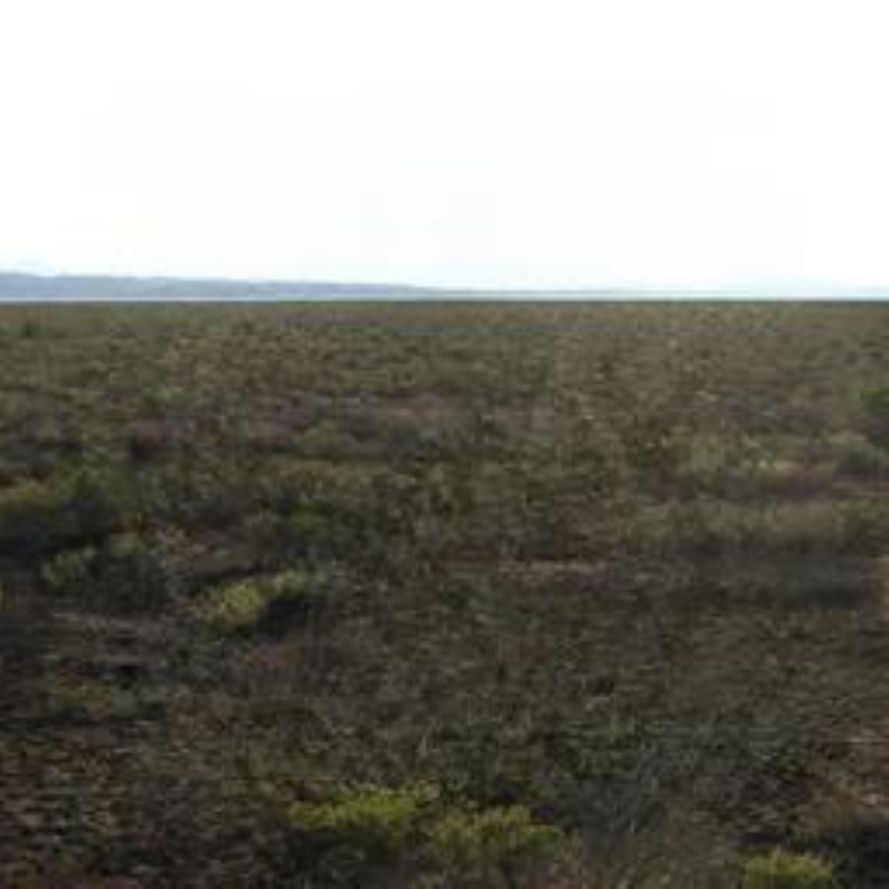}\\
			
			\includegraphics[width=0.12\textwidth]{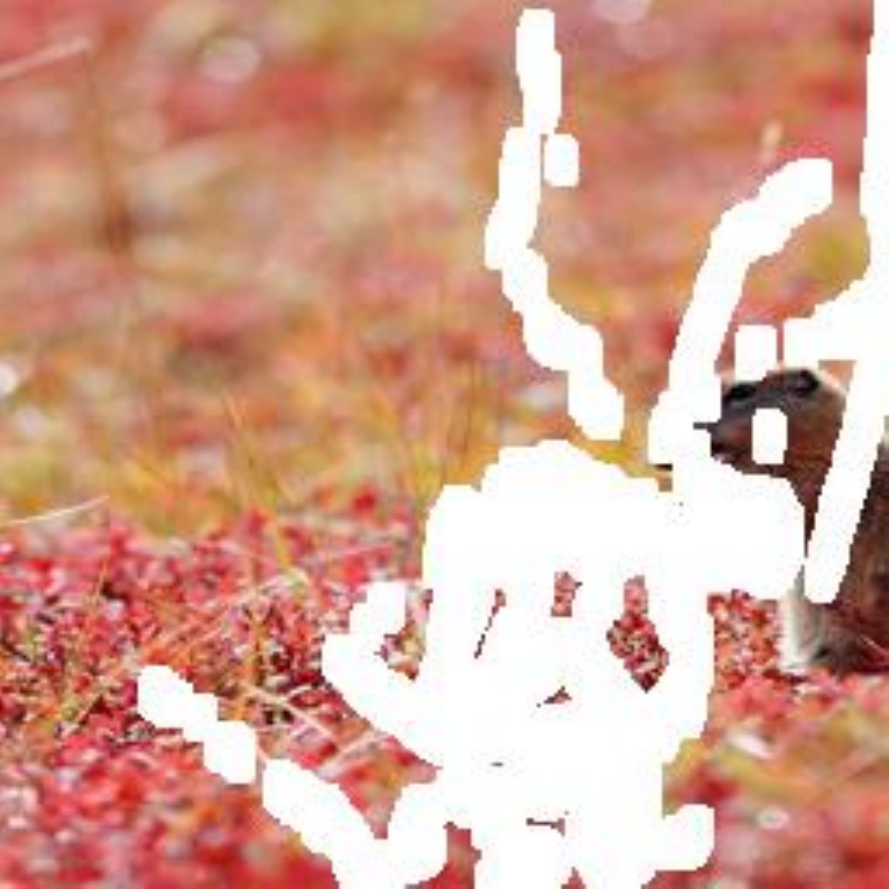} &
			\hspace{-4mm}
			\includegraphics[width=0.12\textwidth]{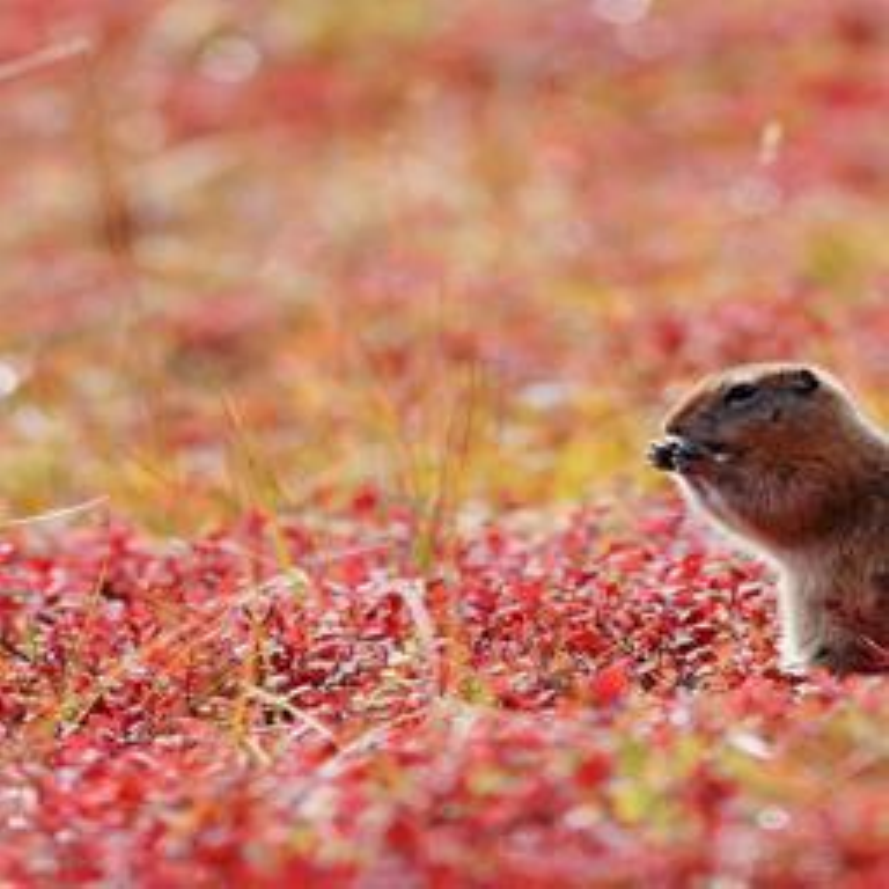} &
			\hspace{-4mm}
			\includegraphics[width=0.12\textwidth]{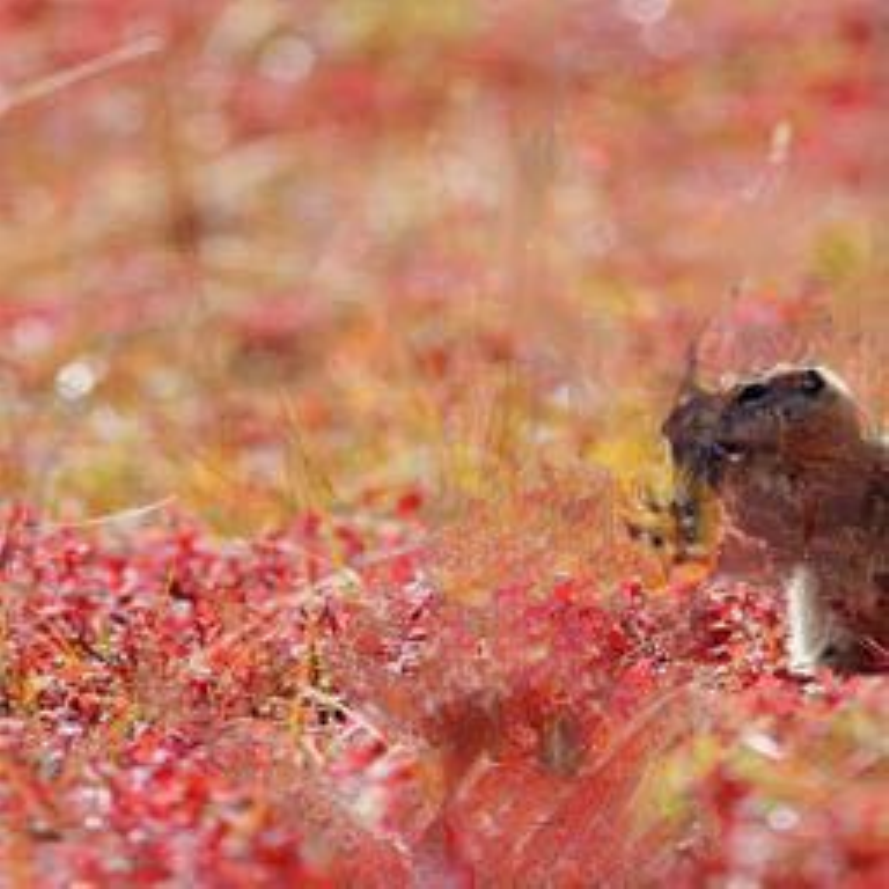} &
			\hspace{-4mm}
			\includegraphics[width=0.12\textwidth]{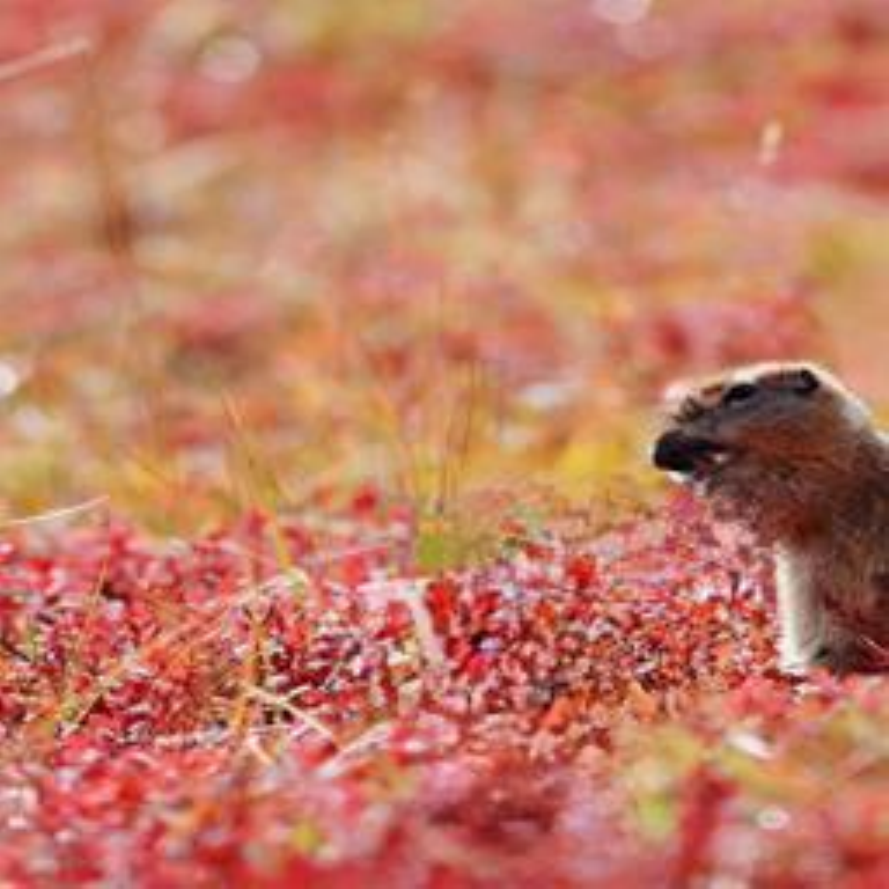}\\
			
			Input & \hspace{-4mm} GT & \hspace{-4mm} PICNet~\cite{PICNet} & \hspace{-4mm} DMFN (Ours) \\
		\end{tabular}
	\end{adjustbox}
	\caption{Visual comparisons on Places2.}
	\label{fig:places2}
\end{figure}

\begin{figure}[ht]
	\centering
	\begin{adjustbox}{valign=t}
		\begin{tabular}{ccc}
			\includegraphics[width=0.15\textwidth]{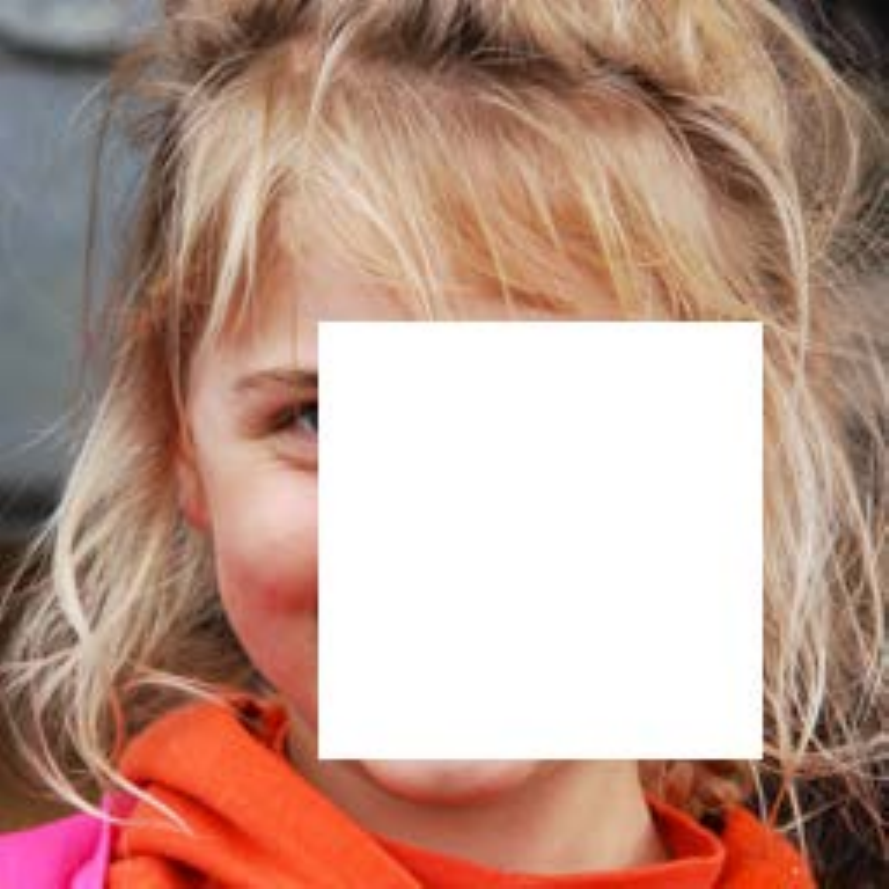} &
			\hspace{-4mm}
			\includegraphics[width=0.15\textwidth]{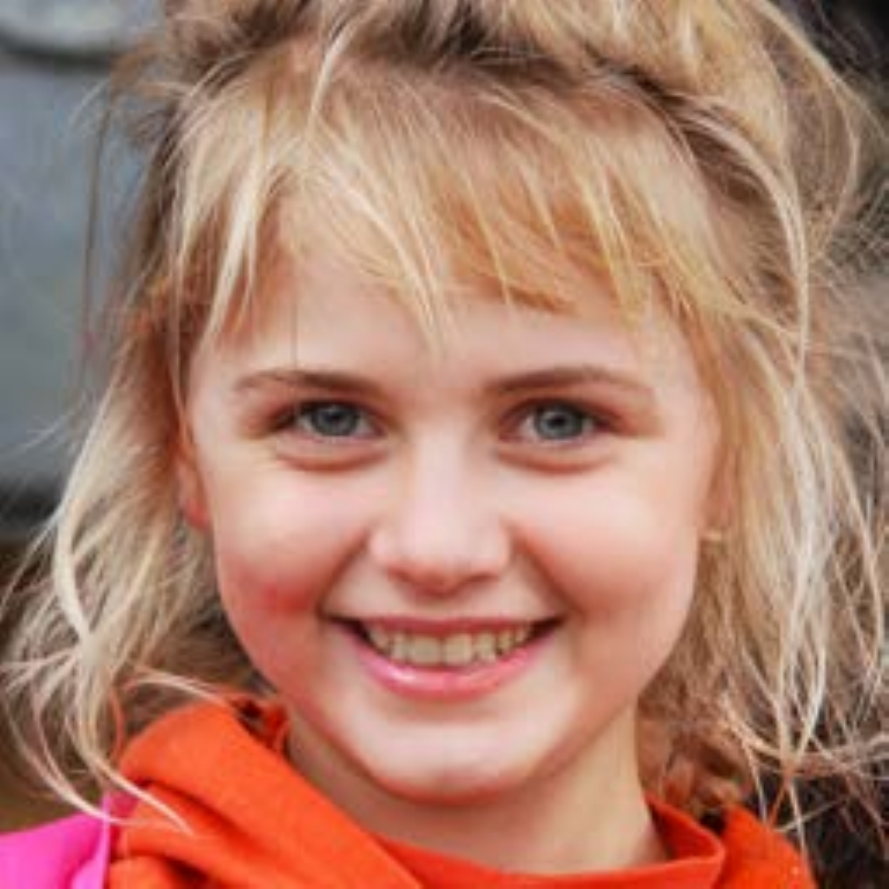} &
			\hspace{-4mm}
			\includegraphics[width=0.15\textwidth]{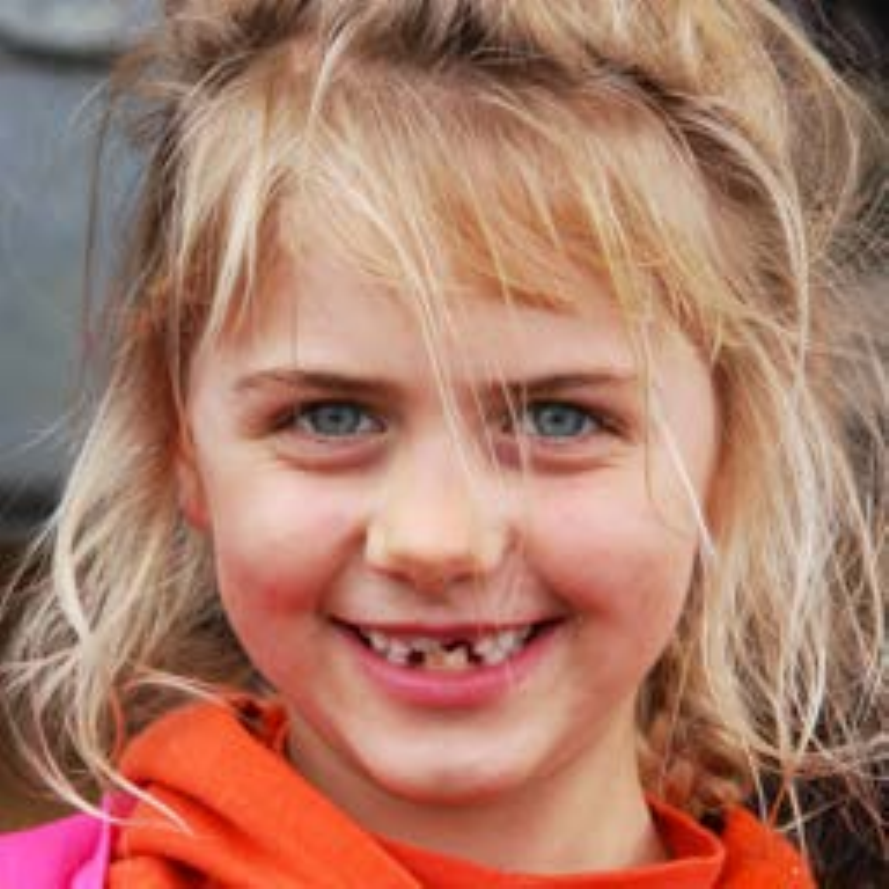} \\

			\includegraphics[width=0.15\textwidth]{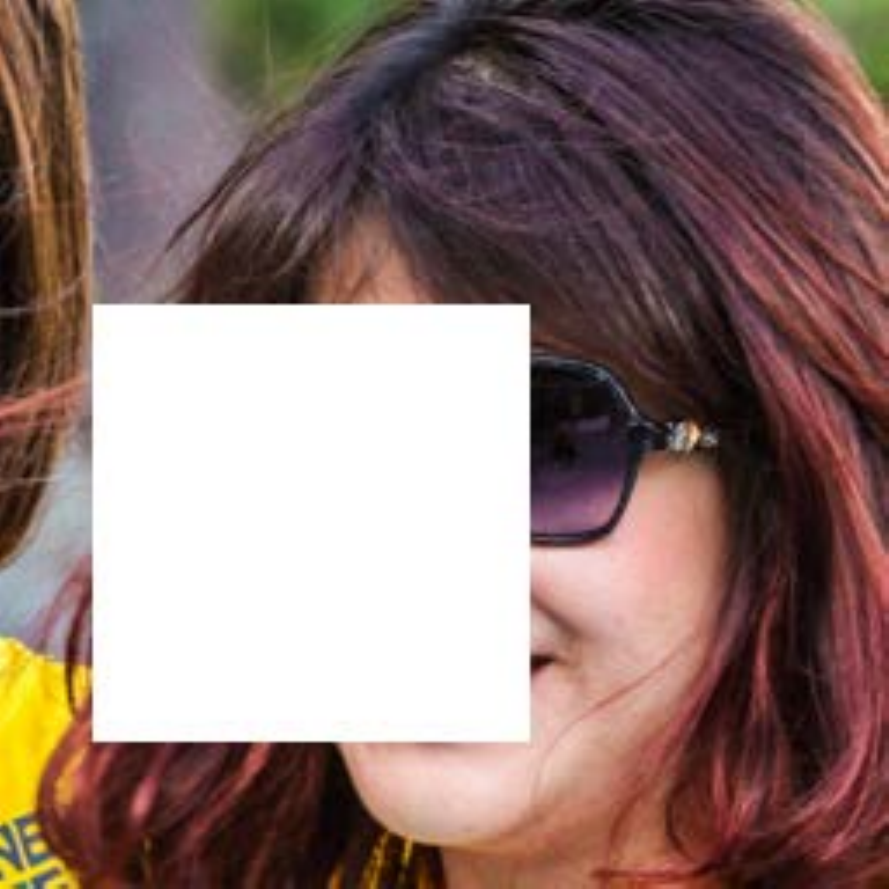} &
			\hspace{-4mm}
			\includegraphics[width=0.15\textwidth]{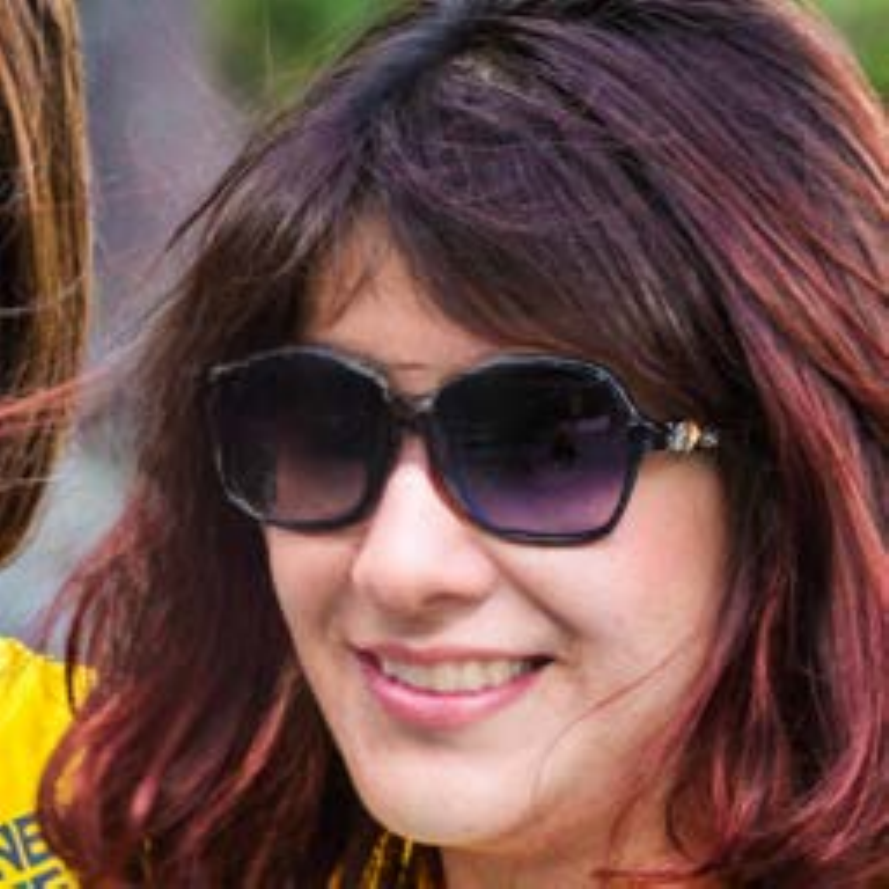} &
			\hspace{-4mm}
			\includegraphics[width=0.15\textwidth]{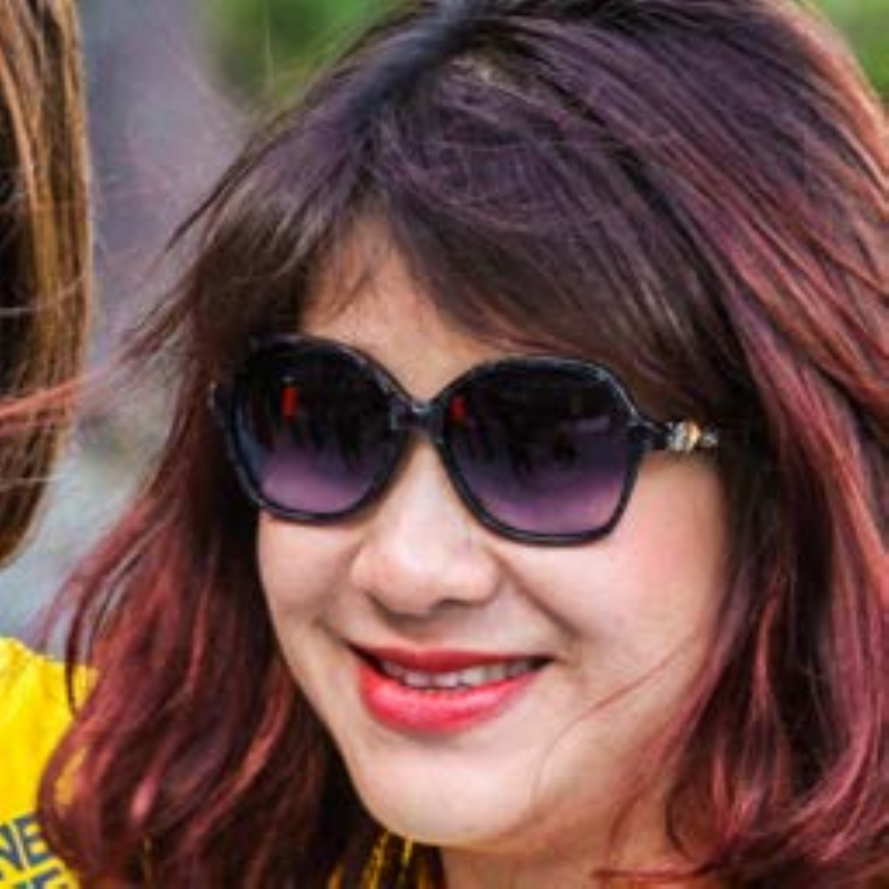} \\

			\includegraphics[width=0.15\textwidth]{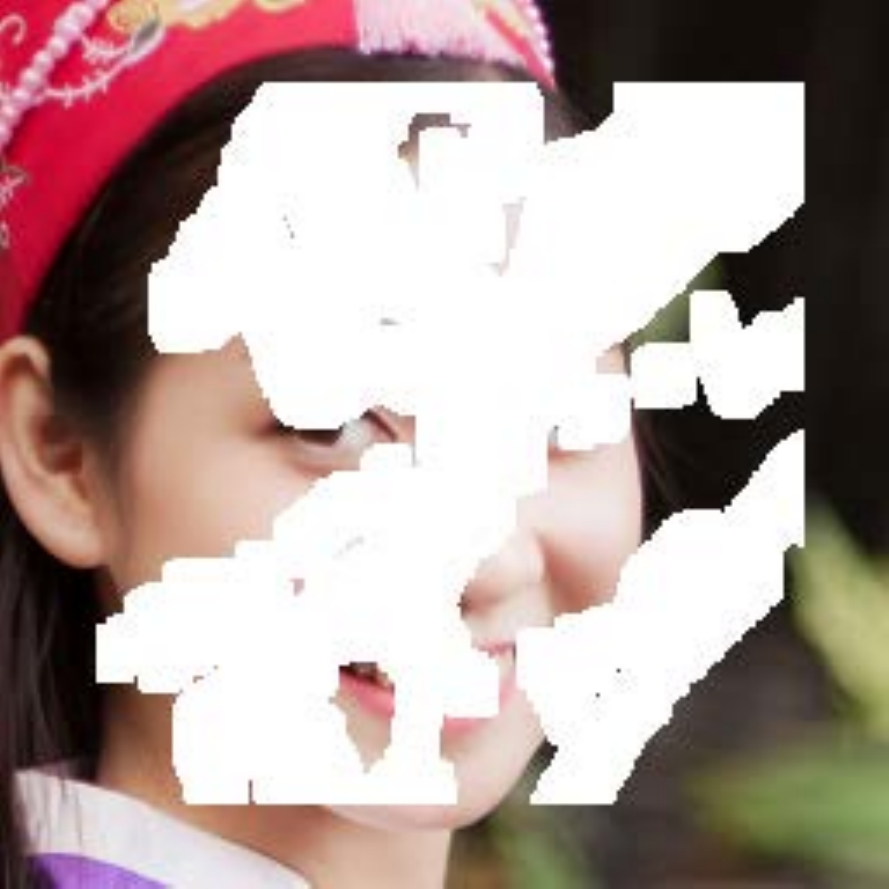} &
			\hspace{-4mm}
			\includegraphics[width=0.15\textwidth]{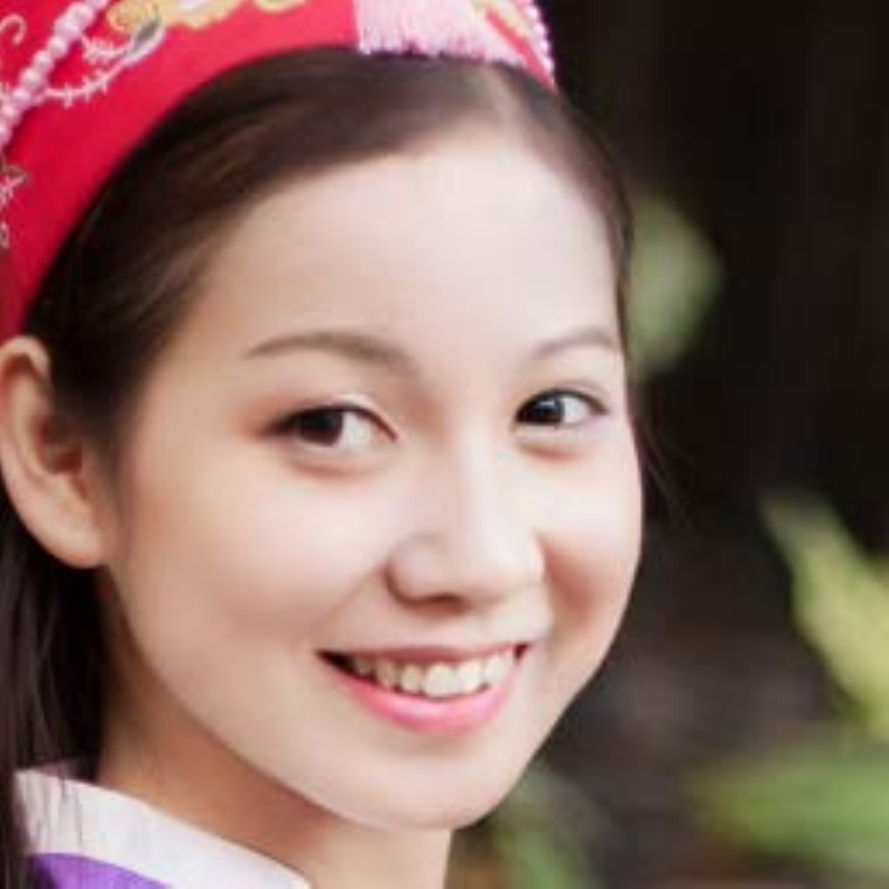} &
			\hspace{-4mm}
			\includegraphics[width=0.15\textwidth]{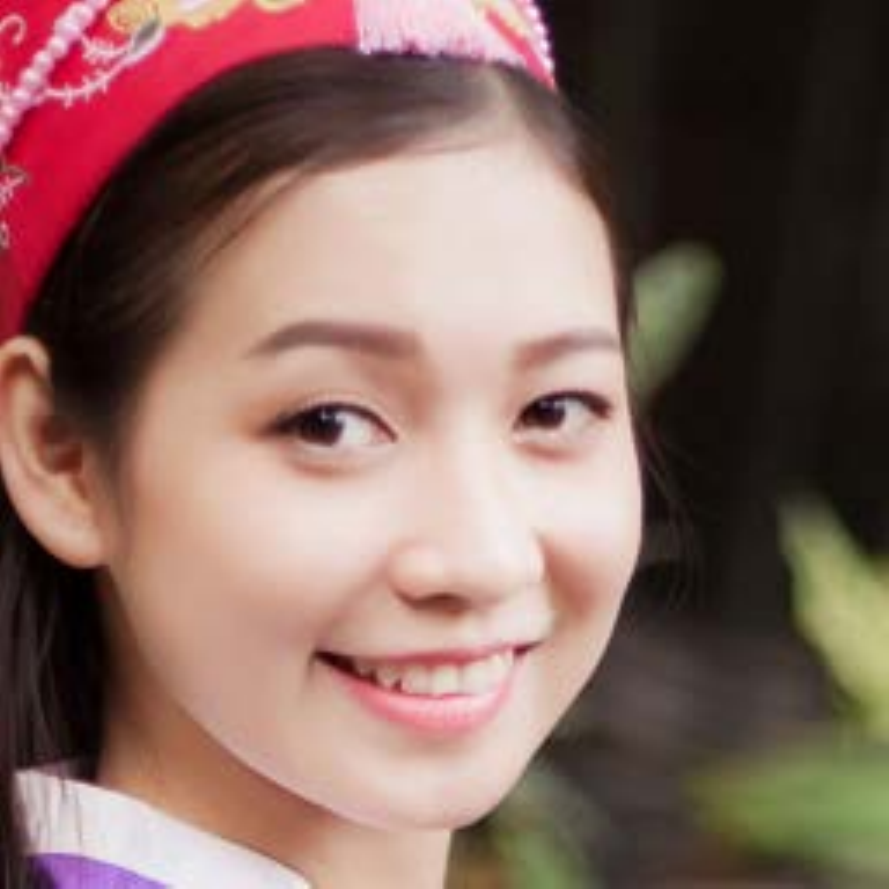} \\
			Input & \hspace{-4mm} DMFN (Ours) & \hspace{-4mm} GT \\
		\end{tabular}
	\end{adjustbox}
	\caption{Visual results on FFHQ dataset.}
	\label{fig:ffhq}
\end{figure}

\begin{figure}[ht]
	\centering
	\begin{adjustbox}{valign=t}
		\begin{tabular}{cccc}
			
			\includegraphics[width=0.115\textwidth]{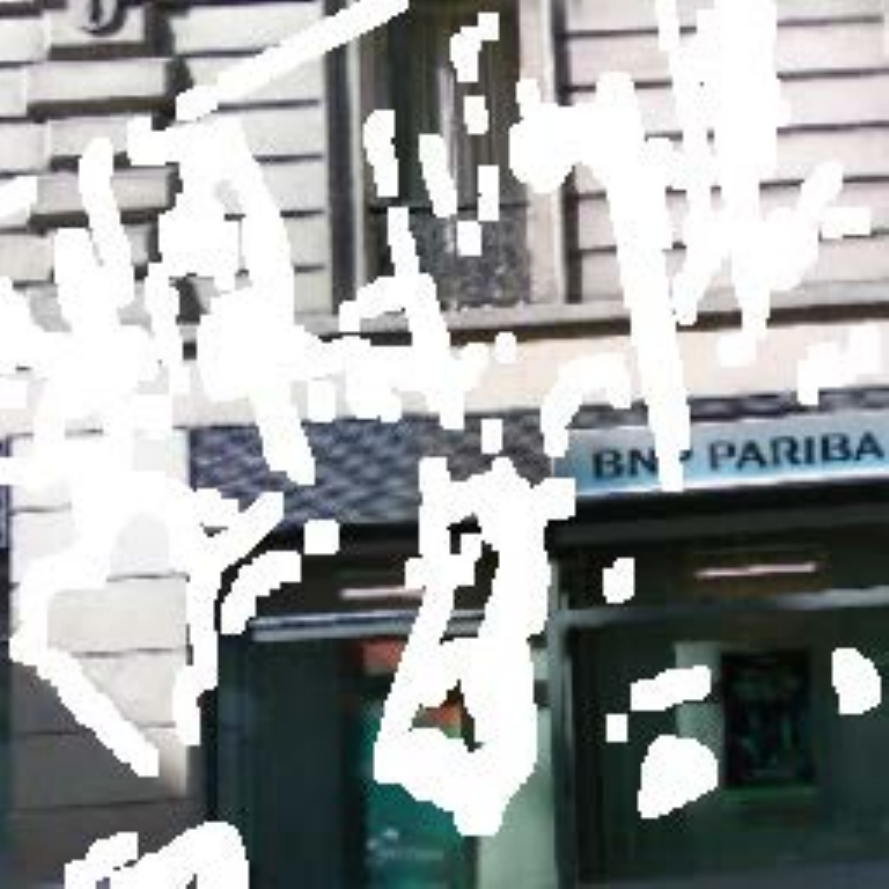} &
			\hspace{-4mm}
			\includegraphics[width=0.115\textwidth]{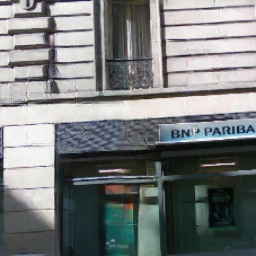} &
			\hspace{-4mm}
			\includegraphics[width=0.115\textwidth]{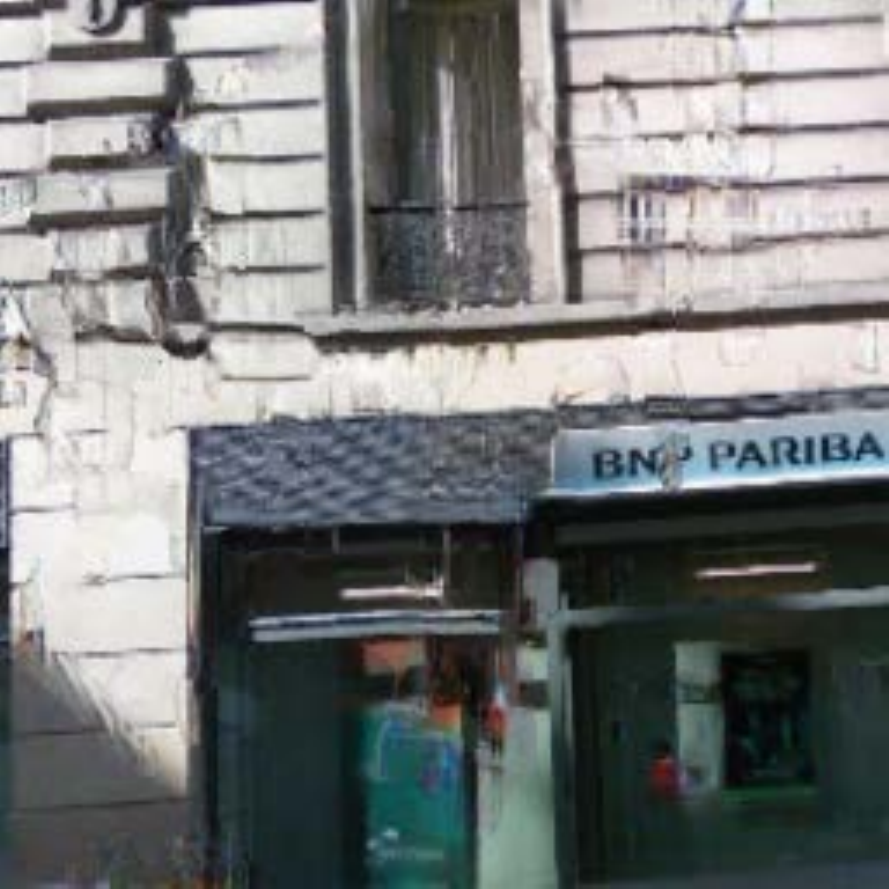} &
			\hspace{-4mm}
			\includegraphics[width=0.115\textwidth]{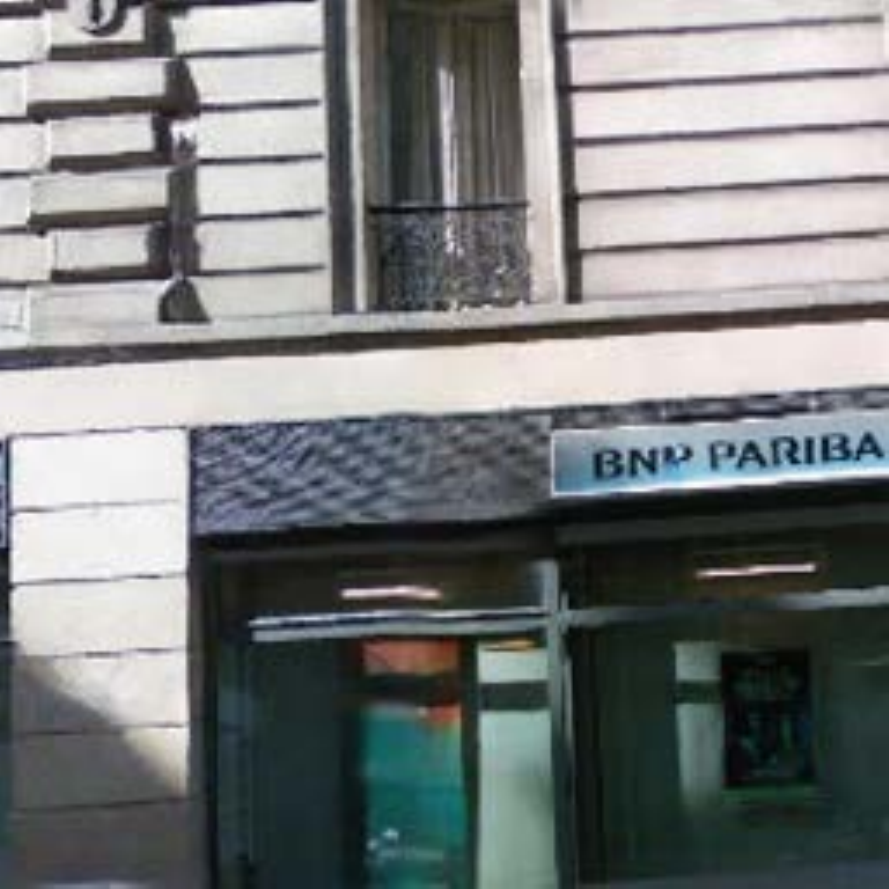} \\

			\includegraphics[width=0.115\textwidth]{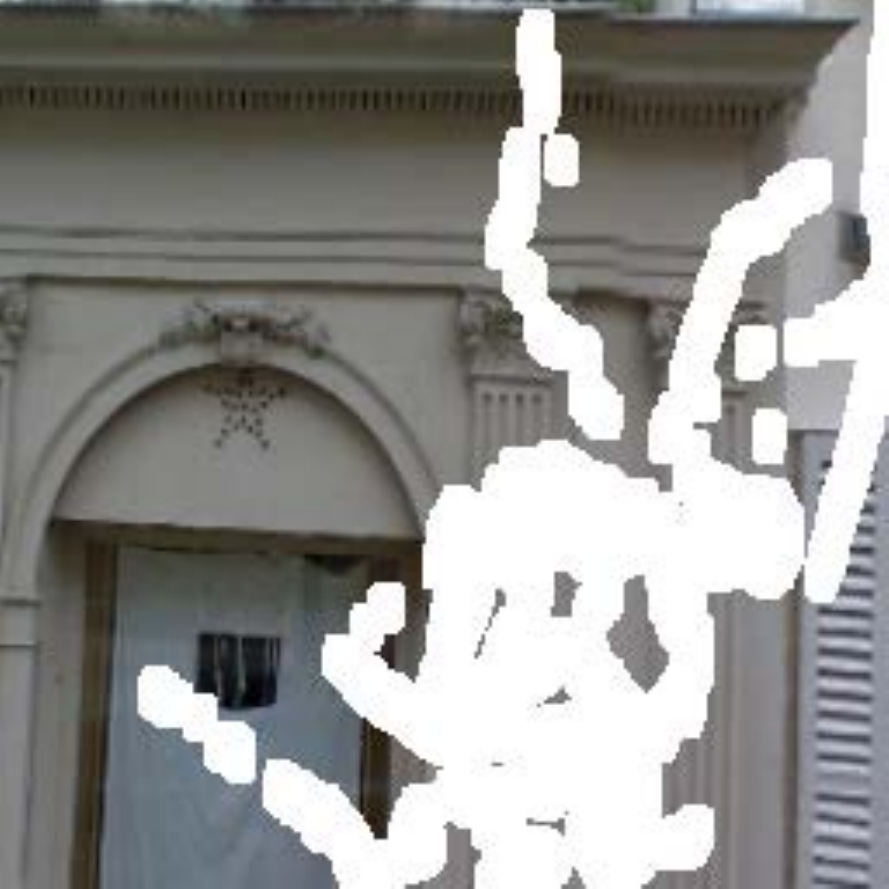} &
			\hspace{-4mm}
			\includegraphics[width=0.115\textwidth]{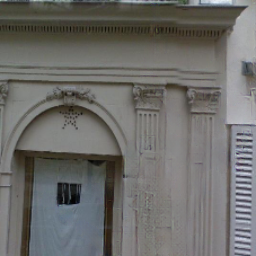} &
			\hspace{-4mm}
			\includegraphics[width=0.115\textwidth]{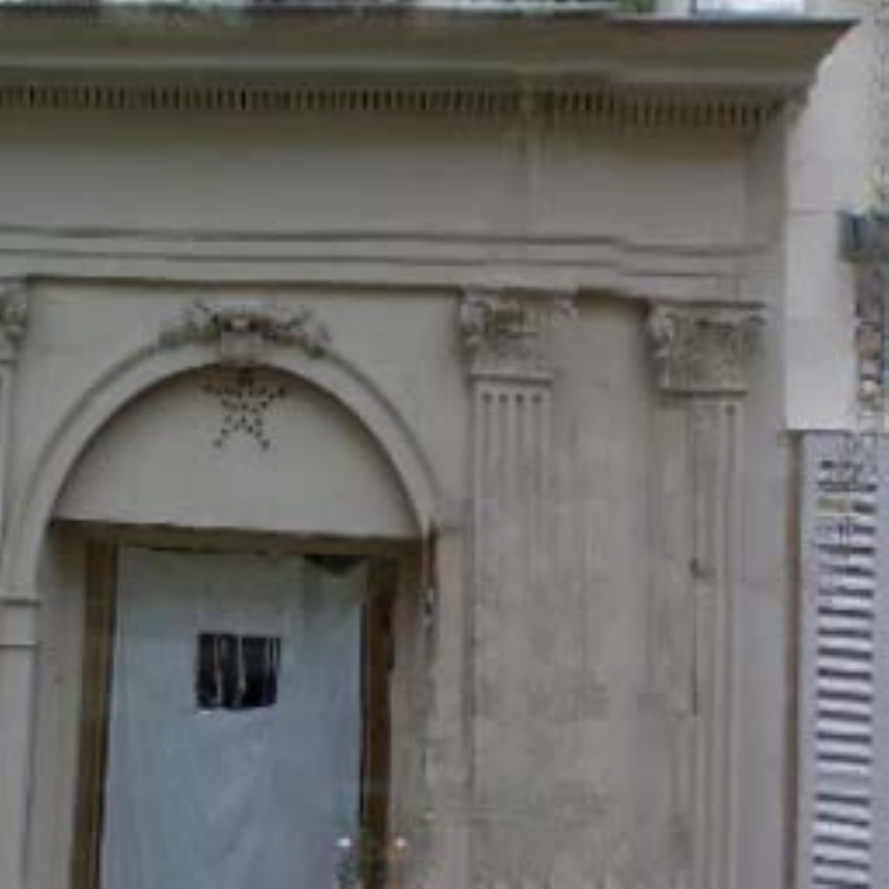} &
			\hspace{-4mm}
			\includegraphics[width=0.115\textwidth]{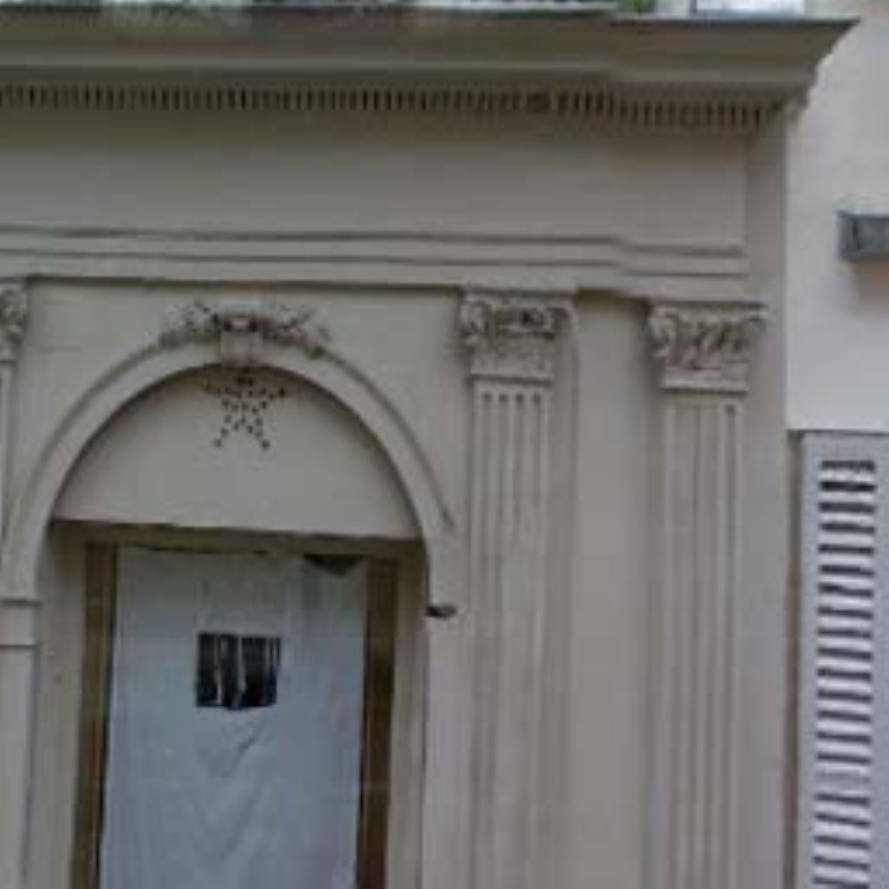} \\

			\includegraphics[width=0.115\textwidth]{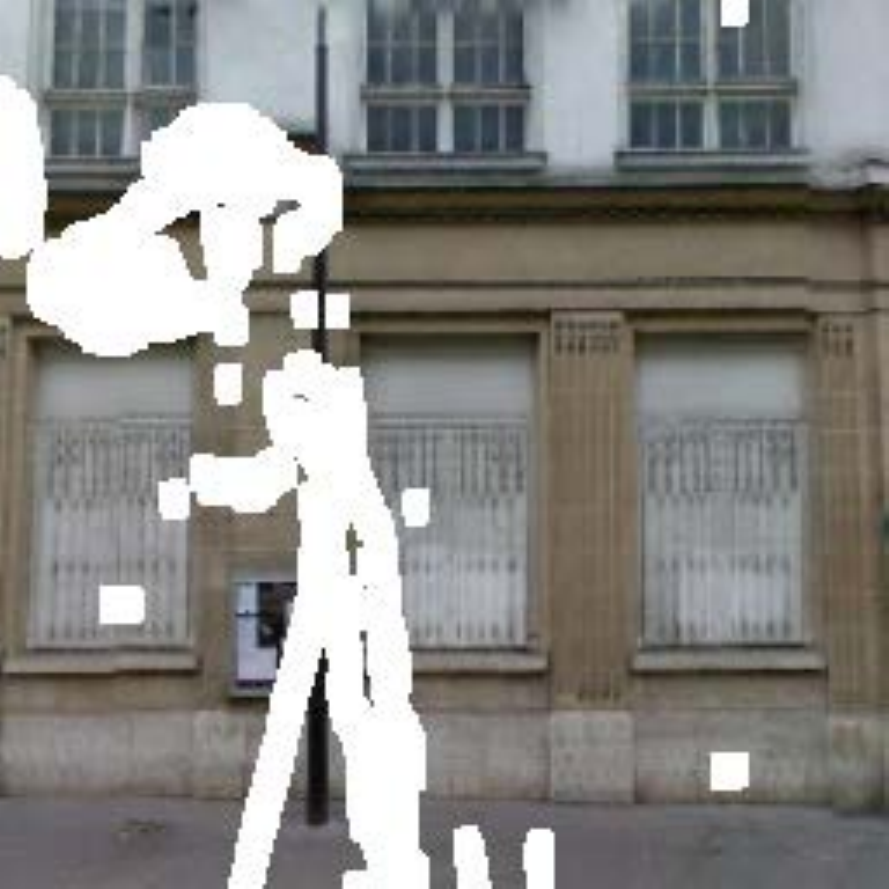} &
			\hspace{-4mm}
			\includegraphics[width=0.115\textwidth]{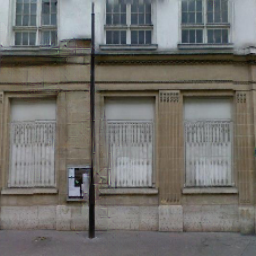} &
			\hspace{-4mm}
			\includegraphics[width=0.115\textwidth]{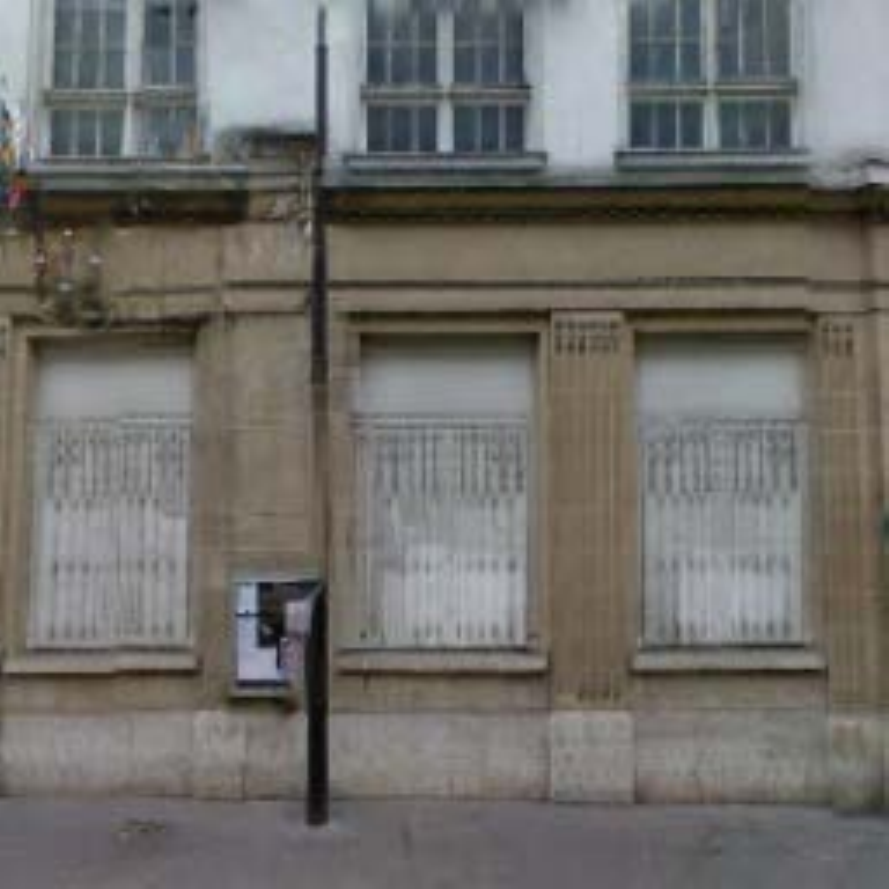} &
			\hspace{-4mm}
			\includegraphics[width=0.115\textwidth]{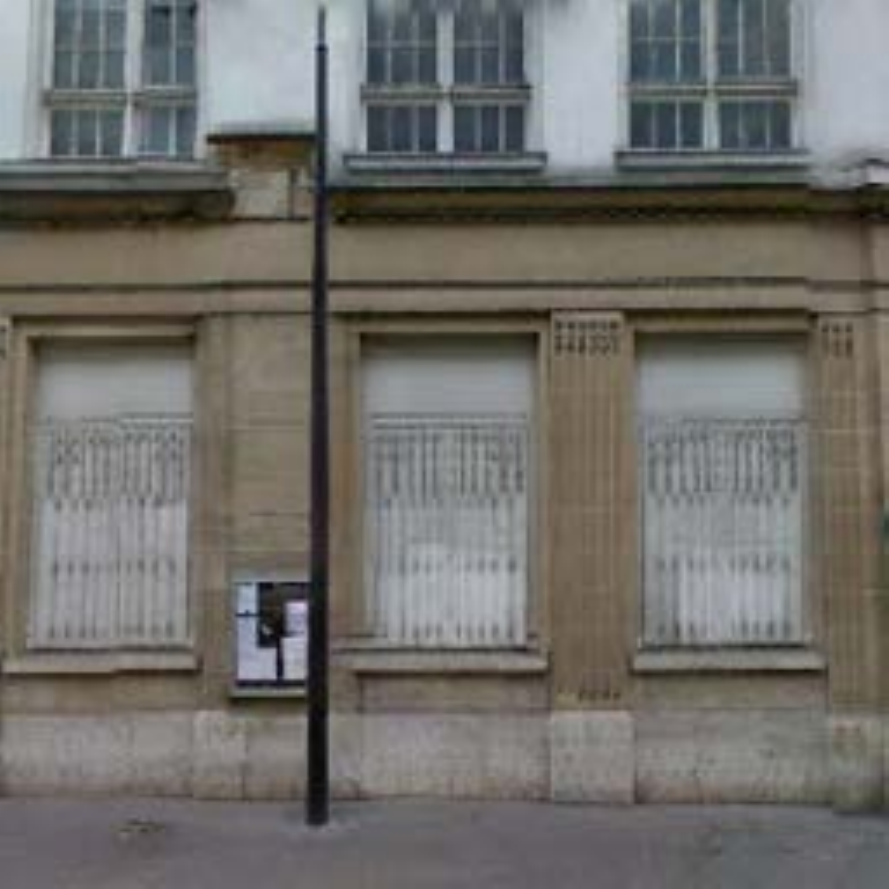} \\

			Input &\hspace{-4mm} LBAM~\cite{LBAM} &\hspace{-4mm} PICNet~\cite{PICNet} &\hspace{-4mm} DMFN (Ours) \\
		\end{tabular}
	\end{adjustbox}
	\caption{Inpainted images with irregular masks on Paris StreetView.}
	\label{fig:irregular1}
\end{figure}

\begin{figure}[ht]
	\centering
	\begin{adjustbox}{valign=t}
		\begin{tabular}{cccc}
			
			\includegraphics[width=0.115\textwidth]{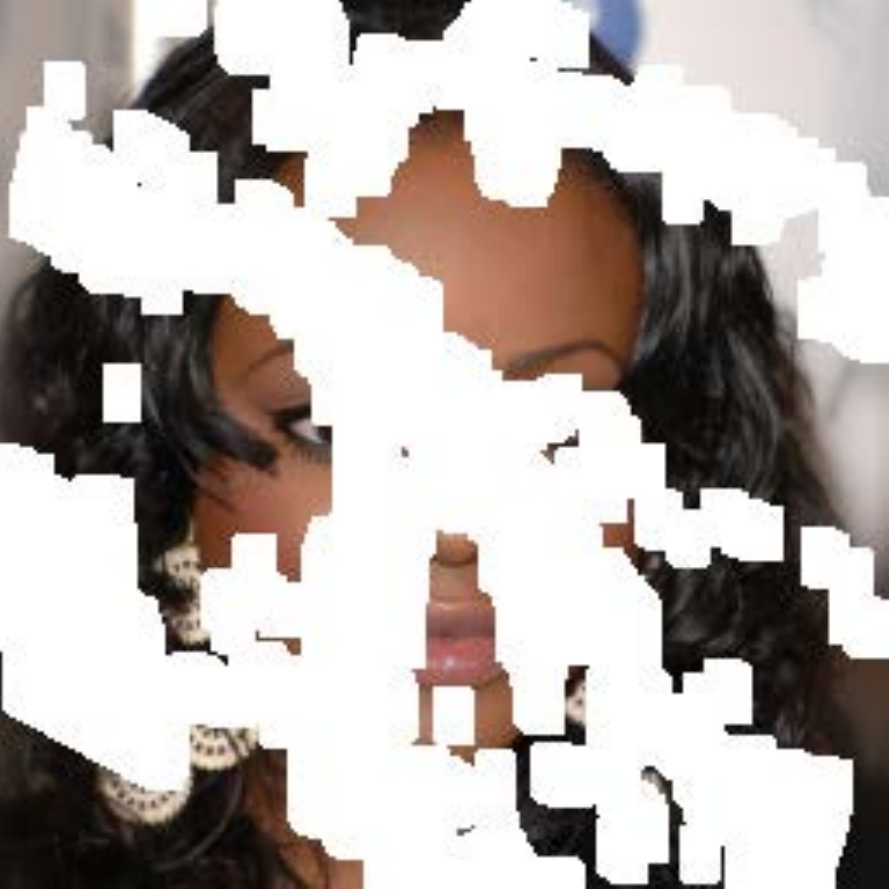} &
			\hspace{-4mm}
			\includegraphics[width=0.115\textwidth]{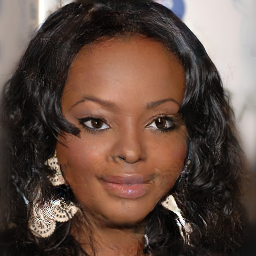} &
			\hspace{-4mm}
			\includegraphics[width=0.115\textwidth]{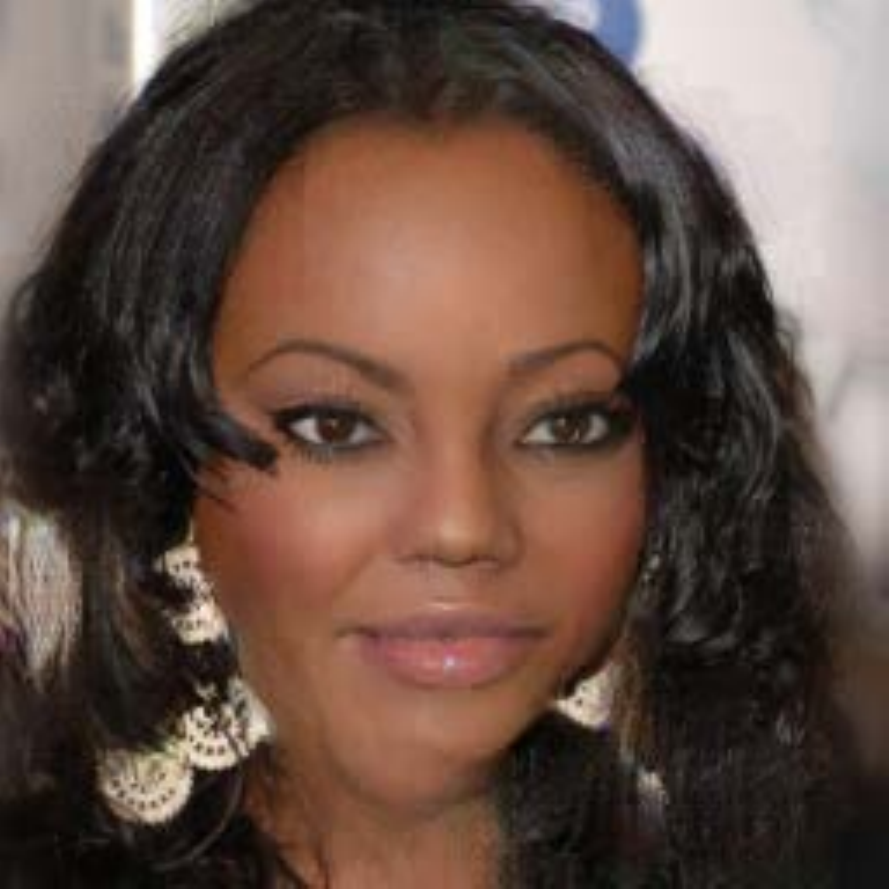} &
			\hspace{-4mm}
			\includegraphics[width=0.115\textwidth]{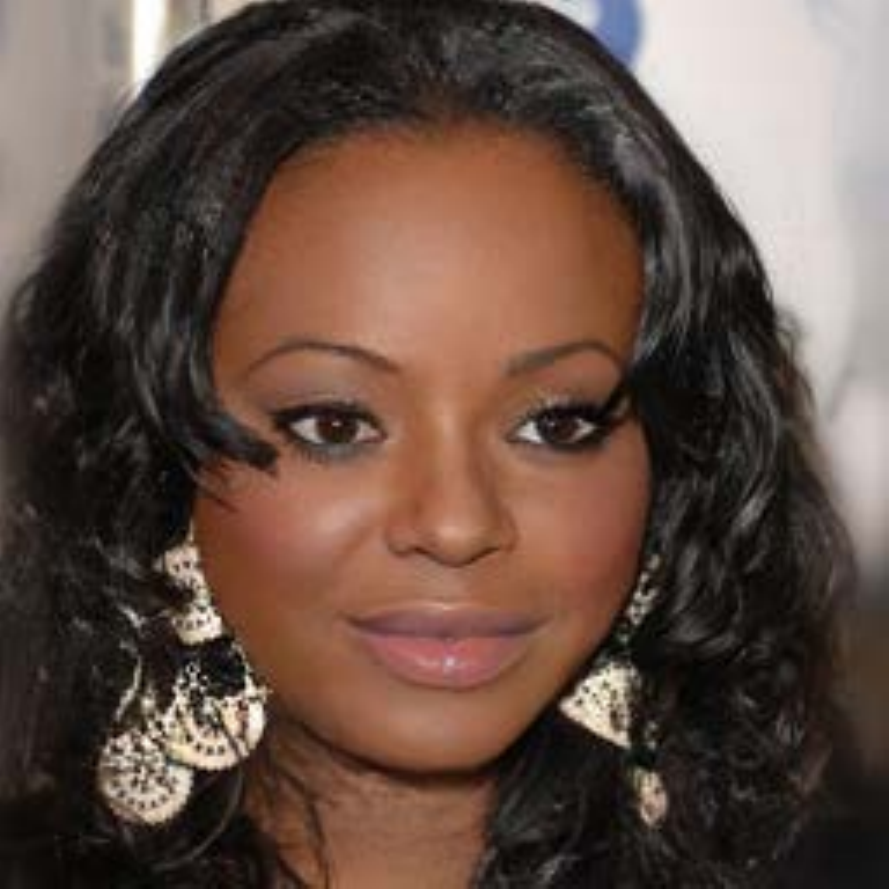} \\

			\includegraphics[width=0.115\textwidth]{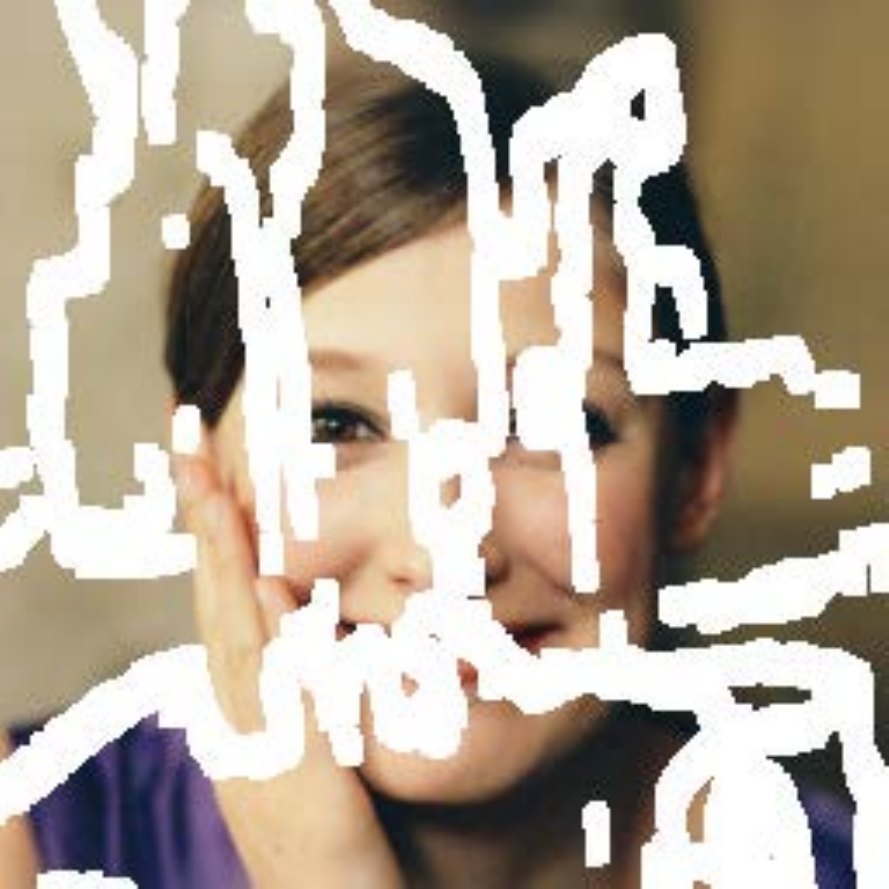} &
			\hspace{-4mm}
			\includegraphics[width=0.115\textwidth]{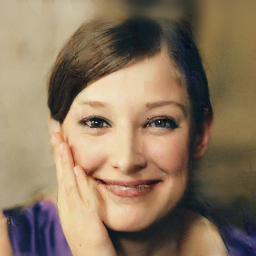} &
			\hspace{-4mm}
			\includegraphics[width=0.115\textwidth]{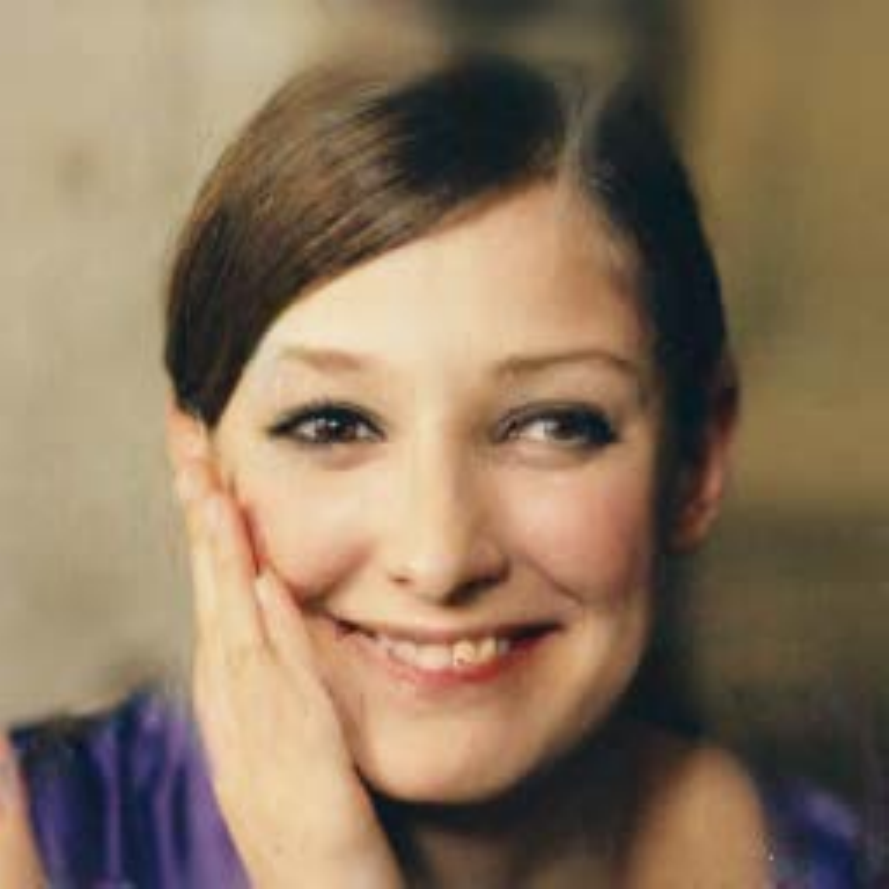} &
			\hspace{-4mm}
			\includegraphics[width=0.115\textwidth]{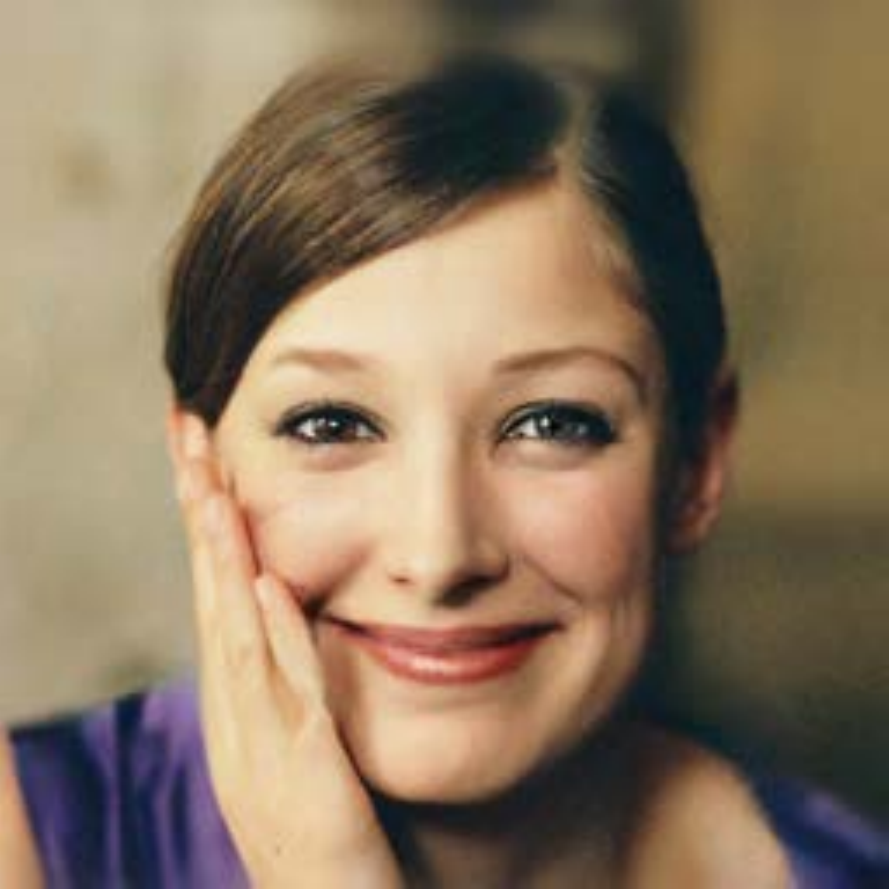} \\

			\includegraphics[width=0.115\textwidth]{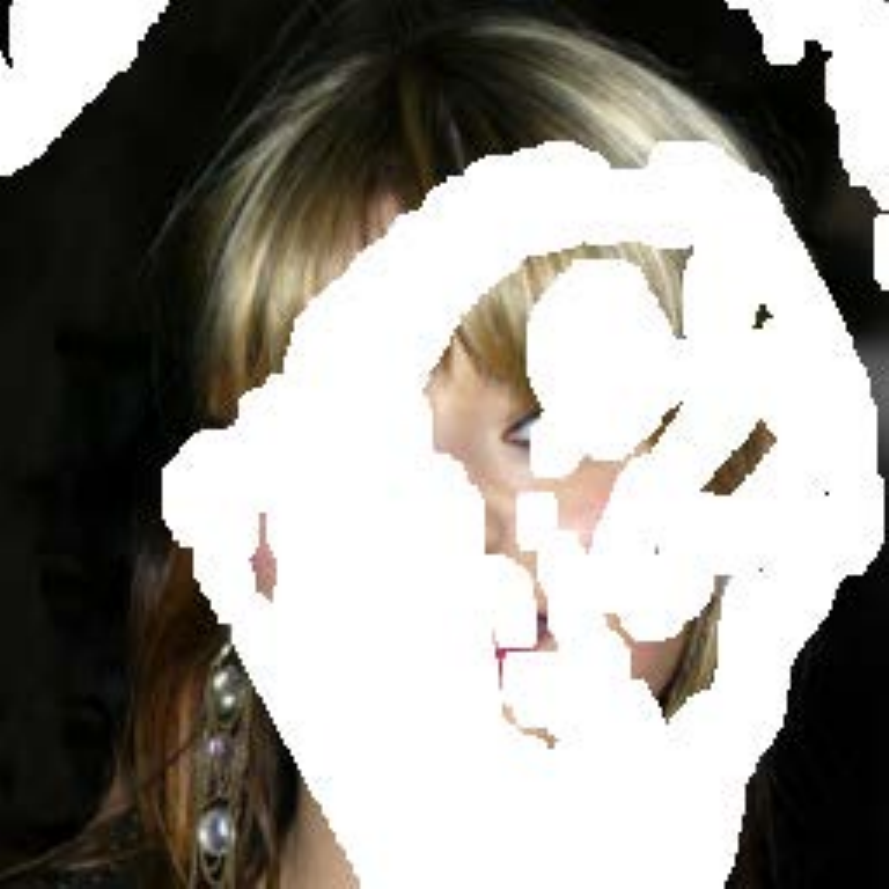} &
			\hspace{-4mm}
			\includegraphics[width=0.115\textwidth]{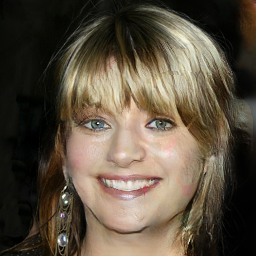} &
			\hspace{-4mm}
			\includegraphics[width=0.115\textwidth]{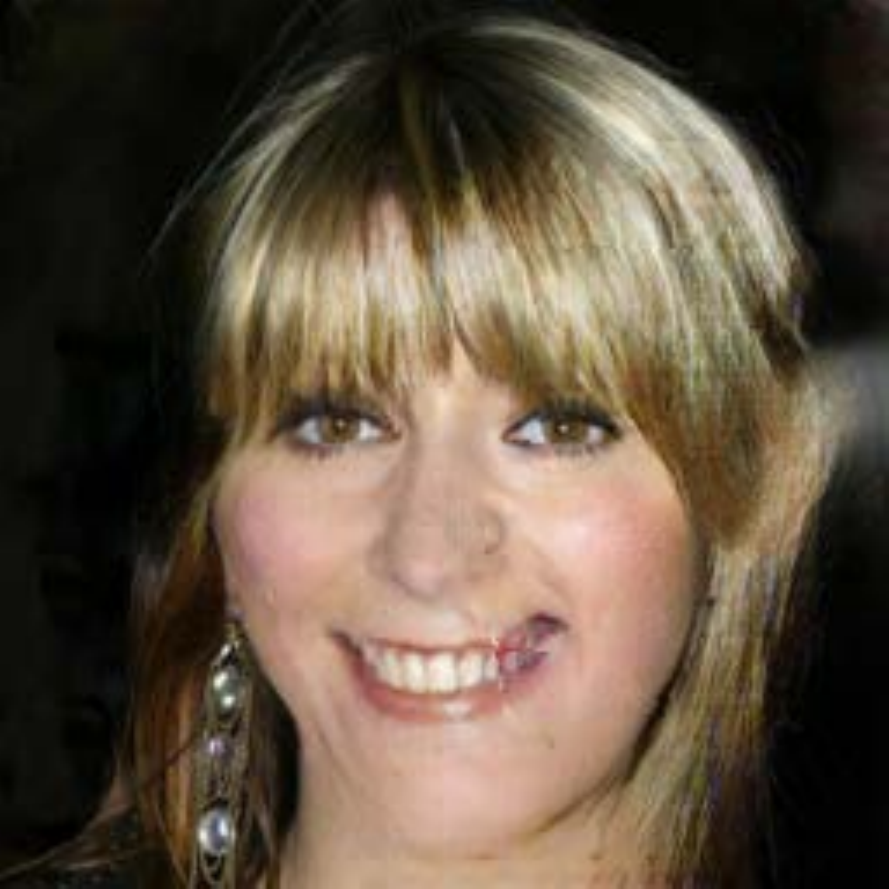} &
			\hspace{-4mm}
			\includegraphics[width=0.115\textwidth]{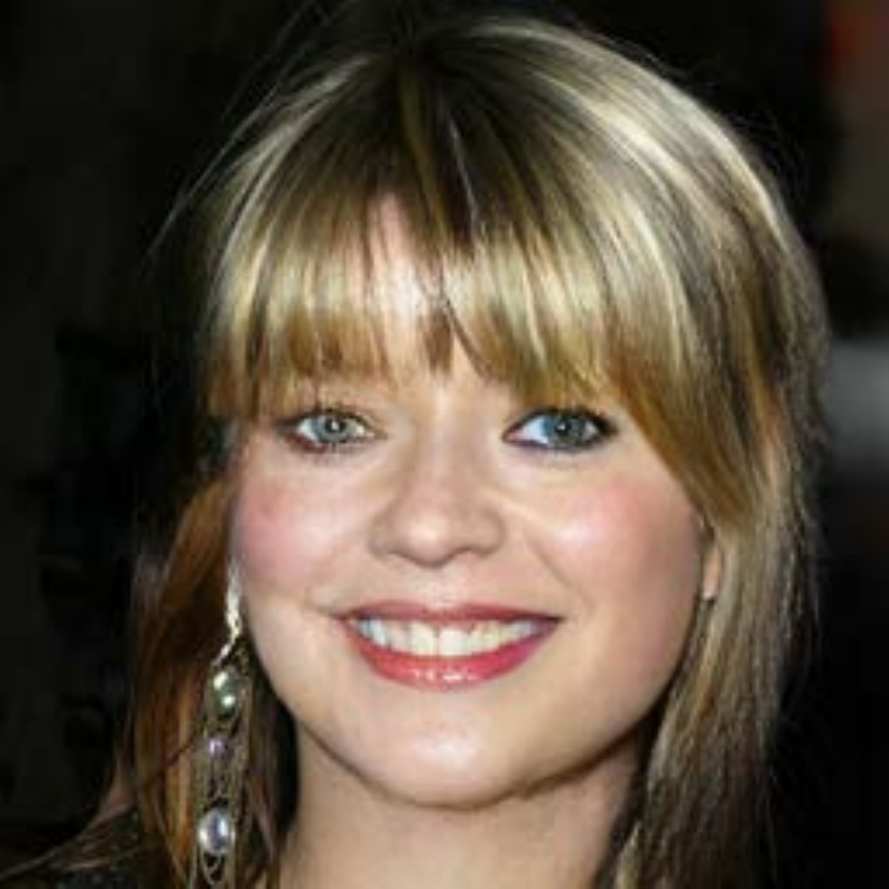} \\
			
			Input &\hspace{-4mm} GC~\cite{GC} &\hspace{-4mm} PICNet~\cite{PICNet} &\hspace{-4mm} DMFN (Ours) \\
		\end{tabular}
	\end{adjustbox}
	\caption{Inpainted images with irregular masks on CelebA-HQ.}
	\label{fig:irregular2}
\end{figure}

As shown in Figures~\ref{fig:paris-streetview},~\ref{fig:celeba-hq}, and~\ref{fig:places2}, compared with other state-of-the-art methods, our model gives a noticeable visual improvement on textures and structures. For instance, our network generates plausible image structures in Figure~\ref{fig:paris-streetview}, which mainly stems from the dense multi-scale fusion architecture and well-designed losses. The realistic textures are hallucinated via feature matching and adversarial training. For Figure~\ref{fig:celeba-hq}, we show that our results with more realistic details and fewer artifacts than the compared approaches. Besides, we give partial results of our method and PICNet~\cite{PICNet} on Places2 dataset in Figure~\ref{fig:places2}. The proposed DMFN creates more reasonable, natural, and photo-realistic images. Additionally, we also show some example results (masks at random position) of our model trained on FFHQ in Figure~\ref{fig:ffhq}. In Figure~\ref{fig:irregular1}, our method performs more stable and fine for large-area irregular masks than compared algorithms. More compelling results can be found in the \emph{supplementary material}.

\subsection{Quantitative comparisons}
\begin{table*}[htpb]
	\centering
	\caption{Quantitative results (center regular mask) on four testing datasets.}
	\label{tab:quantitative-results}
	\begin{tabular}{|c|c|c|c|c|}
		\hline
		\multirow{2}{*}{Method} & Paris street view (100) & Places2 (100) & CelebA-HQ (2,000) & FFHQ (10,000) \\
		\cline{2-5}
		& LPIPS / PSNR / SSIM & LPIPS / PSNR / SSIM & LPIPS / PSNR / SSIM & LPIPS / PSNR / SSIM \\
		\hline
		CA~\cite{contextual-attention} & N/A & 0.1524 / 21.32 / 0.8010 & 0.0724 / 24.13 / 0.8661 & N/A \\
		GMCNN~\cite{GMCNN} & 0.1243 / 24.38 / 0.8444  & 0.1829 / 19.51 / 0.7817 & 0.0509 / 25.88 / 0.8879 & N/A \\
		PICNet~\cite{PICNet} & 0.1263 / 23.79 / 0.8314 & 0.1622 / 20.70 / 0.7931 & N/A & N/A \\
		PENNet~\cite{PEN-Net} & N/A & 0.2384 / 21.93 / 0.7586 & 0.0676 / 25.50 / 0.8813 & N/A \\
		DMFN (Ours) & \textbf{0.1018} / \textbf{25.00} / \textbf{0.8563} & \textbf{0.1188} / \textbf{22.36} / \textbf{0.8194} & \textbf{0.0460} / \textbf{26.50} / \textbf{0.8932} & \textbf{0.0457} / \textbf{26.49} / \textbf{0.8985} \\
		\hline
	\end{tabular}
\end{table*}

Following~\cite{contextual-attention,GMCNN}, we measure the quality of our results using peak signal-to-noise ratio (PSNR) and structural similarity (SSIM). Learned perceptual image patch similarity (LPIPS)~\cite{LPIPS} is a new metric that can better evaluate the perceptual similarity between two images. Because the purpose of image inpainting is to pursue visual effects, we adopt LPIPS as the main qualitative assessment. The lower the values of LPIPS, the better. In Places2, 100 validation images from ``canyon'' scene category are chosen for evaluation. As shown in Table~\ref{tab:quantitative-results}, our method produces acceptable results compared with CA~\cite{contextual-attention}, GMCNN~\cite{GMCNN}, PICNet~\cite{PICNet}, and PENNet~\cite{PEN-Net} in terms of all evaluation measurements. 

\begin{figure}[htpb]
	\centering
	\begin{tabular}{cc}
		\includegraphics[width=0.21\textwidth]{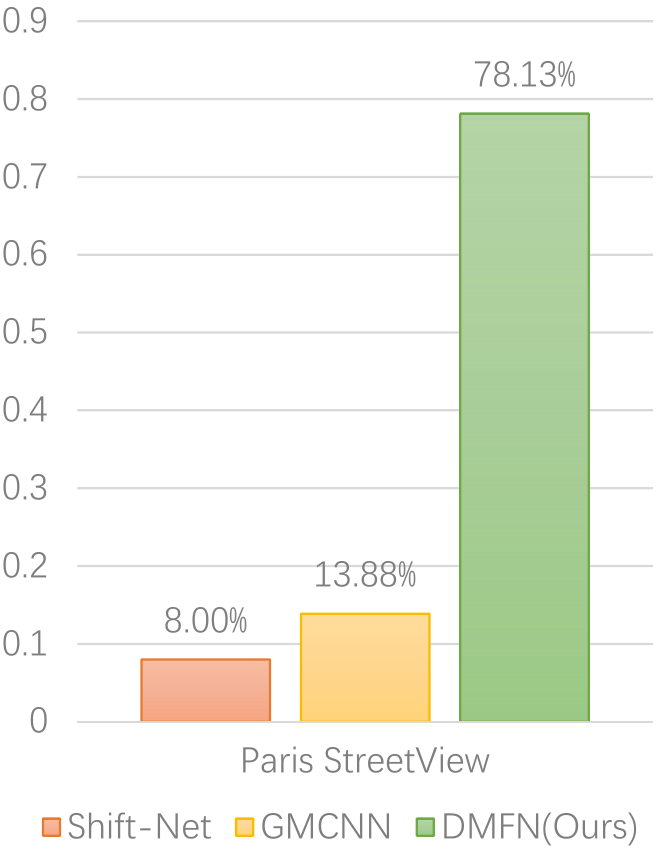} &
		\includegraphics[width=0.21\textwidth]{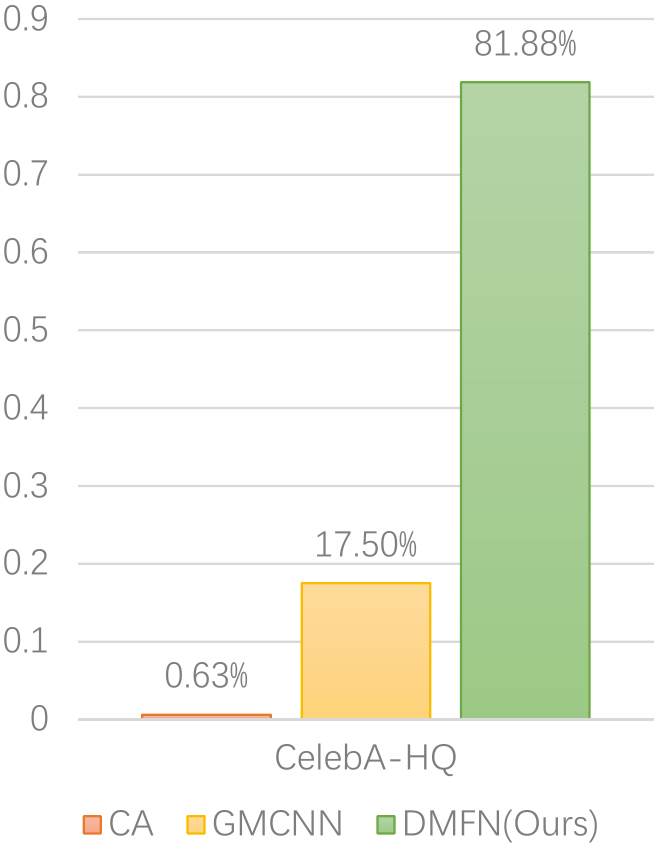} \\
	\end{tabular}
	\caption{Results of user study.}
	\label{fig:user-study}
\end{figure}

We also conducted user studies as illustrated in Figure~\ref{fig:user-study}. The scheme is based on blind randomized A/B/C tests deployed on Google Forms platform as in~\cite{GMCNN}. Each survey includes $40$ single-choice questions. Each question involves three options (completed images that are generated from the same corrupted input by three different methods). There are $20$ participants invited to accomplish this survey. They are asked to select to the most realistic item in each question. The option order is shuffled each time. Finally, our method outperforms compared approaches by a large margin.

\subsection{Ablation study}

\begin{table}[htpb]
	\centering
	\caption{Quantitative results of different structures on Paris street view dataset (center regular mask).}
	\label{tab:different-structures}
	\scalebox{0.9}{
		\begin{tabular}{|r|c|c|c|c|c|}
			\hline
			Model & rate=2 & rate=8 & w/o combination & w/o ${K_i}\left(  \cdot  \right)$ & DMFB \\
			\hline
			Params & 803,392 & 803,392 & 361,024 & 361,024 & 471,808 \\
			\hline
			LPIPS$ \downarrow$ & 0.1059 & 0.1067 & 0.1083 & 0.1026 & \textbf{0.1018} \\
			PSNR$ \uparrow$ & 24.93 & 24.91 & 24.24 & 24.93 & \textbf{25.00} \\
			SSIM$ \uparrow$ & 0.8530 & 0.8549 & 0.8505 & 0.8561 & \textbf{0.8563} \\
			\hline
	\end{tabular} }
\end{table}

\begin{table}[htpb]
	\centering
	\caption{Investigation of self-guided regression loss and geometrical alignment constraint on CelebA-HQ (random regular mask).}
	\label{tab:different-losses}
	\begin{tabular}{|r|c|c|c|c|}
		\hline
		Metric & w/o self-guided & w/o align & w/o dis\_fm & with all \\
		\hline
		LPIPS$ \downarrow$ & 0.0537 & 0.0534 & 0.0542 & \textbf{0.0530} \\
		PSNR$ \uparrow$ & 25.73 & 25.63 & 25.65 & \textbf{25.83} \\
		SSIM$ \uparrow$ & 0.8892 & 0.8884 & 0.8870 & \textbf{0.8892} \\
		\hline
	\end{tabular}
\end{table}

\begin{figure}[htpb]
	\centering
	\tiny
	\scalebox{0.95}{
		\begin{tabular}{cccccc}
			\includegraphics[width=0.08\textwidth]{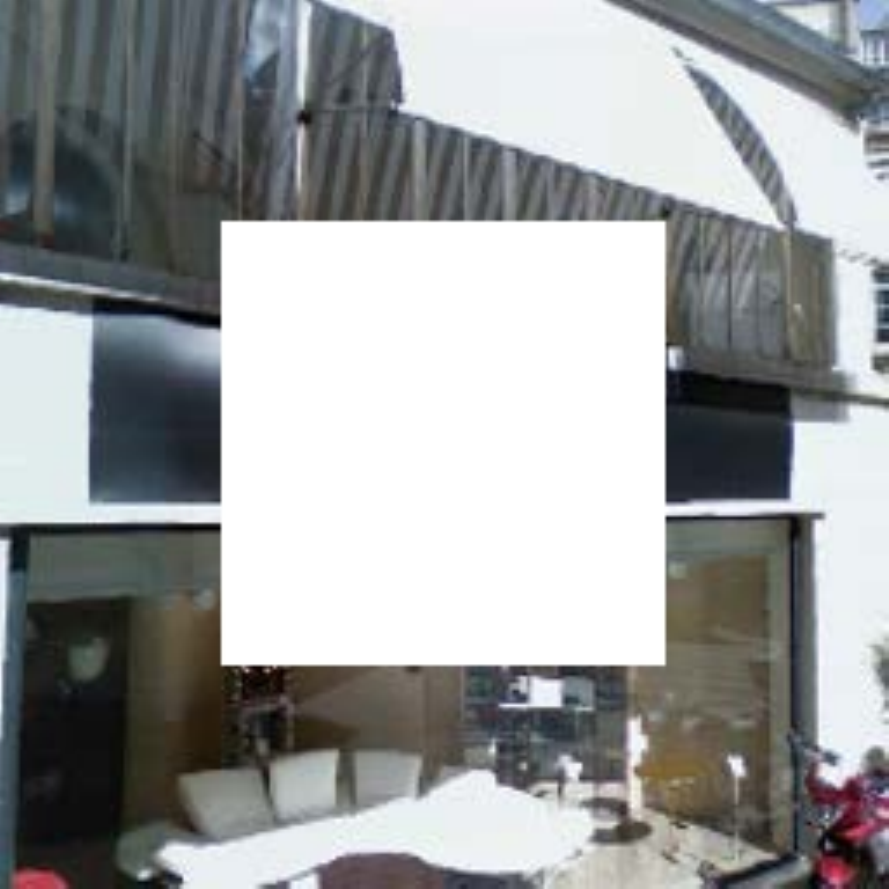} &
			\hspace{-4mm}
			\includegraphics[width=0.08\textwidth]{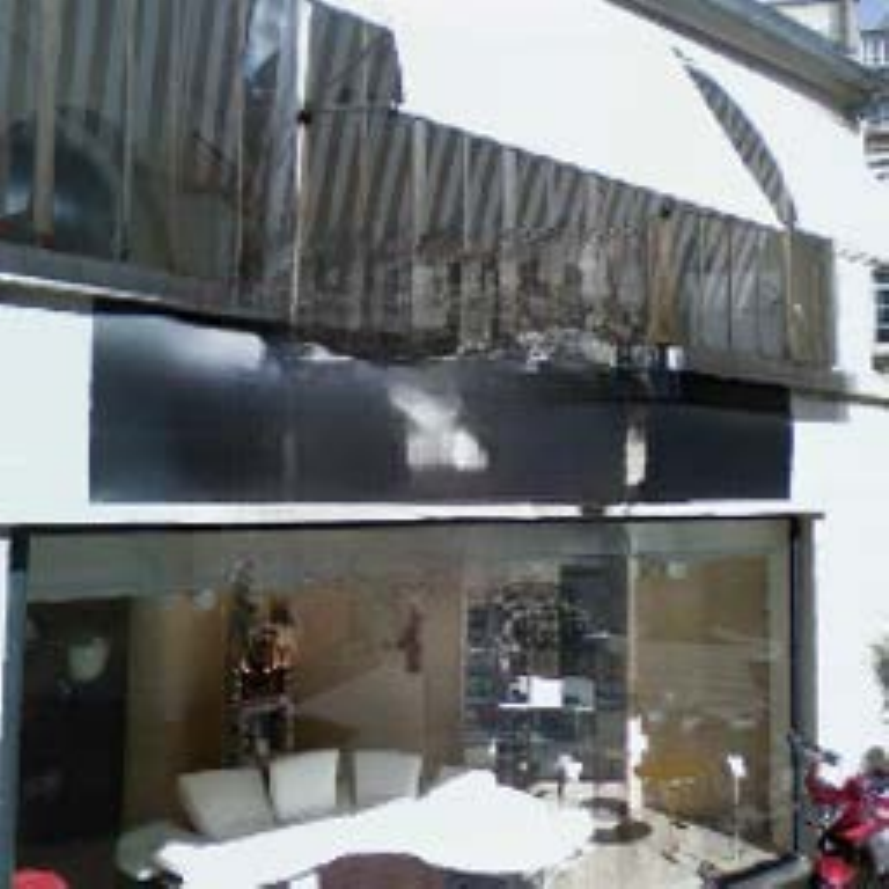} &
			\hspace{-4mm}
			\includegraphics[width=0.08\textwidth]{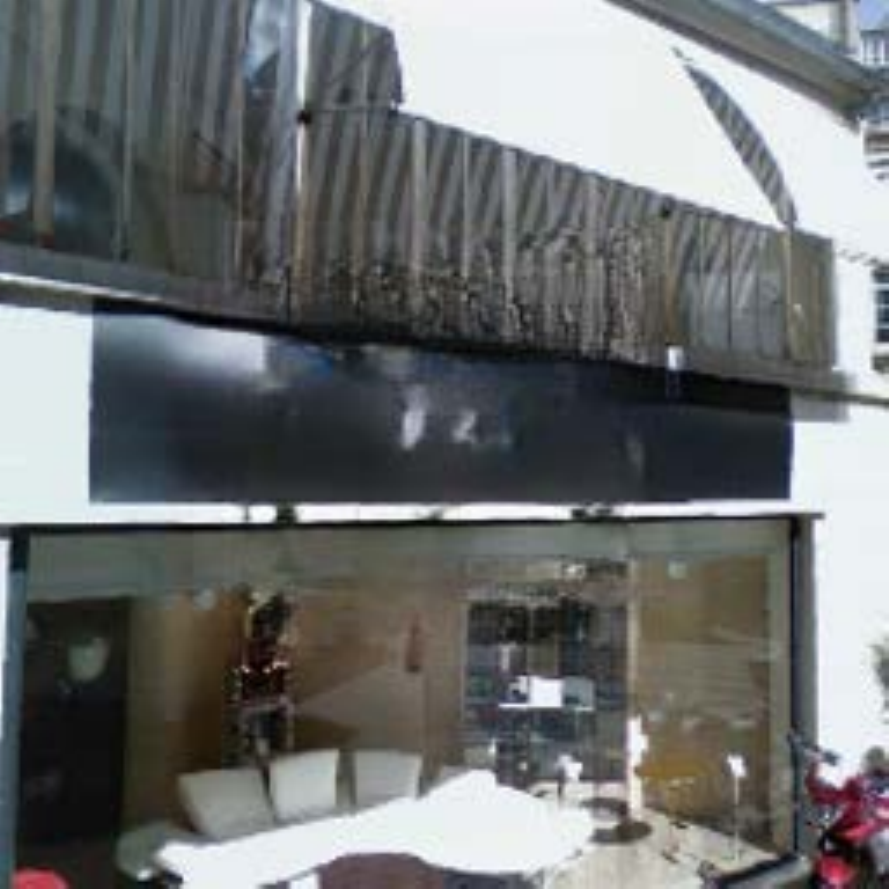} &
			\hspace{-4mm}
			\includegraphics[width=0.08\textwidth]{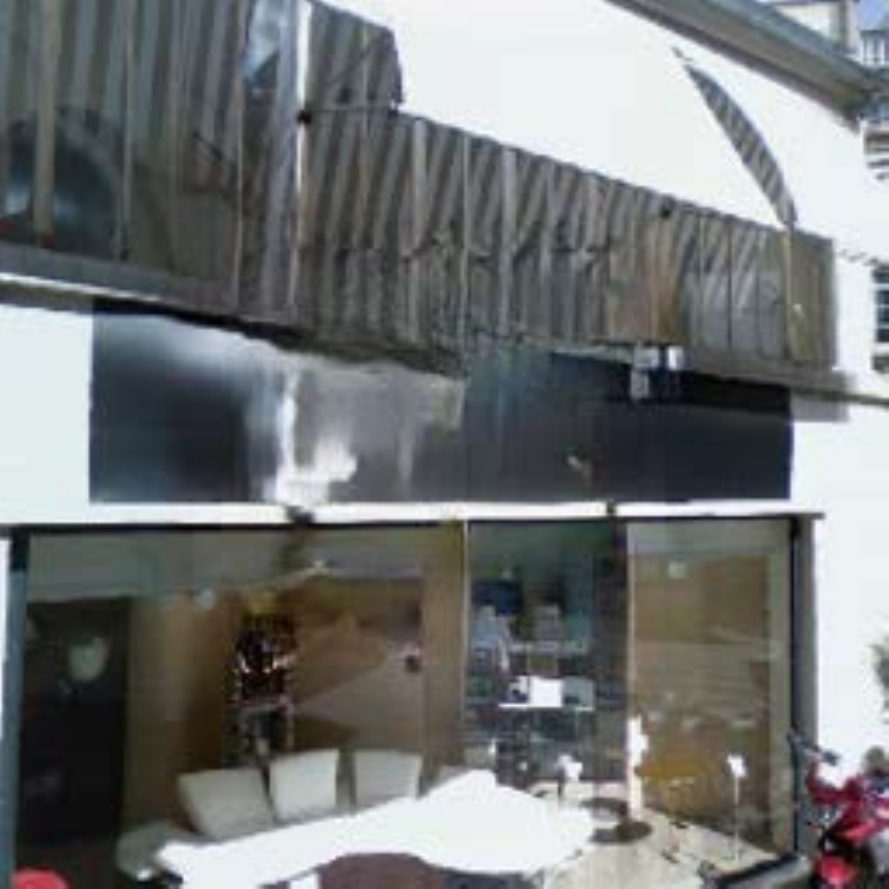} &
			\hspace{-4mm}
			\includegraphics[width=0.08\textwidth]{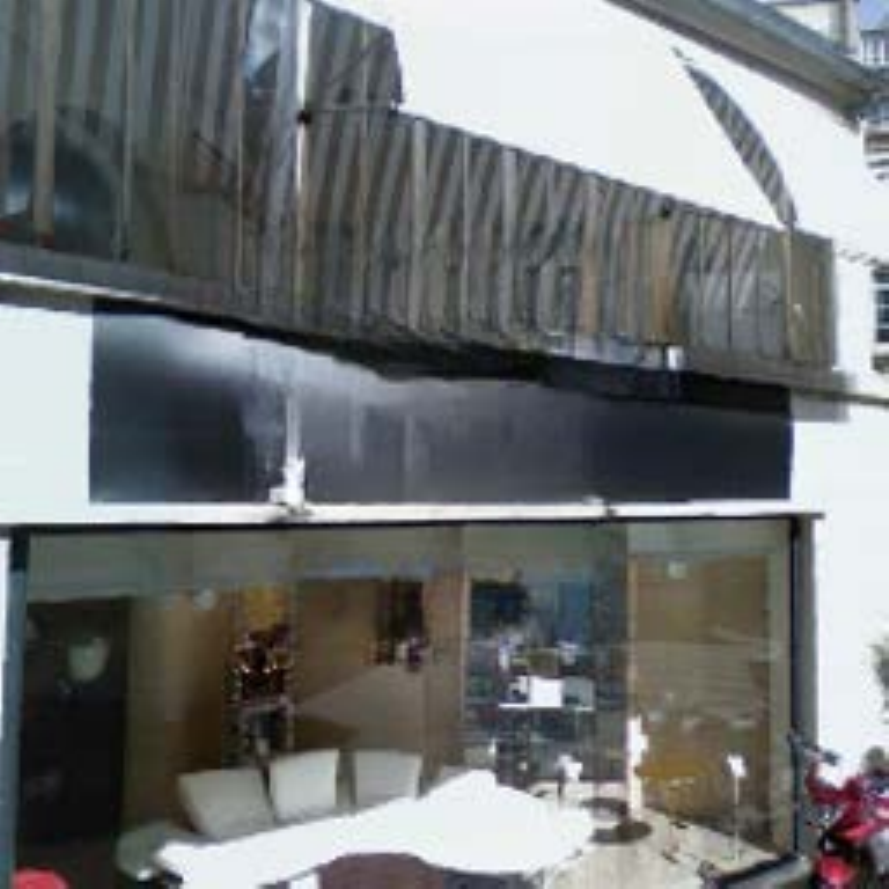} &
			\hspace{-4mm}
			\includegraphics[width=0.08\textwidth]{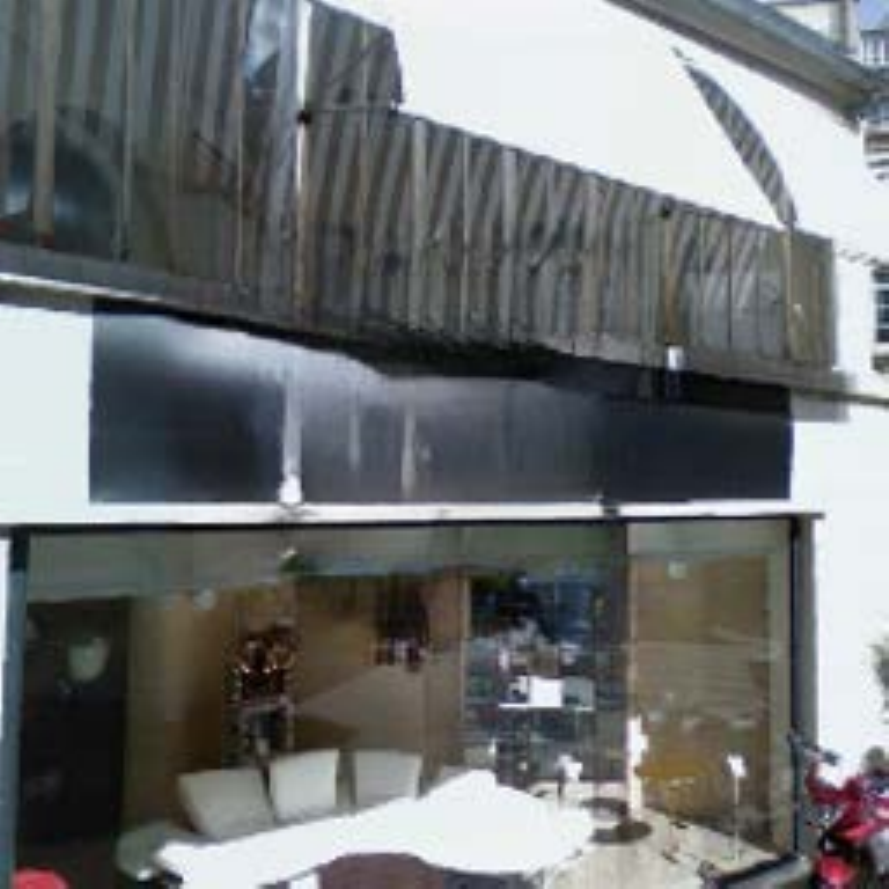} \\
			
			\includegraphics[width=0.08\textwidth]{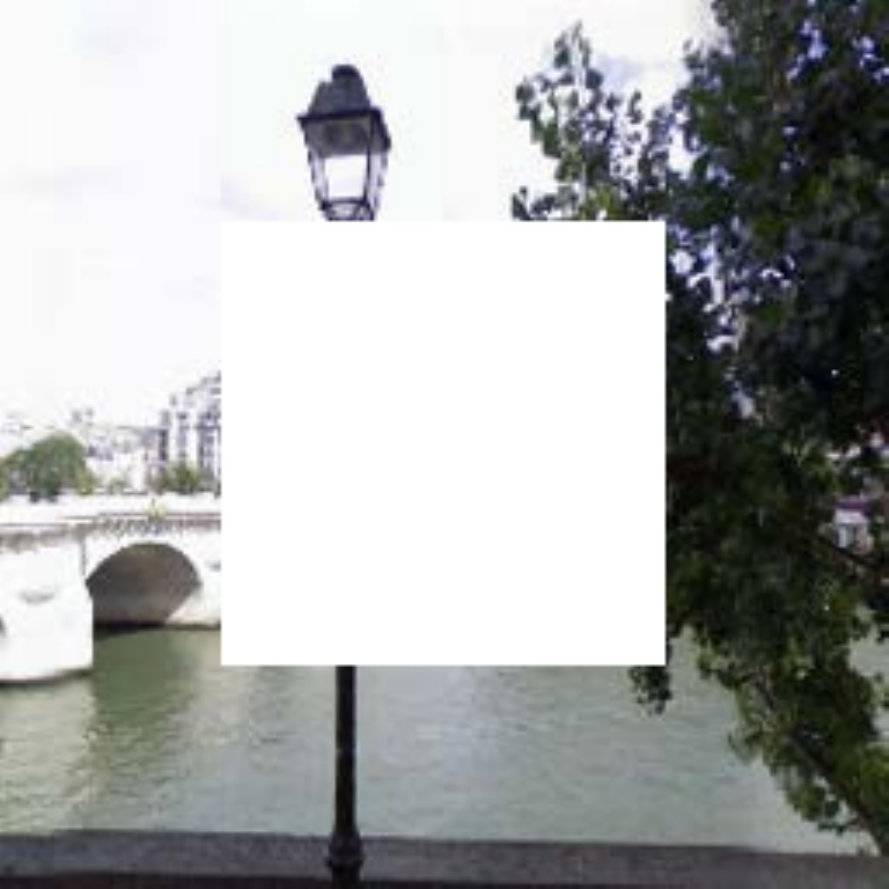} &
			\hspace{-4mm}
			\includegraphics[width=0.08\textwidth]{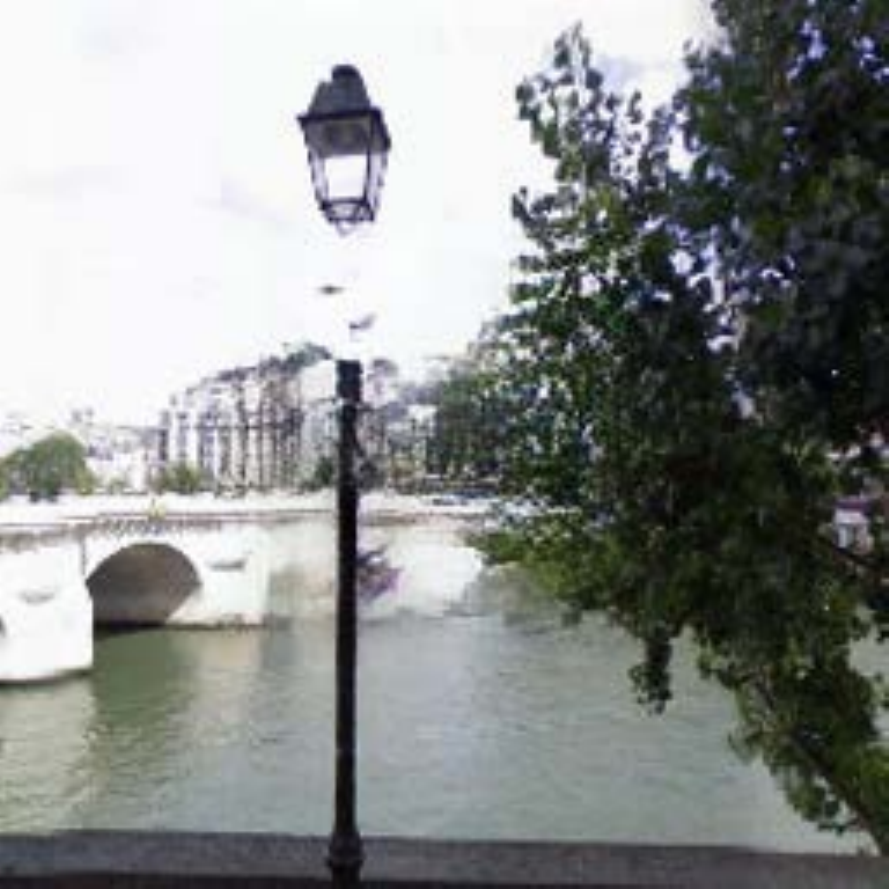} &
			\hspace{-4mm}
			\includegraphics[width=0.08\textwidth]{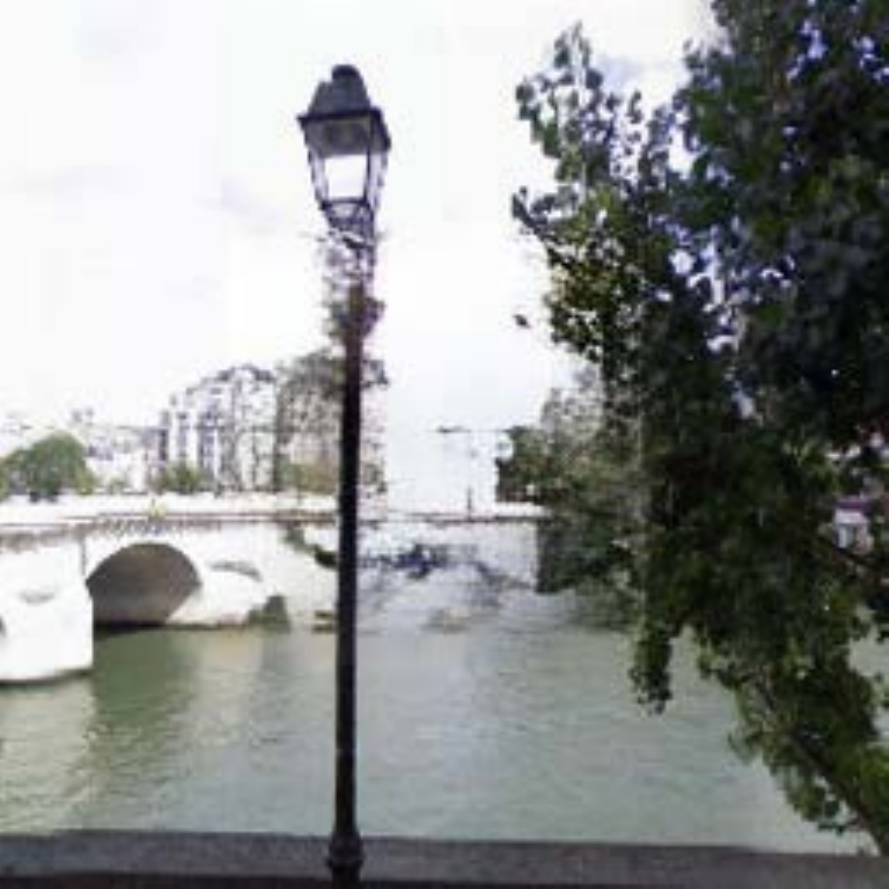} &
			\hspace{-4mm}
			\includegraphics[width=0.08\textwidth]{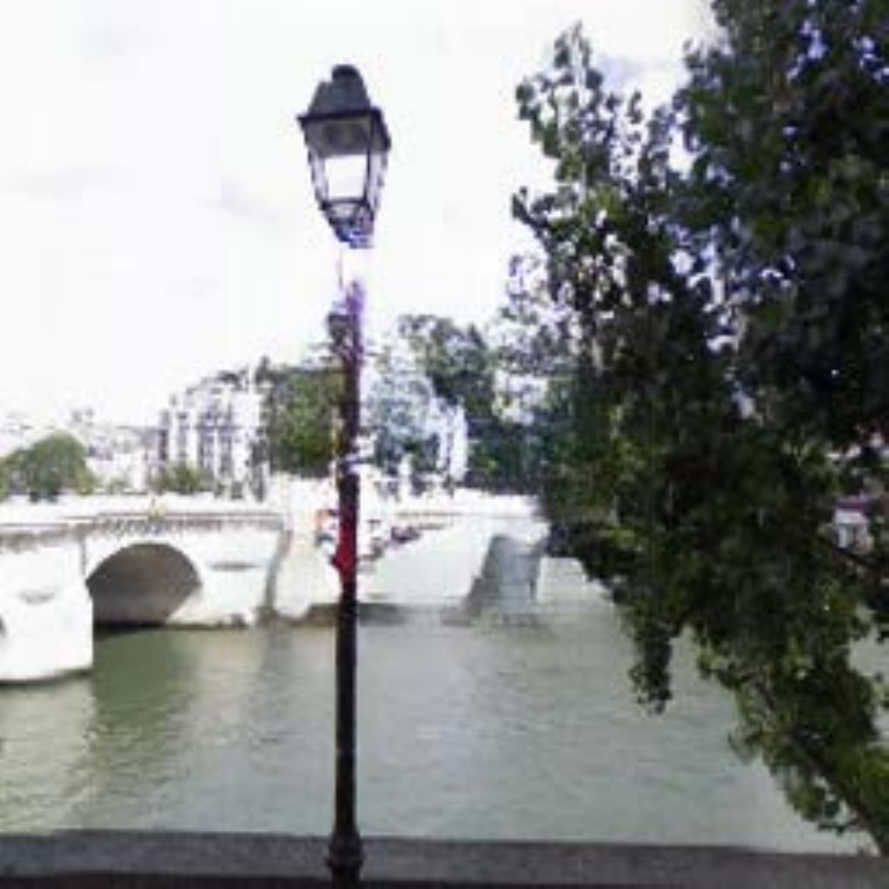} &
			\hspace{-4mm}
			\includegraphics[width=0.08\textwidth]{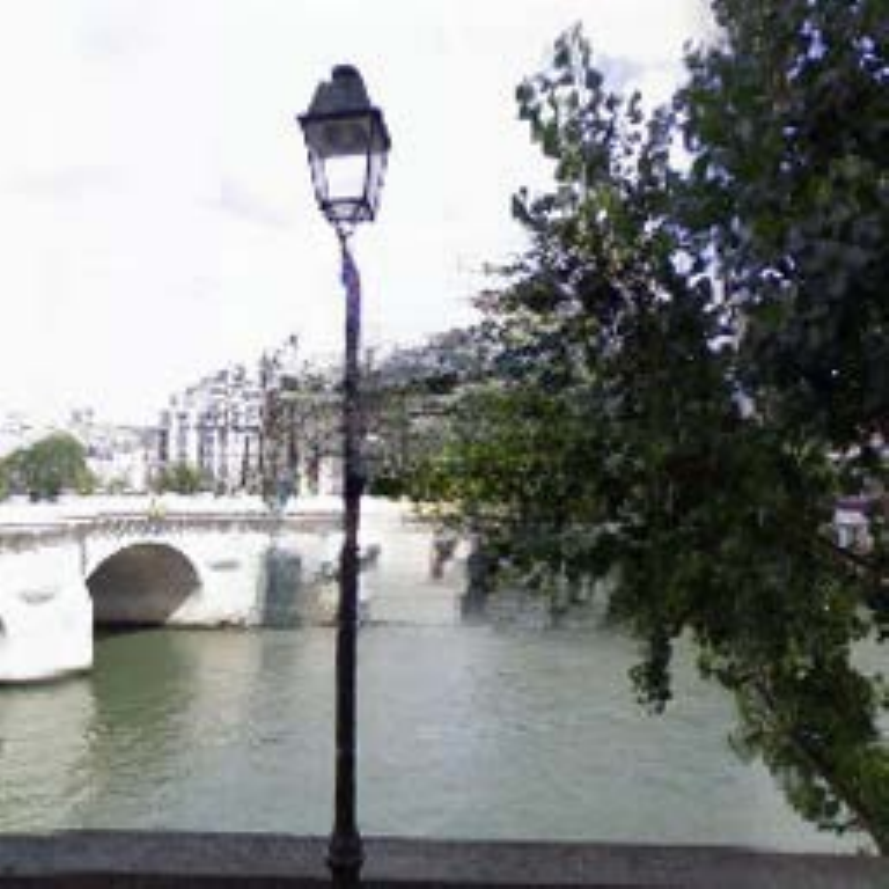} &
			\hspace{-4mm}
			\includegraphics[width=0.08\textwidth]{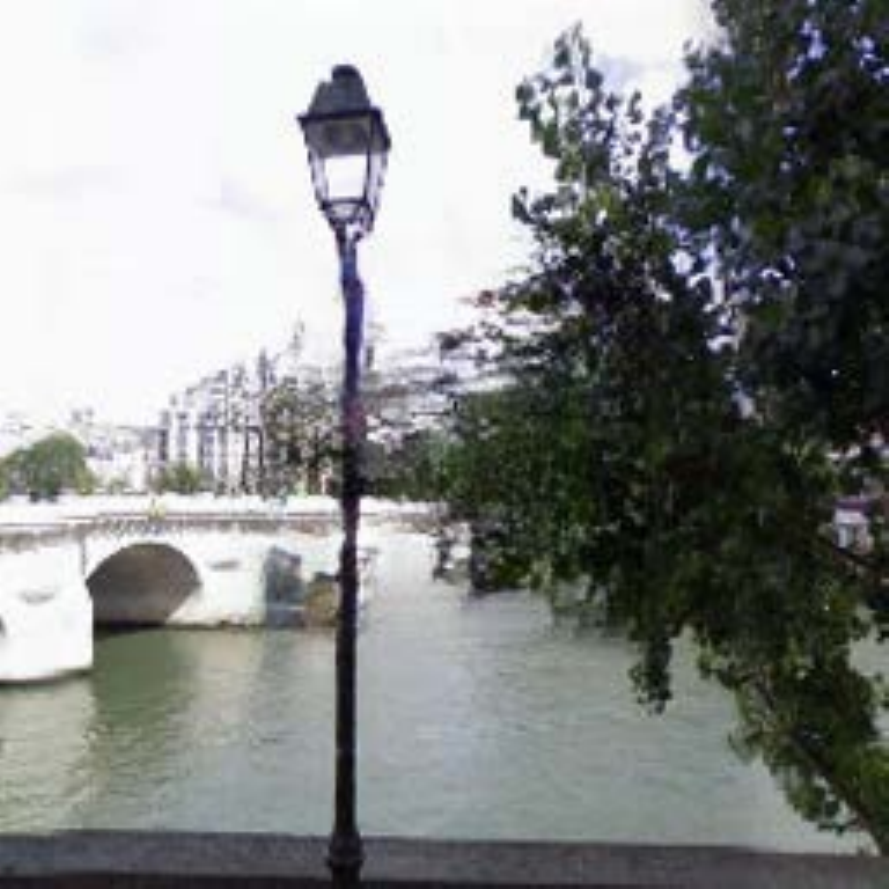} \\
			
			Input & \hspace{-4mm} rate=2 & \hspace{-4mm} rate=8 & \hspace{-4mm} w/o combination & \hspace{-4mm} w/o ${K_i}\left(  \cdot  \right)$ & DMFB (Ours)\\
	\end{tabular} }
	\caption{Visual comparison of different structures. \textit{Best viewed with zoom-in.}}
	\label{fig:ablation-structure}
\end{figure}

\begin{figure}[htpb]
	\centering
	\scriptsize
	\begin{tabular}{cccc}
		\includegraphics[width=0.115\textwidth]{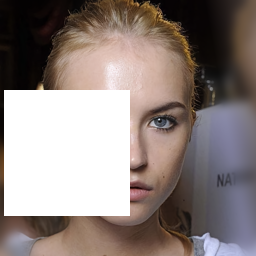} &
		\hspace{-4mm}
		\includegraphics[width=0.115\textwidth]{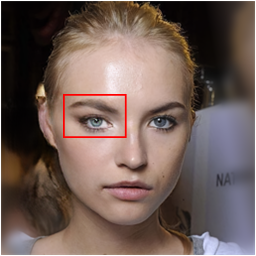} &
		\hspace{-4mm} 
		\includegraphics[width=0.115\textwidth]{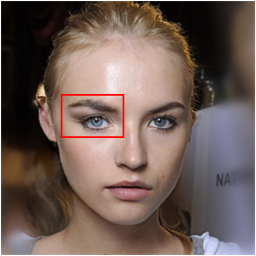} &
		\hspace{-4mm}
		\includegraphics[width=0.115\textwidth]{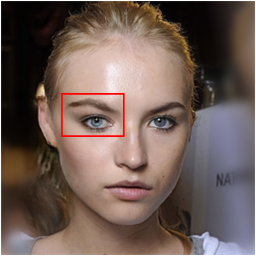} \\

		\includegraphics[width=0.115\textwidth]{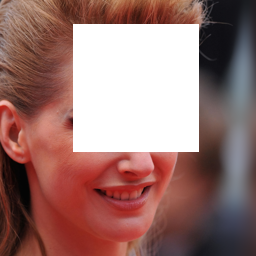}&
		\hspace{-4mm}
		\includegraphics[width=0.115\textwidth]{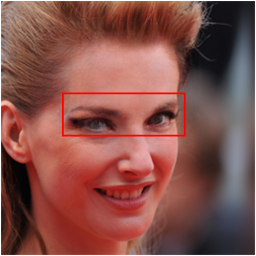}&
		\hspace{-4mm}
		\includegraphics[width=0.115\textwidth]{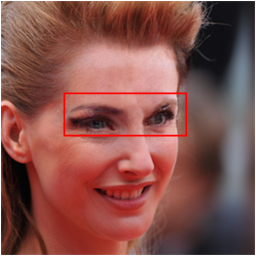}&
		\hspace{-4mm}
		\includegraphics[width=0.115\textwidth]{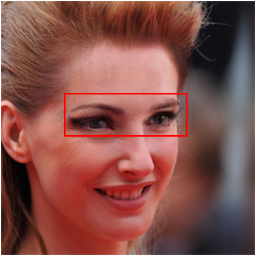} \\
		
		Input &\hspace{-4mm} w/o self-guided &\hspace{-4mm} w/o alignment &\hspace{-4mm} with all\\
	\end{tabular}
	\caption{Visual comparison of results using different losses. \textit{Best viewed with zoom-in.}}
	\label{fig:ablation-loss}
\end{figure}

\subsubsection{Effectiveness of DMFB}

To validate the representation ability of our DMFB, we replace its middle part (4 dilated convolutions and combination operation) to a $3 \times 3$ dilated convolution (256 channels) with dilation rate of $2$ or $8$ (``rate=2'' or ``rate=8'', see in Table~\ref{tab:different-structures}). Additionally, to verify the strength of ${K_i}\left(  \cdot  \right)$ in combination operation, we perform the DMFB without ${K_i}\left(  \cdot  \right)$ that denoted as ``w/o ${K_i}\left(  \cdot  \right)$'' in Table~\ref{tab:different-structures}. Combined with Table~\ref{tab:different-structures} and Figure~\ref{fig:ablation-structure}, we can clearly see that our model with DMFB (\textbf{Parms}: $471,808$) predicts reasonable and less artifact images than ordinary dilated convolutions (\textbf{Parms}: $803,392$). Specifically, in the second row of Figure~\ref{fig:ablation-structure}, both ``w/o combination'' and ``w/o ${K_i}\left(  \cdot  \right)$'' have different degrees of the partial absence of lampposts. The visual effects of them can be obviously ranked as ``DMFB'' $>$ ``w/o ${K_i}\left(  \cdot  \right)$'' $>$ ``w/o combination''. Meanwhile, the results of ``rate=2'' and ``rate=8'' suggest the importance of spatial support as discussed in~\cite{globally-and-locally}. It also demonstrates large and dense receptive field is beneficial to completing images with large holes.

\subsubsection{Self-guided regression and geometrical alignment constraint}
To investigate the effect of proposed self-guided regression loss and geometrical alignment constraint, we train a complete DMFN on CelebA-HQ dataset without the corresponding loss. Thanks to the effectiveness of the generator and the loss functions, the baseline model (w/o self-guided regression loss \& geometrical alignment constraint) can already achieve almost satisfactory results. Because of the subtle problems existing in the baseline model, we propose two loss functions to further refine the results, the so-called image fine-grained inpainting. For instance, in the first row of Figure~\ref{fig:ablation-loss}, ``w/o alignment'' shows that the eyelid lines at the left eye (in \textcolor{red}{red} box) are dislocation, the eyebrows are slightly scattered. ``w/o self-guided'' yields correct double eyelids, but eyebrows are still unnatural. ``with all'' shows the best performance. Although the qualitative performance is not much improved, these new loss functions have a corrective effect on a few problematic results produced by the baseline model. And we give the quantitative results in Table~\ref{tab:different-losses}, which validates the effectiveness of various proposed losses.

\begin{figure*}[ht]
	\centering
	\begin{tabular}{cccc}
		\includegraphics[width=0.2\textwidth]{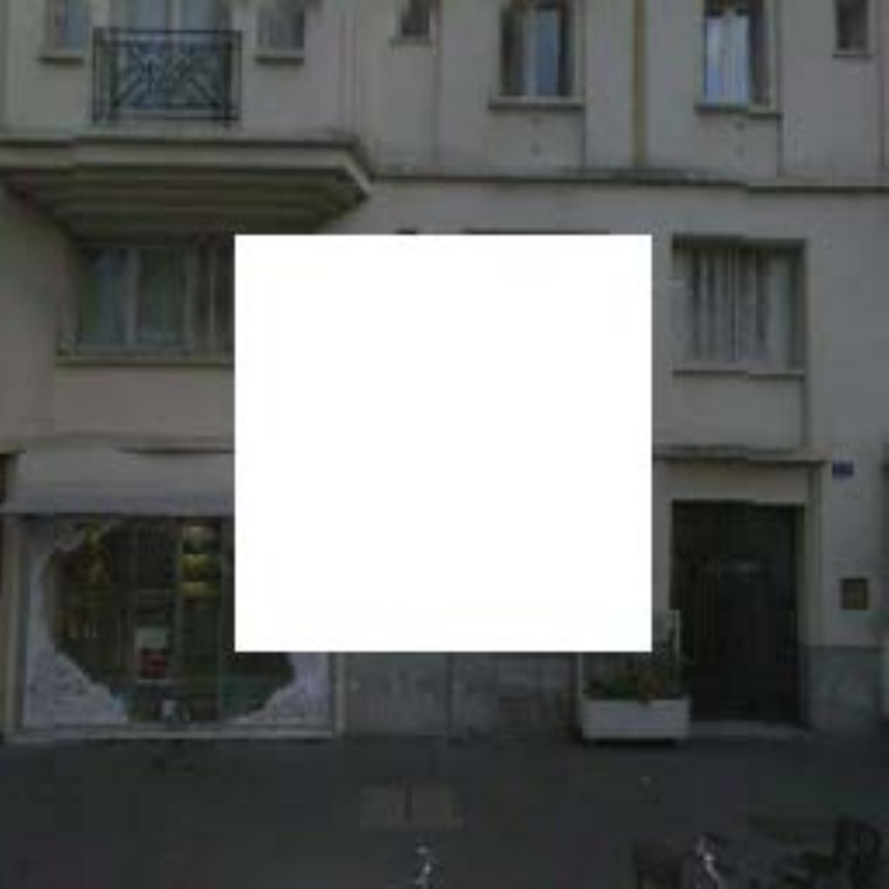} &
		\hspace{-3mm}
		\includegraphics[width=0.2\textwidth]{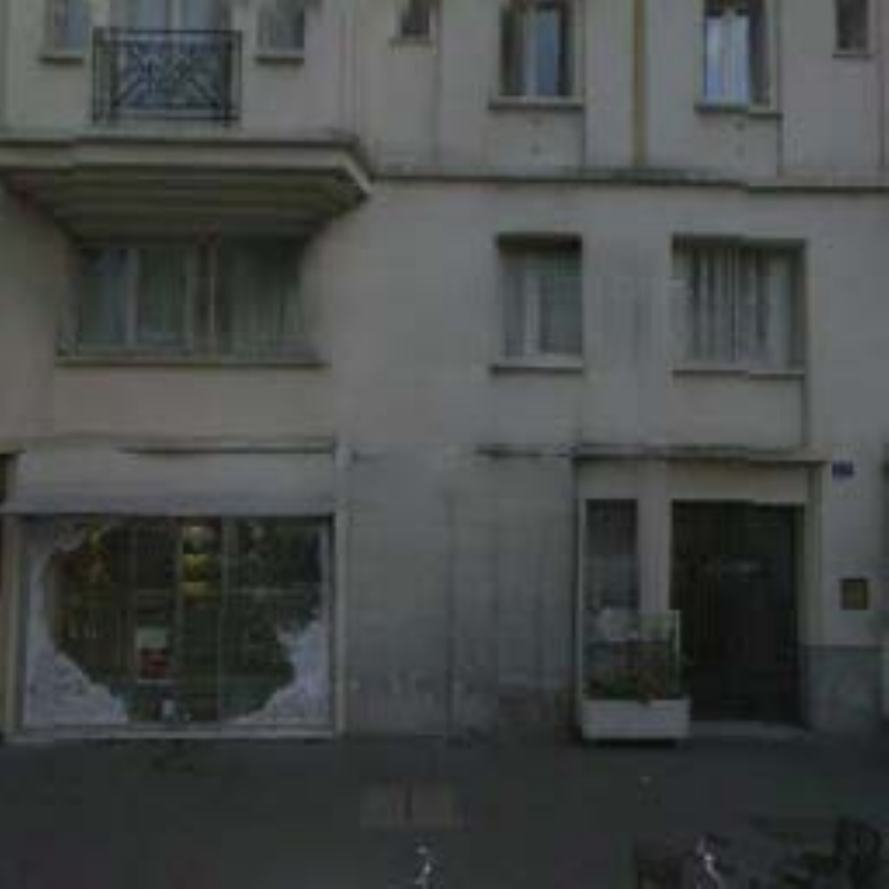} &
		\hspace{-3mm}
		\includegraphics[width=0.2\textwidth]{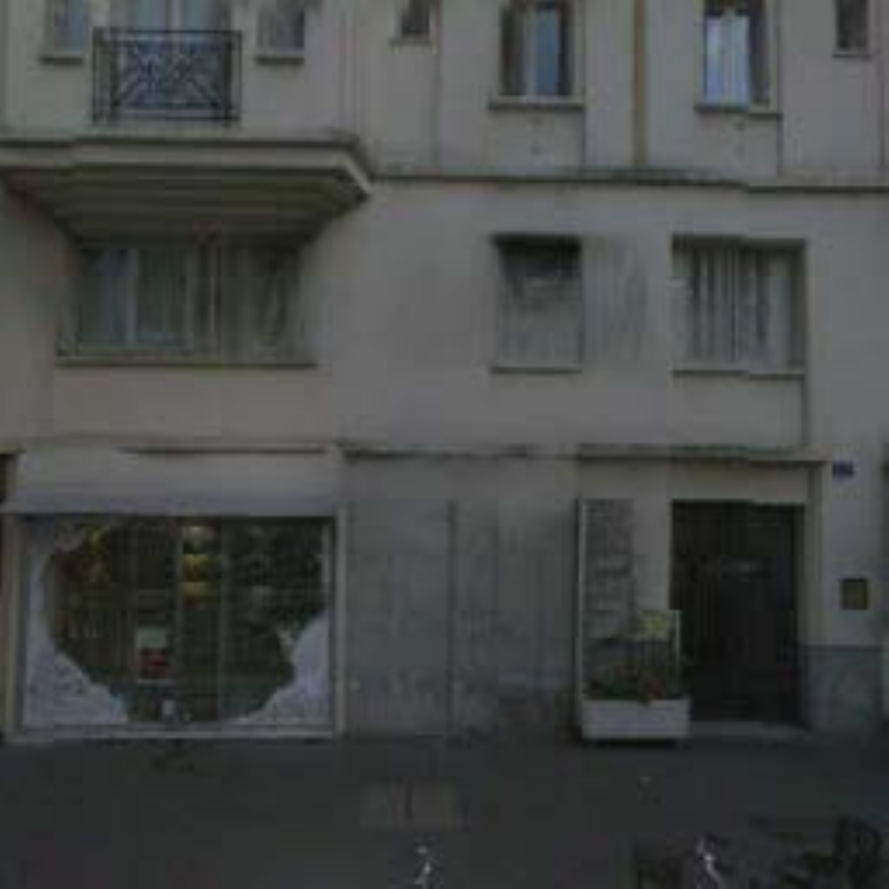} &
		\hspace{-3mm}
		\includegraphics[width=0.2\textwidth]{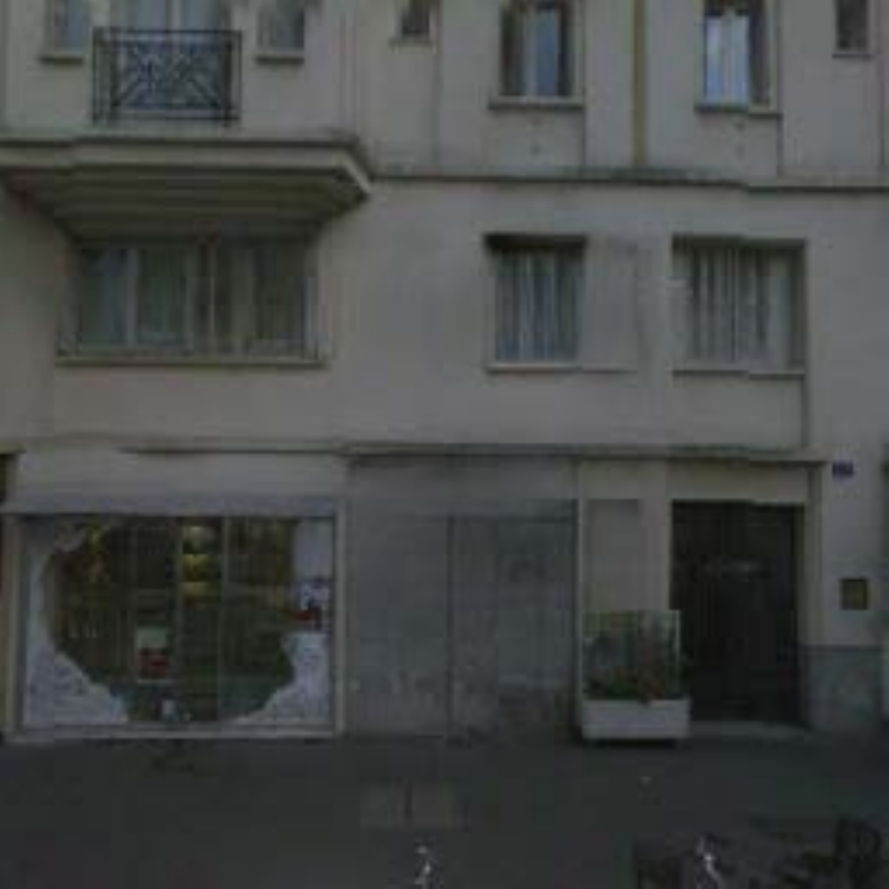} \\
		
		Input & Dot product & Gaussian distance & L2 distance \\
	\end{tabular}
	\caption{Visual comparisons on Paris street view. \textit{Best viewed with zoom-in.}}
	\label{fig:loss}
\end{figure*}

\begin{figure*}[htpb]
	\centering
	\includegraphics[width=0.9\textwidth]{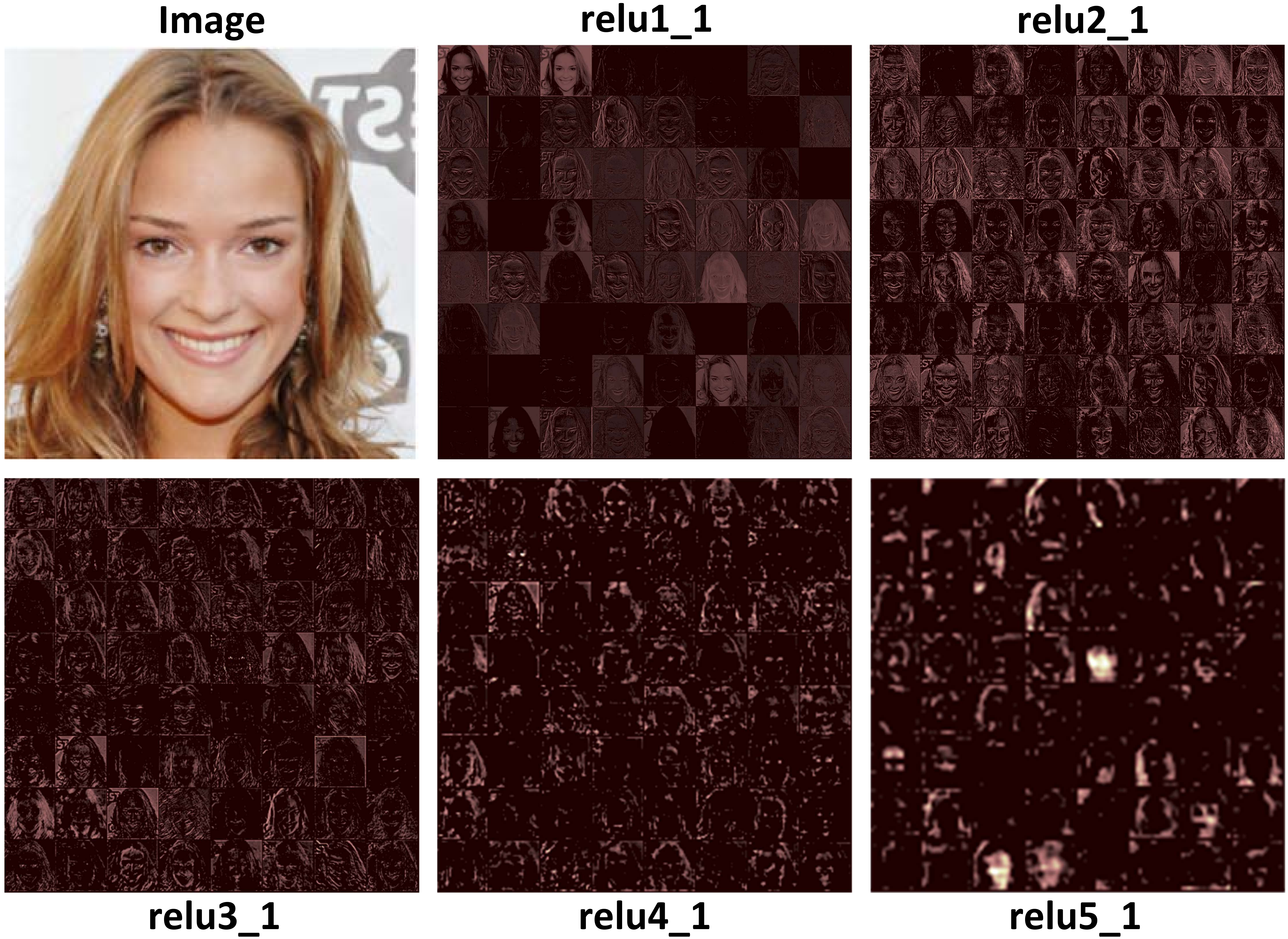}
	\caption{Visualization of VGG feature maps (\textit{the first $64$ pieces}).}
	\label{fig:visualization}
\end{figure*}

\begin{table*}[htpb]
	\centering
	\caption{The comparison of self-guided regression loss with various distance metrics on Paris Streetview. Here, ${X_i} \in \mathbb{R}{^{C \times 1}}$ represents the vector of the image $X \in \mathbb{R}{^{C \times H \times W}}$ at position $i$.}
	\label{tab:distance-metrics}
	\begin{tabular}{|c|c|c|c|c|}
		\hline
		Distance metric & $\phi \left( {{X_i},{Y_i}} \right)$ & PNSR$ \uparrow $ & SSIM$ \uparrow $ & LPIPS$ \downarrow $ \\
		\hline
		Gaussian distance & $\exp \left\{ {{{ - \left\| {{X_i} - {Y_i}} \right\|_2^2} \mathord{\left/
					{\vphantom {{ - \left\| {{X_i} - {Y_i}} \right\|_2^2} {2{\sigma ^2}}}} \right.
					\kern-\nulldelimiterspace} {2{\sigma ^2}}}} \right\}$ & \textbf{25.06} & \textbf{0.8596} & 0.1027 \\
		\hline
		Dot product & ${X_i}Y_i^T$ & 24.77 & 0.8588 & 0.1035 \\
		\hline
		L2 distance & $\left\| {{X_i} - {Y_i}} \right\|_2^2$ & 25.00 & 0.8563 & \textbf{0.1018} \\
		\hline
	\end{tabular}
\end{table*}

\subsection{Discussions}
\subsubsection{Intention of self-guided regression loss}
The prior works (\eg,~CA~\cite{contextual-attention} and GMCNN~\cite{GMCNN}) assign less weight at masked region centers to formulate the variant of L1 loss. CA only use spatial discounted L1 loss in the coarse network (first stage). GMCNN first train their model with only confidence-driven L1 loss. Without the assistant of GAN, these first stages only aim to obtain a coarse results. Different from them, the self-guided regression loss apply to VGG features focuses on learning the \textit{hard} areas measured by the current guidance map. And our framework is one-stage trained with all losses at the same time.

\subsubsection{Analysis of self-guided regression loss}
In this section, we conduct the ablation study of using different distance metrics in the average error map. Table~\ref{tab:distance-metrics} compares instantiations including \textit{Gaussian}, \textit{dot product}, and \textit{L2} when used in self-guided regression loss. The simple L2 performs the best LPIPS performance, and Gaussian achieves the best PSNR and SSIM performance. Considering that the purpose of image inpainting is the pursuit of plausible visual effect, We choose the simple and efficient L2 distance to measure the average error map. Figure~\ref{fig:loss} shows the visual comparisons among these metrics, which indicates L2 can recover the better structural information.

\subsubsection{Investigation of geometrical alignment constraint}
As illustrated in Figure~\ref{fig:visualization}, we visualize the first $64$ feature maps of each selected VGG layer. The response maps of `relu1\_1`, `relu2\_1`, and `relu3\_1` layers almost have the completed face contour, which is unsuited to aligning the part components using geometrical alignment constraint. For the spatial resolution of each response map generated by `relu5\_1` layer is only $16 \times 16$, it will result in a small coordinate range. Thus, we choose the output response maps of `relu4\_1` layer to compute our geometrical alignment constraint, which guides the coordinate expectation registration.

\section{Conclusion}\label{sec:conclusion}

In this paper, we proposed a dense multi-scale fusion network with self-guided regression loss and geometrical alignment constraint for image fine-grained inpainting, which highly improves the quality of produced images. Specifically, dense multi-scale fusion block is developed to extracted better features. With the assistance of self-guided regression loss, the restoration of semantic structures becomes easier. Additionally, geometrical alignment constraint is inductive to the coordinate registration between generated image and ground-truth, which promotes the reasonableness of painted results.

\section*{Acknowledgment}
This work was supported in part by the National Key Research and Development Program of China under Grants 2018AAA0102702, 2018AAA0103202, in part by the National Natural Science Foundation of China under Grant 61772402, 61671339 and 61871308.
\bibliographystyle{IEEEtran}
\bibliography{IEEEabrv,mybibfile}

\end{document}